%% file: main.tex
\documentclass[]{techreport}

\usepackage[utf8]{inputenc}
\usepackage[T1]{fontenc}
\usepackage{xspace}
\usepackage{graphicx}
\usepackage{amsmath}
\usepackage{amssymb}
\usepackage{booktabs}
\usepackage{url}
\usepackage{amsfonts}
\usepackage{nicefrac}
\usepackage{microtype}
\usepackage{xcolor}

\input{preamble}

\title{%
\noindent
\begin{minipage}[c]{0.1\textwidth}
    \centering
    \includegraphics[width=\linewidth]{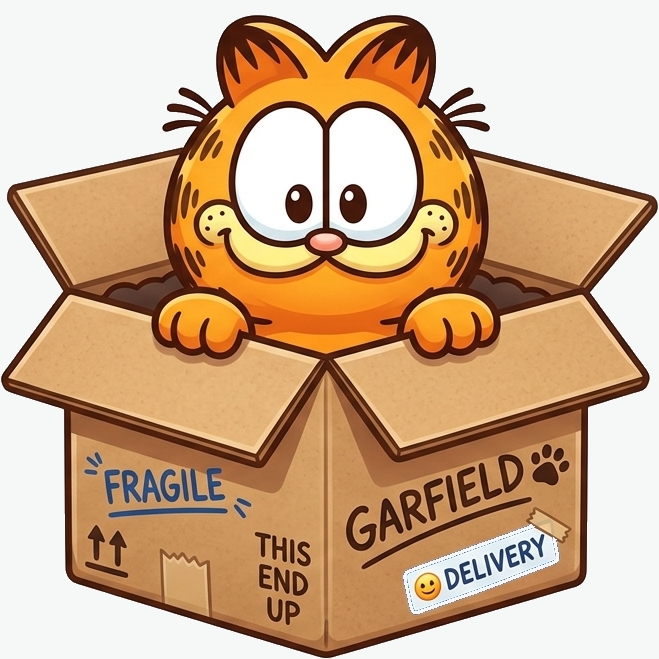}
\end{minipage}%
\hfill
\begin{minipage}[c]{0.87\textwidth}
    \LARGE Schr\"odinger's Cat: Probabilistic Representation and Prediction of Potential Scene Kinematics
\end{minipage}
}

\author{
    {\large \textbf{Timy Phan}\NoHyper\thanks{Equal Contribution.}\endNoHyper, \hspace{0.3em} \textbf{Jannik Wiese}\footnotemark[1], \hspace{0.3em} \textbf{Bj\"orn Ommer}}\\
    CompVis @ LMU Munich, Munich Center for Machine Learning (MCML)
}

\abstract{%
    \input{sec_tech/0_abstract}

    \textbf{Project Page:} \url{https://compvis.github.io/schroedingers_cat}
    \\
    \textbf{Code:} \url{https://github.com/CompVis/schroedingers_cat}
    \\
    \textbf{Accepted at ECCV 2026}
    \vspace{-3.2em}
}

\begin{document}
\maketitle

\vspace{-0.9em}

\begin{figure}[H]
    \centering
    \definecolor{myred}{HTML}{FF3C4F}
    \ifeccv\vspace{-1em}\fi
    \includegraphics[width=0.8\linewidth]{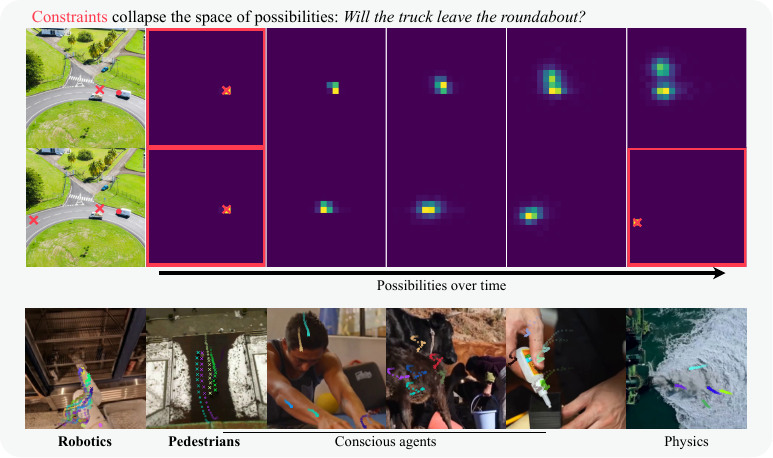}
    \caption{\textbf{Schrödinger's Cat:} Given the initial scene context as an image and a set of \emph{spatio-temporally sparse constraints or goals} (\textcolor{myred}{$\mathbf{x}$}), \ours~encodes the space of possible motions. By sampling from this space, our approach is able to plan realistic motion for conscious agents and inanimate objects, which we \textbf{evaluate} in applications such as robotics and trajectory prediction.}\label{fig:intro:teaser}
    \vspace{0.5em}
\end{figure}

\setcounter{footnote}{0}
\RenewDocumentCommand{\paragraph}{s m}{\vspace{.25em}\noindent\textbf{#2\IfBooleanF{#1}{.}}}

\input{sec_tech/1_intro}
\input{sec_tech/2_related}

\input{sec_tech/3_method}
\input{sec_tech/4_experiments}
\input{sec_tech/5_conclusion}

\input{sec_tech/X_acknowledgment}

\FloatBarrier

{
    \small
    \bibliographystyle{ieeenat_fullname}

    \bibliography{main}
}

\FloatBarrier
\clearpage

\input{sec_tech/X_suppl}

\end{document}

%% file: preamble.tex
\usepackage{adjustbox}
\usepackage{graphicx}
\usepackage{caption}
\usepackage{subcaption}
\usepackage{printlen}
\usepackage{float}
\usepackage{wrapfig}
\usepackage{multirow}
\usepackage{booktabs}
\usepackage{pifont}
\usepackage{makecell}
\usepackage{nicefrac}
\usepackage{url}
\usepackage{yhmath}
\usepackage{stmaryrd}
\usepackage{verbatim}
\usepackage{enumitem}
\usepackage{amsmath}
\usepackage{lipsum}
\usepackage{algorithm}
\usepackage{algpseudocode}
\usepackage{amsfonts}
\usepackage{microtype}
\usepackage{pgf}
\usepackage{pgfplots}
\usepackage{tabularx}
\usepackage{scalerel}
\usepackage{xspace}
\usepackage{ifthen}
\usepackage{soul}
\usepackage{cuted}
\usepackage{mathtools}
\usepackage{tikz}
\usetikzlibrary{angles,quotes,3d,math,arrows.meta,calc,positioning,fit,backgrounds,decorations.pathreplacing,calligraphy,shapes,shapes.multipart}
\usepackage[table]{xcolor}
\usepackage{lineno}

\usepackage{multirow}
\usepackage{booktabs}
\usepackage{array}
\usepackage{rotating}
\usepackage[labelfont=bf]{caption}
\usepackage{afterpage}
\usepackage{xr}
\usepackage{cancel}

\input{macros}

\ifeccv\else
    \setlist[itemize]{nosep}
\fi

\definecolor{ourgreen}{RGB}{46, 204, 113}
\definecolor{ourgreenborder}{RGB}{39, 174, 96}
\definecolor{ourblue}{RGB}{52, 152, 219}
\definecolor{ourblueborder}{RGB}{41, 128, 185}
\definecolor{ourorange}{RGB}{230, 126, 34}
\definecolor{ourorangeborder}{RGB}{211, 84, 0}
\definecolor{ourred}{RGB}{231, 76, 60}
\definecolor{ourredborder}{RGB}{192, 57, 43}
\definecolor{ouryellow}{RGB}{241, 196, 15}
\definecolor{ouryellowborder}{RGB}{243, 156, 18}
\definecolor{ourpurple}{RGB}{155, 89, 182}
\definecolor{ourpurpleborder}{RGB}{142, 68, 173}
\definecolor{ourturquoise}{RGB}{26, 188, 156}
\definecolor{ourturquoiseborder}{RGB}{22, 160, 133}
\definecolor{ourturquoise}{RGB}{26, 188, 156}
\definecolor{ourturquoiseborder}{RGB}{22, 160, 133}
\definecolor{ourwhite}{RGB}{236, 240, 241}
\definecolor{ourwhiteborder}{RGB}{189, 195, 199}
\definecolor{ourgray}{RGB}{149, 165, 166}
\definecolor{ourgrayborder}{RGB}{127, 140, 141}

\definecolor{ourwhite2}{RGB}{246, 247, 248}

\definecolor{matplotlibblue}{HTML}{1f77b4}
\definecolor{matplotliborange}{HTML}{ff7f0e}
\definecolor{matplotlibgreen}{HTML}{2ca02c}

\definecolor{ourhighlightcolor}{RGB}{46, 204, 113}

\newcolumntype{H}{>{\setbox0=\hbox\bgroup}c<{\egroup}@{}}

\newcommand{\tikzstylenodedistance}{4mm}
\newcommand{\tikzstyleinnersep}{2mm}
\newcommand{\tikzstyleminimumheight}{8.75mm}
\newcommand{\tikzstyleminimumwidth}{12mm}

\tikzset{
    node distance=\tikzstylenodedistance,
    text centered,
    anchor=center,
}
\tikzset{
    standard node/.style n args={1}{%
        rectangle,
        rounded corners=0.1cm,
        fill=our#1,
        draw=our#1border,
        line width=0.04cm,
        minimum height=\tikzstyleminimumheight,
        minimum width=\tikzstyleminimumwidth,
        inner sep=\tikzstyleinnersep,
        text centered,
        anchor=center,
        align=center,
    }
}
\tikzset{
    standard node module/.style n args={0}{%
        rectangle,
        rounded corners=0.1cm,
        fill=ourturquoise,
        draw=ourturquoiseborder,
        line width=0.04cm,
        minimum height=\tikzstyleminimumheight,
        minimum width=12mm,
        inner xsep=\tikzstyleinnersep,
        inner ysep=1mm,
        text centered,
        anchor=center,
        align=center,
    }
}
\tikzset{
    standard node image/.style n args={1}{%
        rectangle,
        fill=our#1,
        draw=our#1border,
        line width=0.04cm,
        minimum height=\tikzstyleminimumheight,
        minimum width=\tikzstyleminimumwidth,
        inner sep=0,
        text centered,
        anchor=center,
        align=center,
    }
}
\tikzset{
    standard node circle/.style n args={1}{%
        fill=our#1,
        draw=our#1border,
        circle,
        inner sep=0.1cm,
        minimum height=0,
        minimum width=0,
    }
}
\tikzset{
    standard node circle/.prefix style = standard node
}

\tikzset{
    standard line/.style n args={0}{%
        line width=0.04cm,
        rounded corners=0.1cm,
    }
}
\tikzset{
    standard arrow/.style n args={0}{%
        -latex,
    }
}
\tikzset{
    standard arrow/.prefix style = standard line
}

\tikzset{
    simple node image/.style n args={0}{%
        rectangle,
        inner sep=0,
        text centered,
        anchor=center,
        align=center,
        node distance=0mm
    }
}

\makeatletter
\newcommand{\printfnsymbol}[1]{%
  \textsuperscript{\@fnsymbol{#1}}%
}
\makeatother

\crefname{appsec}{Supp.\ Sec.}{Supp.\ Secs.}
\Crefname{appsec}{Supp.\ Section}{Supp.\ Sections}
\crefname{appfig}{Supp.\ Fig.}{Supp.\ Figs.}
\Crefname{appfig}{Supp.\ Figure}{Supp.\ Figures}
\crefname{apptab}{Supp.\ Tab.}{Supp.\ Tabs.}
\Crefname{apptab}{Supp.\ Table}{Supp.\ Tables}
\crefname{appeq}{Supp.\ Eq.}{Supp.\ Eqs.}
\Crefname{appeq}{Supp.\ Equation}{Supp.\ Equations}

%% file: macros.tex
\newif\ifeccv
\eccvfalse %

\newcommand{\methodname}{GARFIELD\xspace}
\newcommand{\ours}{\methodname}
\newcommand{\oursours}{\methodname (Ours)}
\newcommand{\Ours}{Goal-Aware Representations of Future kInEmatic Latent Distributions\xspace}
\newcommand{\Oursours}{\emph{Goal-Aware Representations of Future kInEmatic Latent Distributions} (GARFIELD)\xspace}

\ifeccv
    
\else
    
\fi

\newcommand{\image}{\mathbf{I}}

\ifeccv
    
\else
    
\fi

\makeatletter
\newcommand{\cancelasX}[1]{\mathpalette\cancelasX@{#1}}
\newcommand{\cancelasX@}[2]{%
  \ooalign{%
    \hfil$\m@th#1\cancel{\phantom{x}}$\hfil\cr
    $\m@th#1#2$\cr
  }%
}
\makeatother

\newcommand{\no}[0]{\textcolor{red}{\ding{55}}}
\newcommand{\yes}[0]{\textcolor{green}{\ding{51}}}
\newcommand{\notavail}[0]{\color{ourgray}{n/a}}

\renewcommand{\image}{\mathcal{I}}

\newcommand{\timestep}{\tau}
\newcommand{\track}{t^i}
\newcommand{\trackset}{T}
\newcommand{\trackpoint}{\track_\timestep}

\newcommand{\cond}{\mathcal{C}}

\newcommand{\condtracks}{\widehat{T}}
\newcommand{\predtracks}{\bar{T}}

\newcommand{\encoder}{\mathbf{E}_\varphi}
\newcommand{\decoder}{\mathbf{D}_\theta}

\newcommand{\decoderhead}{\mathbf{D}^{\text{p}}_\psi}
\newcommand{\decoderdense}{\mathbf{D}^{\text{d}}_\omega}

\newcommand{\latent}{\mathbf{z}}
\newcommand{\indplatent}{\mathbf{z}_{\timestep}^i}

%% file: sec_tech/0_abstract.tex
Predicting how a scene may evolve from partial observations requires reasoning about multiple possible futures rather than committing to a single trajectory.
Existing approaches either generate appearance-dominated video predictions or sample a small number of trajectories without explicitly modeling the distribution of possible motion.
We introduce \Oursours, a probabilistic model of scene kinematics that learns a \emph{structured spatio-temporal latent representation} of the distribution over possible futures given an image and optional \emph{spatio-temporally sparse constraints}.
The same latent representation enables both joint sampling of all trajectories and direct access to the underlying motion distribution through an efficient deterministic density decoder.
As a result, uncertainty about future motion can be localized to specific scene elements and timesteps and progressively refined through additional constraints.~Experiments demonstrate strong motion planning performance competitive with large video generation models while sampling trajectories $97\times$ faster.
Our method further estimates motion densities two orders of magnitude faster than Monte-Carlo sampling from motion generation models, enabling interactive exploration and uncertainty-aware planning.

%% file: sec_tech/1_intro.tex
\section{Introduction}
\label{sec:intro}

Our world is constantly unfolding and every moment carries multiple possible futures: the future is fundamentally under-determined given the present due to stochasticity, free will, and other hidden factors.
Acting in such environments requires anticipating how a scene might change. In particular, we need to predict scene kinematics -- how its constituents may move -- estimating the probability of \emph{what could possibly be} rather than merely perceiving and tracking \emph{what has already happened}.
Video models used as world simulators~\cite{alonso2024diffusionworldmodelingvisual, genie3,hafner2025trainingagentsinsidescalable} generate the future as sequences of images, producing individual frames rather than explicitly modeling scene changes between frames.
Much of their capacity is spent modeling appearance details that do not affect scene kinematics, inducing unnecessary latency and limiting exploration of possibilities.
Modeling motion directly~\cite{pandey2024motionmodeshappennext, boduljak2025happensnextanticipatingfuture, bharadhwaj2024track2act} improves efficiency and focuses capacity on kinematics, but existing approaches predict dense motion fields, wasting computational effort on static background instead of targeting relevant scene elements.
Moreover, generative motion models sample individual realizations rather than predicting the probability of various potential outcomes, their distribution, and uncertainty.

We take inspiration from Schr\"odinger’s famous thought experiment: under limited observability, the state of a system is represented as a distribution of possible states, which collapses as new observations become available.
Analogously, the key idea of \Oursours~is to represent the future as a \emph{distribution over trajectories of scene constituents} that can be \emph{progressively refined} by incorporating additional observations or constraints.
Our encoder takes an image of a scene together with possible intermediate or goal positions to 
produce a latent representation that encodes the distribution over possible future movements of scene elements. More constraints consistently sharpen this distribution, providing users with effective interactive control.

This kinematics distribution must be localized to specific scene elements and timesteps, so that uncertainty can be attributed correctly.
We therefore learn a probabilistic embedding of scene kinematics using a joint motion encoder of the entire scene that produces \emph{spatio-temporally localized latent variables}.
Training with a generative decoder ensures that the representation captures the space of possible motions rather than a single trajectory.
The structured latent space, which allows inspecting possible motion of specific elements at specific times is enforced by training the encoder with a point-wise decoder model.
We propose efficient decoders which -- based on the same latent representation -- enable both {(i)} joint trajectory sampling that captures interdependencies, as well as a {(ii)} \emph{deterministic density decoding} that produces probability heatmaps in a single ViT forward pass.
The density decoder estimates densities and uncertainty orders of magnitude faster than Monte-Carlo sampling, which is crucial for real-time planning and fast exploration of uncertainty.

We consider the task of \emph{motion planning}: generating trajectories that satisfy sparse goals while respecting scene constraints. Prior work typically relies on conditioning signals such as goal images~\cite{bharadhwaj2024track2act}, temporally dense trajectories~\cite{shi2024motioni2v}, or text prompts~\cite{shi2024motioni2v, yang2024cogvideox, HaCohen2024LTXVideo}, which are either ambiguous or difficult to obtain. In contrast, we enable \emph{spatio-temporally sparse conditioning}, specifying constraints only for selected objects and timesteps while leaving the remaining motion for the model to infer. Coupled with fast density estimation, this enables efficient exploration of possible futures and uncertainty-aware motion planning with minimal user input.

\textbf{Contributions:} We learn a (i) \emph{structured probabilistic latent representation of motion} (\cref{sec:poss_encoder}).
The same latent representation enables (ii) jointly sampling \emph{mutually consistent trajectories} that respect provided constraints (\cref{sec:met:coherent_traj}) and (iii) direct access to the underlying motion distribution via an efficient \emph{density decoder} (\cref{sec:met:density_decoder}, \cref{fig:exp:qual_merged}).
Our model further supports (iv) \emph{flexible spatio-temporally sparse conditioning}, allowing users to shape the distribution of possible futures through sparse constraints (\cref{fig:exp:more_cond_more_certv2}).
As a result, \Oursours~enable uncertainty-aware motion planning and interactive exploration of future possibilities (\cref{sec:met:interactive_cond}, \cref{fig:exp:policy_comparison}).
Our approach samples full trajectories $97\times$ faster than video world models (\cref{tab:exp:motion_comparison}) and estimates motion densities \emph{two orders of magnitude faster} than Monte-Carlo sampling from motion prediction methods (\cref{fig:exp:energy_v_latency}).

%% file: sec_tech/2_related.tex
\section{Related Work}\label{sec:relatedwork}

Recently, video world models~\cite{yang2024cogvideox, HaCohen2024LTXVideo, wan2025wan, brooks2024video, alonso2024diffusionworldmodelingvisual, hafner2025trainingagentsinsidescalable, genie3} based on diffusion transformers~\cite{peebles2023scalablediffusionmodelstransformers} have become the de facto standard approach to predict future scene evolution.
These models are typically conditioned on a start image, text, and sometimes action classes, producing videos that depict future events in full pixel detail.
They achieve impressive results in robotics~\cite{wen2024vidman, zhu2025unified}, autonomous driving~\cite{bartoccioni2025vavim, zhang2025epona, wang2023drivedreamer}, and physical simulation~\cite{brooks2024video, genie3, lu2023vdt}, but represent the future as a sequence of images instead of focusing on dynamics. The motion in the generated pixels can be extracted and analyzed post-hoc, but this approach adds additional overhead on top of the already expensive video synthesis.

Low-level temporal dynamics can be represented as dense optical flow~\cite{horn1981determining,teed2020raft} or sparse point tracks~\cite{karaev2024cotracker3, zholus2025tapnexttrackingpointtap}, which capture motion without modeling appearance.
Recent methods such as Track2Act~\cite{bharadhwaj2024track2act} and \textit{What Happens Next?}~\cite{boduljak2025happensnextanticipatingfuture} apply diffusion transformers to generate point tracks for motion-centric tasks (e.g.~robotic control).
Similarly, Motion-I2V~\cite{shi2024motioni2v} generates optical flow from a start image, text prompt, and spatially sparse conditioning tracks. The sampled flow then conditions a video generator to improve quality and control.
However, these approaches rely on ambiguous text~\cite{boduljak2025happensnextanticipatingfuture,stracke2026zipmo} or unavailable goal-state images~\cite{bharadhwaj2024track2act}.
In contrast, \ours~conditions on spatio-temporally sparse goal states.
Moreover, these methods predict motion on dense grids, wasting compute on mostly static background instead of focusing on relevant scene elements.

In comparatively narrow application domains such as autonomous driving and pedestrian prediction, methods like Social-LSTM~\cite{alahi2016social}, Social-GAN~\cite{gupta2018social}, and Trajectron++~\cite{salzmann2021trajectron} already focus on predicting motion only for relevant scene elements.
This highlights the practical value of element-centric modeling.
However, these models are tailored to specific domains and do not generalize to open-set motion prediction conditioned on arbitrary scene appearance.

Tracks and optical flow capture low-level motion but remain agnostic to semantics and constraints.
Recent work therefore explores richer motion representations, including training with VAEs~\cite{chen2021personalized}, normalizing flows~\cite{henter2020moglow}, or self-supervised representation learning~\cite{suris2022representing,wang2023fend,ressler2025dismo}.
Motion Diffusion Autoencoders~\cite{richardson2025motion} further learn semantic motion representations in closed-domain settings.
However, most approaches focus on reconstruction or semantics and do not model the stochasticity of future motion.
Early works represented this stochasticity using probabilistic motion primitives~\cite{paraschos2013probabilistic}, B\'ezier curve distributions~\cite{hug2020introducing}, or Gaussian processes~\cite{kihwankim2011gaussian}.
Recent diffusion models~\cite{shi2024motioni2v, bharadhwaj2024track2act, boduljak2025happensnextanticipatingfuture, stracke2026zipmo, baumann2026envisioning} enable sampling possible futures but do not provide direct access to the underlying distribution, requiring expensive Monte-Carlo sampling to estimate distributions.
Motion Modes~\cite{pandey2024motionmodeshappennext} improves exploration of this distribution through guided sampling but remains limited by high inference cost.
In contrast, \ours~provides access to a non-parametric distribution over future positions in constant ($\mathcal{O}(1)$) time.

The Flow Poke Transformer (FPT)~\cite{baumann2025whatif} made a significant step towards open-set, distribution-centric motion modeling by predicting the parameters of a Gaussian Mixture Model (GMM) over future positions given a set of spatially sparse known movements.
However, the model is limited by the functional form of their output distribution, constraining the output space to those possibilities that can be represented by a GMM.
More importantly, their approach only predicts distributions at one specific future timestep and does not account for scene evolution over multiple steps or allow specifying intermediate goals.

We propose \ours~to train an open-set motion foundation model that encodes possible future positions of scene elements given appearance cues and optionally sparsely known future goals. We further introduce decoder models that provide access to the underlying non-parametric distribution without repeated sampling and probabilistic sampling of future trajectories.
Our general model can be applied to application domains either in a zero-shot setting or with minimal fine-tuning, highlighting the advantages of scalable open-set pretraining.

%% file: sec_tech/3_method.tex
\section{\Ours}\label{sec:method}

\begin{figure}[tb]
  \centering
  \includegraphics[width=.85\textwidth]{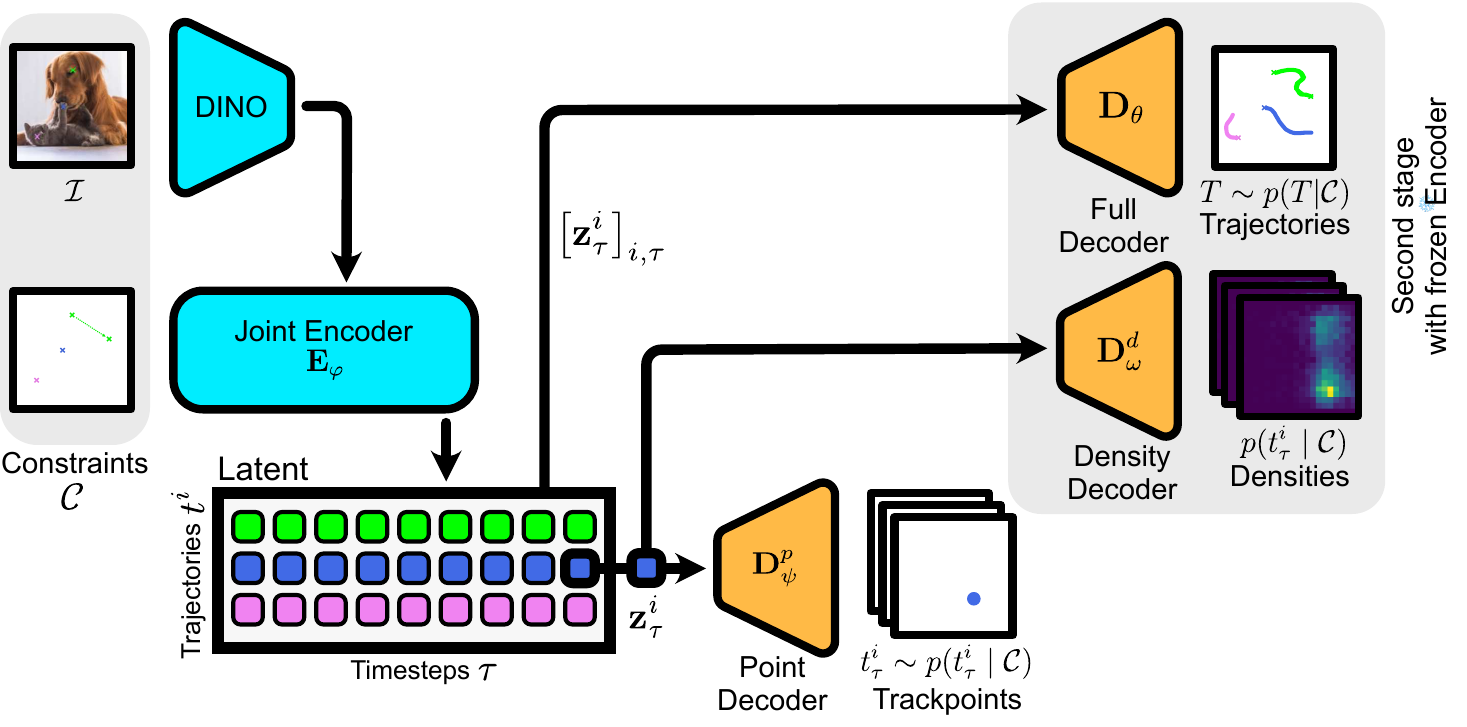}
  \caption{
  \textbf{\ours~Architecture Overview}.
  Given a set of constraints $\cond$, namely an image $\image$ and kinematics $\condtracks$, our encoder produces conditional latents $\{\indplatent\}$ representing distributions over possible future kinematics. These can be decoded into point estimates $\trackpoint$ with $\decoderhead$, densities $p(\trackpoint)$ with $\decoderdense$, and sets of kinematics $\trackset$ with $\decoder$.}
  \label{fig:method:overview}
\end{figure}

Predicting scene kinematics given a finite set of constraints or goals requires modeling the probability distribution over many plausible futures.
We represent one possible realization by the motion of scene elements $i$ over time, described by trajectories $t^i = (t_1^i, t_2^i, \dots)$, where $t_\tau^i \in \mathbb{R}^2$ denotes the position of element $i$ in the 2D image plane at timestep $\tau$.
The distribution $p(\trackset)$ of possible futures, with $T = \{t_\tau^i\}_{i,\tau}$, can be narrowed and steered by conditioning on constraints or goals.
Analogously to the \textit{``Schrödinger's Cat''} thought experiment, additional constraints or observations collapse the space of possible futures towards trajectories that match the conditioning.

Providing scene context by means of a start frame $I$ and a sparse set of positions $\condtracks \subset T$ through which certain trajectories must pass reduces uncertainty by partially controlling scene movement through intermediate goals.
\Oursours~therefore aim to model the conditional distribution $p(\bar{T} \mid I, \condtracks)$ over the unknown trajectory points $\bar{T} = T \setminus \widehat{T}$. For brevity, we summarize the conditioning via start frame and goal positions as $\cond = (I, \condtracks)$.

\subsection{Learning the Distribution of Scene Kinematics}\label{sec:poss_encoder}

We learn a latent representation of the distribution of plausible scene kinematics given scene context and constraints $\cond$.
Our goal with \ours~is twofold: (i) to jointly sample trajectories of all scene elements,
$\predtracks \sim p(\predtracks \mid \cond)$, and (ii) to estimate the probability density
$p(t_\tau^i \mid \cond)$ for the position of a scene element $i$ at time $\tau$.
Modeling such point-wise probability densities enables efficient parallel exploration of motion distributions across the spatio-temporal volume and reveals how changes to $\cond$ affect individual scene elements at specific points in time.

By annotating video data with an optical tracker~\cite{zholus2025tapnexttrackingpointtap, karaev2024cotracker3}, we can harness a rich and scalable source of data for learning and understanding scene kinematics.
However, the videos and their track annotations are only one possible realization $\trackset \sim p(\trackset)$ of kinematics for the respective scene $I$.
This leaves the distribution of \emph{possible} kinematics out of reach and thus there is no direct ground truth for $p(\trackset)$.
Generative modeling is an established way to capture a data distribution without direct access to its underlying density function by learning from samples of that data distribution.

Current methods~\cite{shi2024motioni2v,bharadhwaj2024track2act,boduljak2025happensnextanticipatingfuture,baumann2026envisioning,stracke2026zipmo} can sample $\trackset$ with generative models which underlines the viability of learning scene kinematics from video data.
However, the ability to sample $\trackset$ does not equate to easily accessible densities $p(\trackpoint | \cond)$.
In theory, the density function $p(\trackpoint | \cond)$ can be approximated with Monte-Carlo estimation by drawing multiple realizations of $\trackset$.
In practice, reliable estimates require many samples and the expensive sampling procedure of generative models renders this avenue infeasible.

Because both $\predtracks \sim p(\predtracks | \cond)$ and $p(\trackpoint | \cond)$ are conditioned on the same $\cond$, we propose to learn a shared latent representation which represents $\cond$ and informs both sampling and density estimation. Density estimation $p(\trackpoint | \cond)$ needs to be spatio-temporally localized. Thus, we factorize the latent
\begin{equation}\label{eq:met:latent_factorization}
    \latent = \big [ \indplatent \big]_{i,\tau}
\end{equation}
into separate $\indplatent$ components which each corresponds to a specific point-wise motion distribution $p(\trackpoint | \cond)$. This representation is learned end-to-end with a joint encoder $\encoder$, which produces $\latent$, and a point-wise decoder $\decoderhead$ which samples track positions from the point-wise distribution $\trackpoint \sim \decoderhead(\indplatent) \approx p(\trackpoint | \cond)$. We train $\encoder$ and $\decoderhead$ jointly to enforce this point-wise structural property in $\latent$.

\subsection{Sampling-free Density Estimation}\label{sec:met:density_decoder}

Latent components $\indplatent$ represent the distribution of possible positions of scene elements $i$ at time $\tau$.
Using the point-wise decoder $\decoderhead$ with $N_{MC}$ different initial noise seeds allows sampling $N_{MC}$ track points from the point-wise distribution and thus enable a Monte-Carlo (MC) estimation of the outcome probabilities.
However, this presents a trade-off: accurate estimation requires many samples, leading to substantial computational overhead.

To circumvent this trade-off, we propose a deterministic density estimator $\decoderdense$ for $\indplatent$ with \ours.
We define the marginal probability density on a regular grid $\mathcal{G} = \{1,\dots,g\} \times \{1,\dots,g\}$ with $g \in \mathbb{N}$ as
\begin{equation}
    p_\omega^{(u,v)}(\indplatent) = \text{P}\!\left(\trackpoint \in \mathcal{B}_{u,v} \mid \indplatent\right),
\end{equation}
which gives the probability that $\trackpoint$ lies within the spatial region $\mathcal{B}_{u,v}$ corresponding to grid location $(u,v)\in\mathcal{G}$. We interpret the grid $\mathcal{G}$ relative to the conditional image $\image$, such that each bin $\mathcal{B}_{u,v}$ covers a patch within $\image$.

The density estimator then directly predicts the point-wise non-parametric distribution over the possible future motion for scene element $i$ at timestep $\timestep$ given the known context $\cond$ through $\indplatent$:
\begin{equation}\label{eq:met:decoderdense}
    \decoderdense(\indplatent) = \{p_\omega^{(u,v)}(\indplatent)\}_{(u,v) \in \mathcal{G}} = p(\trackpoint | \indplatent) \approx p(\trackpoint | \cond)
\end{equation}
We train the density estimator $\decoderdense$ with a grid cross-entropy loss
\begin{equation}\label{eq:met:grid_ce}
    \begin{aligned}
        \mathcal{L}_{\text{grid}}
        &= - \sum_{(u,v)\in\mathcal{G}} \mathbf{1}\!\left[\trackpoint \in \mathcal{B}_{u,v}\right] \log p_\omega^{(u,v)}(\trackpoint | \indplatent),
    \end{aligned}
\end{equation}
where $\mathbf{1}\!\left[\trackpoint \in \mathcal{B}_{u,v}\right]$ is $1$ if $\trackpoint$ falls into bin $\mathcal{B}_{u,v}$ and $0$ otherwise, resulting in a one-hot target distribution.

Note that the density estimator also receives only the point-wise latent $\indplatent$ as input.
Thus, $\decoderdense$ is a decoder of the distribution encoded in the latent representation, learned through generative pre-training. Instead of estimating densities through Monte Carlo sampling, we decode them directly with a single ViT forward pass in less than 0.8\,ms (s. \Cref{tab:exp:runtime}) from the latent representation $\indplatent$, enabling fast uncertainty estimation, uncertainty-aware planning, and interactive exploration of these motion distributions.

\subsection{Interactive Exploration of Possibilities}\label{sec:met:interactive_cond}

At inference, the density decoder $\decoderdense$ provides access to the distribution $p(\trackpoint \mid \cond)$ in a single forward pass. This enables uncertainty quantification with \ours~by measuring the Shannon entropy of the predicted distribution
\begin{equation}\label{eq:met:entropy}
    H(\trackpoint) = - \sum_{(u,v)\in\mathcal{G}} p^{(u,v)}_\omega(\trackpoint | \indplatent) \log p^{(u,v)}_\omega(\trackpoint | \indplatent).
\end{equation}
Thus, we can efficiently evaluate where future motion remains under-determined: high entropy indicates multiple plausible futures, while low entropy corresponds to collapsed regions of the possibility space.
This allows users to identify and explore regions where the range of possibilities remains diverse, instead of wasting effort on collapsed low-variance regions of the kinematics distribution.

The density estimator enables interactive visualization and exploration of possible futures in real-time with no need for sampling multiple realizations from a generative model.
Users can therefore explore possible futures and iteratively modify the constraints $\cond$ in a feedback loop, enabling interactive planning by guiding and narrowing the set of feasible trajectories.
Once the user is satisfied with the result, the latents $\indplatent$ can be decoded \emph{once} into a full, coherent trajectory realization with a generative model.

\subsection{Sampling Coherent Realizations}\label{sec:met:coherent_traj}

For many downstream applications in motion planning, concrete trajectory realizations are required.
Naively, \ours\ could infer trajectories from the point-wise decoder by factorizing the distribution of scene kinematics as
\begin{equation}\label{eq:met:head_independent}
    p(\trackset \mid \cond) \approx \prod_{i} \prod_{\timestep} p(\trackpoint \mid \cond).
\end{equation}
However, if the space of possibilities is not fully collapsed and multimodal futures are possible, the point-wise decoder cannot resolve mutual dependencies between points within and between trajectories.
Thus, a sample for timestep $\tau$ might be drawn from one mode while the next timestep's sample $\tau+1$ is drawn from another, resulting in temporally incoherent trajectories.
The alternative of resolving these dependencies via auto-regressive updates of $\cond$ would entail computational overhead and early greedy commitment to sampled  points resulting in performance deterioration (\cref{tab:sup:ar_vs_ours} in supplementary materials).

Instead, we train a full decoder $\decoder$ as a joint generative model conditioned on the set of all latents, which learns to sample
\begin{equation}\label{eq:met:full_decoder}
    \trackset \sim \decoder \bigg (\big \{\indplatent \big \}_{i, \timestep} \bigg ) = \decoder(\latent) = \decoder(\encoder(\cond)) \approx p(\trackset \mid \cond)
\end{equation}
using latents produced by the frozen encoder (\cref{sec:poss_encoder}). By sampling trajectories jointly, \ours's full decoder correctly handles inter-dependencies.

By learning structured latents (\cref{eq:met:latent_factorization}) with a point-wise generative decoder described in \cref{sec:poss_encoder}, \ours~enables access to density estimates through \cref{eq:met:decoderdense} and can sample coherent trajectories from $p(\trackset \mid \cond)$ using \cref{eq:met:full_decoder}.

%% file: sec_tech/4_experiments.tex
\input{tab/4_exp/main_eval}

\section{Experiments}
\label{sec:experiments}

Our experiments evaluate \ours's capabilities in goal-conditioned motion planning and its ability to predict distributions over future kinematics. We further show our method's usefulness as a foundation for downstream tasks.

\subsection{General setup}
\label{sec:experiments:setup}

\paragraph{Data}
For training and evaluation we annotate videos with off-the-shelf trackers~\cite{zholus2025tapnexttrackingpointtap} to obtain a pseudo-ground truth for supervision and metric calculation.
We train on a dataset of 20M $224^2$\,px videos at 12 FPS that cover a wide range of subjects.
Comparisons are performed on public OpenVid-1M~\cite{nan2024openvid} while some ablation studies are performed on a held-out validation set of the training data.
We further evaluate on domain-specific data~\cite{robicquet2016sdd, pellegrini2009eth, lerner2007ucy, rt12022arxiv, walke2023bridgedata} with either human annotated trajectories or CoTracker3~\cite{karaev2024cotracker3} annotations following standard practice for application domains.

\paragraph{Implementation details} The encoder, full decoder, and density estimator of \ours~are implemented as transformer networks~\cite{dosovitskiy2020image, peebles2023scalablediffusionmodelstransformers}. We extract image features using DINOv2R-B~\cite{oquab2023dinov2, darcet2023vitneedreg} and unfreeze DINO during training.
We use 3D Axial RoPE~\cite{su2023roformerenhancedtransformerrotary}, yet mask the starting position for the full decoder to avoid information leakage.
We use $g=$20 for the density grid and append an out of bounds token.
To ensure smoothness of $\indplatent$ we apply a $\tanh$ function and add Gaussian noise with $\sigma=$1e-5 to $\indplatent$ during training~\cite{schusterbauer2026probabilistic}. The encoder adopts static camera conditioning from FPT~\cite{baumann2025whatif}.
$\decoderhead$ follows MAR~\cite{li2024autoregressivea} and is a 34M parameter MLP. More details are in \cref{sec:sup:implementation} in the appendix.

\paragraph{Training details}
The \ours~encoder $\encoder$ is trained for 450k steps with the point-wise generative decoder and a linearly increasing goal sparsity $\condtracks / \trackset$ from 0.5 to 0.01 in the first 50k steps. The density estimator $\decoderdense$ and full decoder $\decoder$ are trained for 150k and 200k steps while keeping the encoder $\encoder$ frozen. All training runs use a global batch size of 256, AdamW with a learning rate of 1e-4 (except for the density decoder with 1e-5), and bfloat16 precision. We generally model $i \in \{1 \dots 64\}$ tracks across $\timestep \in \{1 \dots 32\}$ timesteps.

\paragraph{Metrics}
For open-set evaluation, we measure motion accuracy using positional errors normalized to $[0,1]$, where 1 corresponds to the image size.
Our main metric is the \emph{endpoint error} $\mathrm{EPE} = \|\bar{t}^i_{\tau}- \trackpoint \|_2$, averaged over all trajectories $i$ and timesteps $\tau$.
We also report the \emph{percentage of correct keypoints} $\mathrm{PCK}@\alpha = \mathbb{E}[\mathrm{EPE}<\alpha]$ with $\alpha \in \{10\%,1\%\}$ relative to image size and the Final Distance Error (FDE), i.e., EPE at the final timestep.
Calibration is evaluated using the Energy Score (ES), a proper scoring rule for probabilistic predictions.
Domain-specific evaluations follow standard protocols from prior work.
Appendix \cref{sec:sup:metrics} provides more metric details.

\subsection{Open-set Motion Modeling}\label{sec:experiments:properties}

\paragraph{Motion Planning}
We compare the \ours~full decoder to open-set baselines for motion planning on the public OpenVid-1M~\cite{nan2024openvid} dataset.
Models predict future motion from a start frame and sparse future constraints (text, goal images, or positions depending on the method). Because multiple futures may satisfy the constraints, we generate five samples per model and report the best.
We evaluate both RGB video models and open-set trajectory models. For the former, we use TI2V models~\cite{yang2024cogvideox, HaCohen2024LTXVideo} conditioned on the start frame and OpenVid prompt informing the model about temporal dynamics. Motion is extracted from generated videos using TapNext~\cite{zholus2025tapnexttrackingpointtap}.
Motion-I2V~\cite{shi2024motioni2v} predicts optical flow from spatially sparse trajectories and interpolates temporally sparse signals. We generally run Motion-I2V with default settings, we compare to other configurations in Appendix~\cref{sec:sup:extended_eval}.
Track2Act~\cite{bharadhwaj2024track2act} generates trajectories on a dense grid conditioned on start and goal RGB frames.
FPT~\cite{baumann2025whatif} predicts the distributions of changes between a start and final timestep. To obtain full trajectories, we rerun FPT for multiple timesteps using only the spatio-temporally sparse conditioning points \ours~observed.

\cref{tab:exp:motion_comparison} shows \ours~achieves the best EPE, PCK, and FDE. Therefore, we are competitive with much larger and orders-of-magnitude slower video generation models.
We also outperform Track2Act, which relies on a dense final RGB frame that is typically unavailable in open-set motion planning.
Rerunning FPT for each timestep yields subpar results, highlighting the importance of modeling temporal dependencies.
Motion-I2V receives the same spatio-temporally sparse points as \ours~and additional text prompts, yet performs worse and incurs higher latency, highlighting the advantage of modeling motion only for relevant scene elements and not having to temporally interpolate conditioning.
Overall, \ours~provides state-of-the-art motion planning in open-set domains, which we attribute to explicit motion modeling of relevant scene elements and spatio-temporally sparse conditioning that provides a more precise control signal than text or RGB-based conditioning used in prior video~\cite{yang2024cogvideox, HaCohen2024LTXVideo} and motion generation~\cite{shi2024motioni2v, bharadhwaj2024track2act} methods.

\input{fig/4_exp/plot/accuracy}

\paragraph{Goal Conditioning}
\Cref{fig:exp:just_pred} shows the impact of adding more sub-goals to the conditioning set $\cond$. \ours~is able to perform more accurate trajectory planning when using very sparse information, i.e.~limited future positional information.
\ours, given only $| \condtracks | = 4$, matches Motion-I2V using $| \condtracks | = 16$ in EPE. Further, our approach is competitive with FPT~\cite{baumann2025whatif} using only image conditioning and outperforms FPT consistently when using more than 2 goal positions.

\input{fig/4_exp/qualitative/qual_plus_dens_merged}
\input{fig/4_exp/qualitative/more_cond_more_cert}

\input{tab/4_exp/robotics}

\paragraph{Qualitative Evaluation}
We present qualitative samples from \ours~with only $\condtracks / \trackset = 1\%$ known trajectory points on the left side of \Cref{fig:exp:qual_merged}. \ours~is able to predict complex trajectories for a range of objects and activities given minimal future knowledge. \cref{fig:sup:qual_comparison} shows qualitative comparisons to baselines in the Supplementary Materials.
The right side of \Cref{fig:exp:qual_merged} shows that $\indplatent$ can encode multiple possible futures given the same constraints $\cond$. We show the zoomed-in densities obtained from the \ours~density estimator $\decoderdense$ as well as two samples from $\decoder$ for the same latent representation, highlighting \ours's ability to capture the distribution of future outcomes.

\Cref{fig:exp:more_cond_more_certv2} demonstrates qualitatively that the estimated densities reflect \ours's improved motion understanding with more conditioning. While the model predicts a mostly static scene without future positions, and fails to account for object inter-dependencies with only one conditional position, it is able to infer complex motion distributions with as few as two conditioning points. 

\input{fig/4_exp/plot/calibration}

\paragraph{Distributions} 
We aim to verify our encoded latent captures meaningful uncertainty about scene dynamics.
\Cref{fig:exp:calibration:ES} shows the energy score of our point-wise and full decoders. Compared to Motion-I2V~\cite{shi2024motioni2v} and FPT~\cite{baumann2025whatif} our approach achieves superior energy score.
Evaluating the density estimator $\decoderdense$, we further compute a discretized energy score (Appendix \cref{sec:sup:metrics}) and find that the density estimator achieves superior discrete energy scores while speeding up computation by two orders of magnitude.
Additionally, \cref{fig:exp:calibration:tightening} shows that providing more conditioning tightens the distributions.
We provide further evaluation of calibration in \cref{sec:sup:extended_eval} in the supplementary materials.

\paragraph{Automated interactive conditioning}
We consider how we should perform interventions to collapse the distribution to the ground truth with minimal outside information (e.g. by prompting the user to provide further input).
Using the \ours~density estimator we can quantify the model's uncertainty in each track position using entropy (\cref{eq:met:entropy}).
We propose using entropy as an indicator for where to provide conditioning by finding the most uncertain trackpoint given the current constraints.
\cref{fig:exp:policy_comparison} compares our entropy-based conditioning to an oracle-based variant that has access to the current $\mathrm{EPE}$ for each trackpoint and a random baseline. For automated evaluation, we insert ground-truth conditioning at either the most uncertain, the highest-error, or a random trackpoint.
While using the error oracle performs even worse than random selection, we find that interactive entropy-based conditioning performs strongly.
The entropy-based conditioning matches oracle-based conditioning with only half the conditional information.
Note that no oracle is available in deployment while entropy-based conditioning selection does not rely on ground truth information.
We additionally evaluate the robustness of entropy-based conditioning w.r.t. noisy user input in Appedix \cref{sec:sup:extended_ablations}.

\subsection{Application Domains}

\input{tab/4_exp/pedestrian_merged}

\paragraph{Robotics}
Planning is commonly used to evaluate motion models~\cite{bharadhwaj2024track2act, wen2024anypoint, pandey2024motionmodeshappennext,stracke2026zipmo}. We compare \ours~following the protocol from Track2Act~\cite{bharadhwaj2024track2act} and sample trajectories for a robotic arm given start and goal states.
While Track2Act uses dense RGB for start and end states, we use readily available positional information.
We follow Track2Act and report \emph{area under the curve} $\Delta= \frac{1}{A} \sum^{A}_{a = 1} PCK@\frac{a}{S}$, where $A=10$ is the highest PCK threshold in pixel units and $S=256$ is the pixel resolution (more details: Appendix \cref{sec:sup:metrics}).
\cref{tab:exp:robotics} highlights \ours~is competitive with Track2Act without training on robotics data. In comparison, Track2Act trains on splits from RT1~\cite{rt12022arxiv} and BridgeData~\cite{walke2023bridgedata}.
This highlights fundamental understanding of motion and kinematics constraints learned by our model.

\paragraph{Pedestrian Trajectory Prediction}
Pedestrian trajectory prediction important for autonomous driving and a long-standing benchmark.
We compare zero-shot and finetuned performance of \ours. As we require RGB input, we use only a subset of ETH/UCY. Further details are provided in Appendix \cref{sec:sup:extended_eval}.
\cref{tab:exp:sdd_eth_combined} shows our zero-shot performance is competitive with established methods such as Social-LSTM~\cite{alahi2016social} even though we \emph{do not train for prediction}. With minimal finetuning ($<$15k iterations) \ours~achieves performance comparable to Social-GAN~\cite{gupta2018social}.
While we are outperformed by more recent domain-specific methods, our results underscore the utility of our representations.

\subsection{Ablations}
\label{sec:experiments:ablations}

\input{tab/4_exp/ablation/enc_ablations}

\paragraph{Entangled and Global Representations}
The left side of \cref{tab:exp:enc_ablation} shows training the point-wise decoder with a merged global token or without enforcing structure yields substantially degraded performance and does not allow decoding into accurate points.
Intuitively, factorized $\indplatent$ encode $p(\trackpoint \mid \cond)$ and can be seamlessly decoded by the point-wise decoder and density estimator. In comparison, for non-localized embeddings the decoder has to identify relevant information by itself, resulting in a substantially harder task for $\decoderhead$ and $\decoderdense$.
In comparison, our disentangled $\indplatent$ enable high-quality track predictions and estimation of point-wise densities.

\paragraph{Latent Dimensionality}
\cref{tab:exp:enc_ablation} further shows the performance on in-distribution samples ($\condtracks/\trackset = 1\%$ conditioning) is almost independent of latent dimensionality. However, in OOD settings, e.g.~when substantially more conditioning ($\condtracks/\trackset = 90\%$) is given, smaller latents outperform larger variants, potentially due to over-adaptation. Unless noted otherwise, we use a latent dimensionality of $64$ as a balance with strong performance on both in-domain and OOD tasks.

\input{fig/4_exp/plot/training_objectives}

\paragraph{Pretraining with other Decoders}
\cref{fig:sup:comp_training} compares pre-training the encoder $\encoder$ with the point-wise decoder $\decoderhead$, full decoder $\decoder$, or density decoder $\decoderdense$.
\cref{fig:sup:comp_training:planning} shows motion planning performance is similar between directly training with the full decoder or first pretraining with a point-wise decoder. However, in \cref{fig:sup:comp_training:des} the density estimator fails to produce high-quality uncertainty estimates when the encoder is pre-trained with the full decoder.
Also, pretraining the encoder with the density estimator and a grid Cross-Entropy objective from~\cref{eq:met:grid_ce} leads to inferior performance of the full and density decoder.
We conclude that (1) the density decoder's discretization limits the encoder's learning signal and (2) the density decoder $\decoderdense$ struggles to interpret entangled latents learned by pretraining the encoder with the full decoder $\decoder$ similar to \cref{tab:exp:enc_ablation}.
Thus, pretraining the encoder with our point-wise decoder $\decoderhead$ preserves both strong full trajectory sampling with a second stage decoder $\decoder$ and interpretability of $\indplatent$ for the density decoder $\decoderdense$.

\input{tab/4_exp/ablation/latency_pw_v_full_merged}

\paragraph{Full vs.\ Point-wise Decoder}
Our point-wise decoder $\decoderhead$ is able to infer the positions of scene elements at a specific timestep given a latent $\indplatent$.
Full trajectories can be reconstructed from independent point-wise estimates. However, in case of multi-modal uncertainty, independent point-wise estimation breaks down as samples per timestep can be assigned to different modes. The full decoder $\decoder$ resolves these inter-dependencies.
\cref{tab:exp:full_v_pointwise} verifies this assumption: in uncertain cases with limited $\cond$ the full decoder produces higher quality estimates, while the point-wise model performs competitively when the distribution is collapsed.

\input{fig/4_exp/plot/ablation_grid_size}

\paragraph{Grid Resolution}
Intuitively, higher $g$ balance fine-granular density estimates with compute cost.
\cref{fig:exp:ablation_grid_size} shows this trade-off empirically. As $g$ influences the value range of the discrete energy score, we keep $g = 20$ fixed for all other evaluations and additionally report results with normalized grids ($g=40$) for the ablation.

\paragraph{Alignment of Direct and MC Densities}
The density decoder $\decoderdense$, point-wise decoder $\decoderhead$, and full decoder $\decoder$ decode the same latents. Thus, heatmaps from $\decoderdense$ should align with MC density estimates from $\decoder$ and $\decoderhead$.
\Cref{fig:exp:alignment} confirms this: both decoders achieve lower Jensen--Shannon divergence and Wasserstein distance to $\decoderdense$ than 1-hot ground-truth heatmaps or baseline samples.

\paragraph{Component Latency} 
\cref{tab:exp:runtime} shows runtimes for our individual components. The density estimator is able to decode a non-parametric distribution faster than either the point-wise or full decoder can produce a single example.

%% file: tab/4_exp/main_eval.tex
\begin{table}[b]
    \centering
    \newcommand{\inte}{\color{ourgray}}
    \newcommand{\vcat}[1]{\rotatebox{90}{\scriptsize#1}}
    {
    \adjustbox{max width=\linewidth}{
    \footnotesize
    \begin{tabular}{c l c c c c c c c}
        \toprule
        \multicolumn{2}{c}{\textbf{Method}}
        & \makecell{\textbf{Text}\\\textbf{Free}}
        & \makecell{\textbf{Num.}\\\textbf{Goals}}
        & $\mathbf{\mathrm{\mathbf{EPE}}\downarrow}$
        & \makecell{$\mathrm{\mathbf{PCK}}\uparrow$\\@$\mathbf{10\%}$}
        & \makecell{$\mathrm{\mathbf{PCK}}\uparrow$\\@$\mathbf{1\%}$}
        & $\mathbf{\mathrm{\mathbf{FDE}}\downarrow}$
        & \makecell{\textbf{Latency}$\downarrow$\\\textbf{[s]}}\\
        \midrule

        \multirow{2}{*}{\vcat{video}} 
        & CogVideoX~\cite{yang2024cogvideox} & \no & \notavail 
        & 0.027 & 0.952 & 0.466 & 0.032 & 178 \\
        & LTX~\cite{HaCohen2024LTXVideo}     & \no & \notavail 
        & 0.025 & 0.960 & 0.464 & 0.033 & 39 \\
        \midrule

        \multirow{7}{*}{\vcat{explicit motion}}
        & \multirow{2}{*}{Motion-I2V~\cite{shi2024motioni2v}} 
            & \multirow{2}{*}{\no} 
            & 4  
            & 0.055 & 0.864 & 0.150 & 0.060 & \multirow{2}{*}{13.2} \\
        &   &   & 16 
            & 0.033 & 0.957 & 0.263 & 0.037 &  \\
        & Track2Act~\cite{bharadhwaj2024track2act} 
            & \yes & \notavail 
            & 0.041 & 0.931 & 0.091 & 0.046 & 4.10 \\

        & \multirow{2}{*}{FPT~\cite{baumann2025whatif}} 
            & \multirow{2}{*}{\yes} 
            & 4  
            & 0.029 & 0.936 & 0.493 & 0.035 & \multirow{2}{*}{0.60} \\
        &   &   & 16 
            & 0.026 & 0.943 & 0.523 & 0.032 &  \\
        \cmidrule(l{1pt}r{4pt}){2-9} %
        & \multirow{2}{*}{\oursours} 
            & \multirow{2}{*}{\yes} 
            & 4  
            & 0.020 & 0.973 & 0.544 & 0.026 & \multirow{2}{*}{\textbf{0.40}} \\
        &   &   & 16 
            & \textbf{0.013} & \textbf{0.989} & \textbf{0.666} & \textbf{0.017} &  \\
        \bottomrule
    \end{tabular}
    }}
    \caption{\textbf{Comparison on Motion Planning.} \ours~achieves strong motion planning performance in open-set domains with reduced latency by modeling motion only for a subset of all scene elements and using spatio-temporally sparse goal conditioning.}\label{tab:exp:motion_comparison}
\end{table}

%% file: fig/4_exp/plot/accuracy.tex
\begin{figure}[t]
    \centering
    \begin{subfigure}{0.3\linewidth}
        \includegraphics[width=\linewidth]{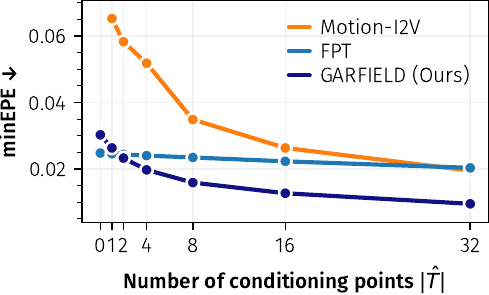}
        \caption{Average over time.}
    \end{subfigure}
    \hfill
    \begin{subfigure}{0.3\linewidth}
        \includegraphics[width=\linewidth]{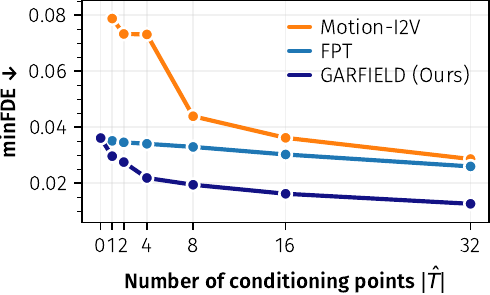}
        \caption{Final positional error.}
    \end{subfigure}
    \hfill
    \begin{subfigure}{0.3\linewidth}
        \includegraphics[width=\linewidth]{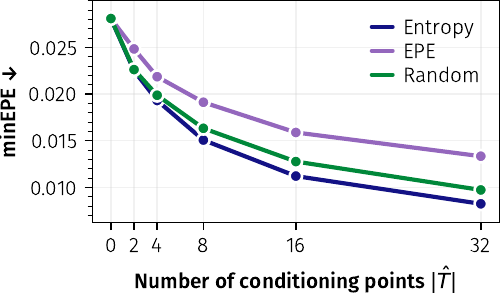}
        \caption{Conditioning policy.}\label{fig:exp:policy_comparison}
    \end{subfigure}
    \caption{\textbf{Impact of Goal Conditioning.} \ours~is able to infer accurate motion in terms of \textbf{(a)} minEPE and \textbf{(b)} minFDE with limited conditioning. \textbf{(c)} Selecting conditioning based on entropy yields faster collapse than an oracle or random selection.}
    \label{fig:exp:just_pred}
\end{figure}

%% file: fig/4_exp/qualitative/qual_plus_dens_merged.tex
\begin{figure}[t]
  \centering
  \setlength{\tabcolsep}{4pt}
  \renewcommand{\arraystretch}{1.0}

  \newcommand{\Lsamplewidth}{0.105\textwidth} %
  \newcommand{\Limg}[2]{\includegraphics[width=\Lsamplewidth]{fig/4_exp/qualitative/#1/#1_t#2.pdf}}
  \newcommand{\Limgrow}[1]{\Limg{#1}{8} & \Limg{#1}{16} & \Limg{#1}{24} & \Limg{#1}{30}}

  \newcommand{\Rsmall}{0.075\textwidth}
  \newcommand{\Rlarge}{0.155\textwidth}
  \newcommand{\Rimg}[1]{\includegraphics[width=\Rsmall]{fig/4_exp/qualitative/#1.pdf}}
  \newcommand{\Rimgvert}[1]{\includegraphics[width=0.045\textwidth]{fig/4_exp/qualitative/#1.pdf}}
  \newcommand{\Rimglarge}[1]{\includegraphics[width=\Rlarge]{fig/4_exp/qualitative/#1.pdf}}

  \begin{tabular}{@{}c@{\hspace{10pt}}c@{}}

    \begin{tabular}{cccc}
      \multicolumn{1}{c}{\textbf{$\timestep \leq 8$}} &
      \multicolumn{1}{c}{\textbf{$\timestep \leq 16$}} &
      \multicolumn{1}{c}{\textbf{$\timestep \leq 24$}} &
      \multicolumn{1}{c}{\textbf{$\timestep \leq 32$}} \\
      \Limgrow{bmx} \\
      \Limgrow{gardening} \\
      \Limgrow{walking}
    \end{tabular}
    &
    \begin{tabular}{@{}ccccc@{}}
      \multicolumn{2}{c}{\Rimglarge{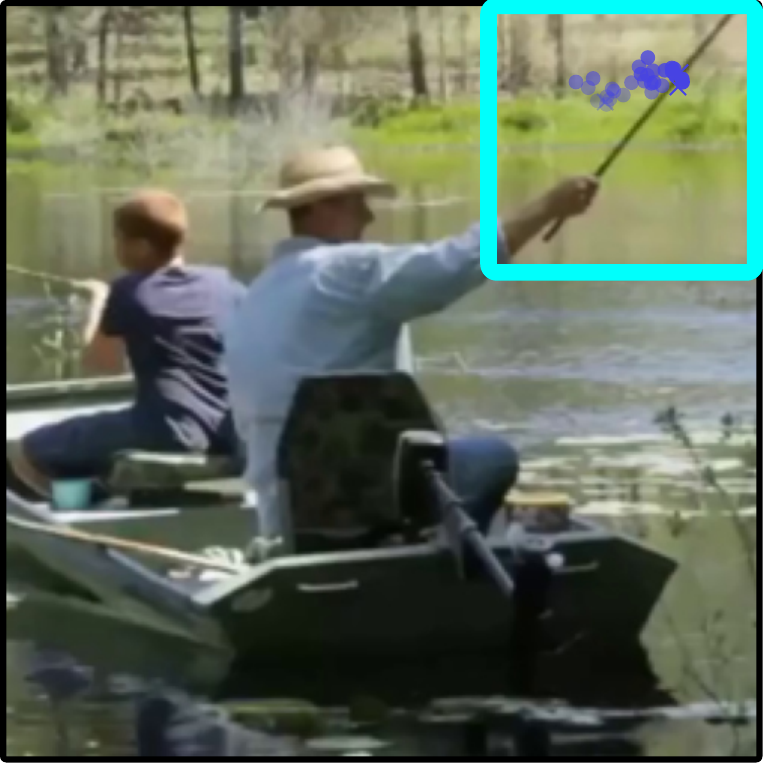}} &
      \multicolumn{2}{c}{\Rimglarge{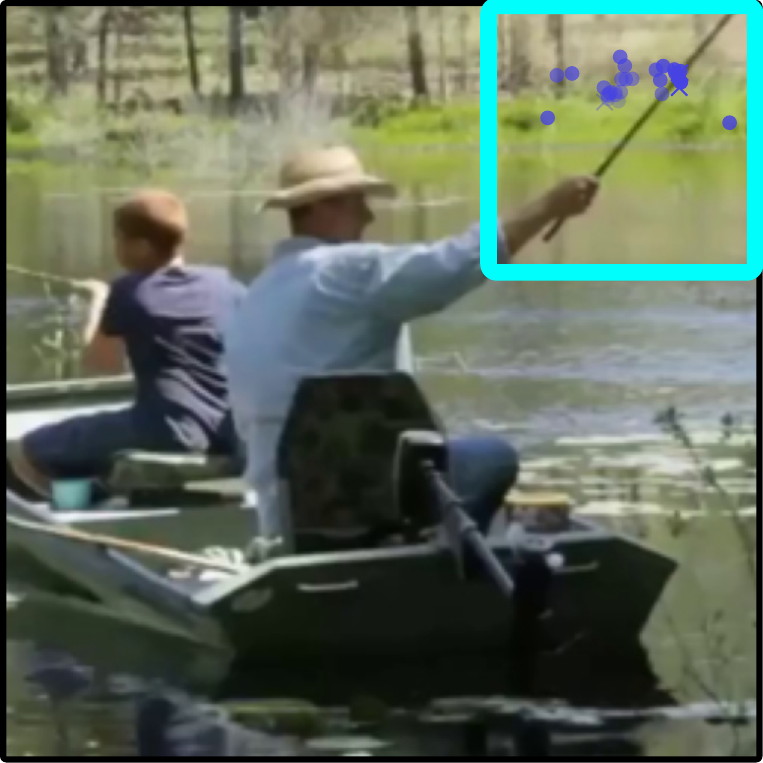}} & \multirow{3}{*}{\Rimgvert{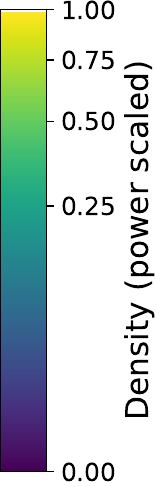}} \\
      $\timestep{=}8$ & $\timestep{=}16$ & $\timestep{=}24$ & $\timestep{=}32$ & \\
      \Rimg{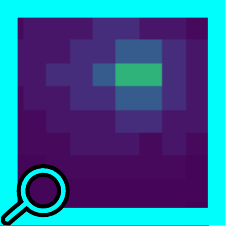} &
      \Rimg{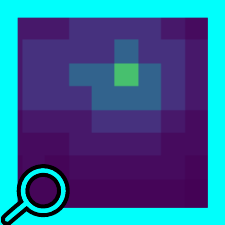} &
      \Rimg{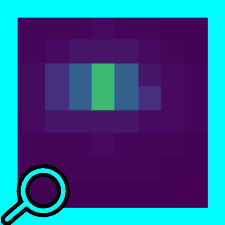} &
      \Rimg{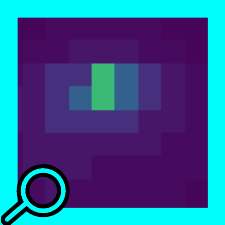} \\
    \end{tabular}

  \end{tabular}
  \caption{\textbf{Qualitative open-set motion and uncertainty.} (\textbf{left}) With sparse conditioning ($\condtracks / \trackset=1\%$), \ours~samples realistic motion. Dots denote predictions and crosses conditioning points. (\textbf{right}) Complex motion gives non-collapsed distributions.}
  \label{fig:exp:qual_merged}
\end{figure}

%% file: fig/4_exp/qualitative/more_cond_more_cert.tex
\begin{figure}[t]
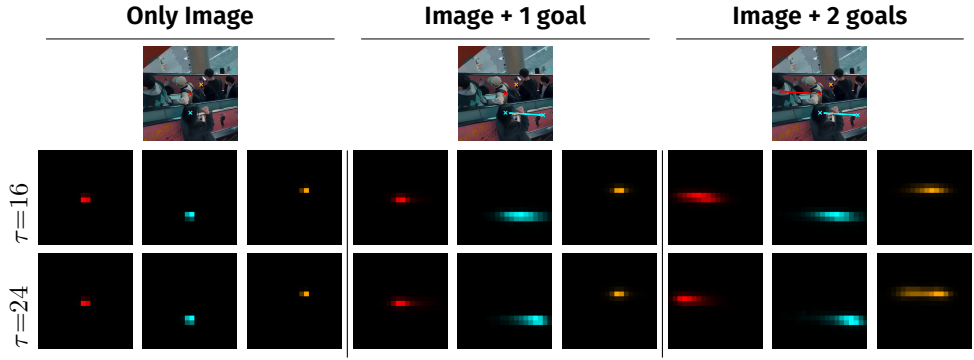

  \centering
  \setlength{\tabcolsep}{2pt}
  \renewcommand{\arraystretch}{1.0}

  \newcommand{\samplewidth}{0.075\textwidth}
  \newcommand{\img}[2]{\includegraphics[width=\samplewidth]{fig/4_exp/qualitative/more_cond_more_cert/#1/#2.pdf}}

  \newcommand{\samplesAtH}[2]{%
    \img{#1}{n0_t#2} & \img{#1}{n1_t#2} & \img{#1}{n2_t#2}%
  }

  \newcommand{\Hlabel}[1]{\rotatebox{90}{\textbf{$\timestep{=}#1$}}}

  \begin{tabular}{c ccc | ccc | ccc}
    & \multicolumn{3}{c}{\textbf{Only Image}} & \multicolumn{3}{c}{\textbf{Image + 1 goal}} & \multicolumn{3}{c}{\textbf{Image + 2 goals}} \\
    \cmidrule(lr){2-4}\cmidrule(lr){5-7}\cmidrule(lr){8-10}

    &
      \multicolumn{3}{c}{\img{0cond}{cond_frame}} &
      \multicolumn{3}{c}{\img{1cond}{cond_frame}} &
      \multicolumn{3}{c}{\img{2cond}{cond_frame}} \\

    \Hlabel{16} &
      \samplesAtH{0cond}{16} &
      \samplesAtH{1cond}{16} &
      \samplesAtH{2cond}{16} \\

    \Hlabel{24} &
      \samplesAtH{0cond}{24} &
      \samplesAtH{1cond}{24} &
      \samplesAtH{2cond}{24} \\

  \end{tabular}
  \caption{\textbf{Distributions with minimal conditioning.} Given $| \condtracks |$=0, \ours~assumes a static scene. With $| \condtracks |$=1 the model captures motion but produces inaccurate object relations. With $| \condtracks |$=2, \ours~remains uncertain only about exact speed.}
  \label{fig:exp:more_cond_more_certv2}
\end{figure}

%% file: tab/4_exp/robotics.tex
\begin{table}[b]
    \centering
    {\footnotesize
    \begin{tabular}{lcc}
    \toprule
         \multicolumn{1}{c}{\textbf{Method}} & \textbf{BridgeData} ~\cite{walke2023bridgedata} $\uparrow$ & \textbf{RT1}~\cite{rt12022arxiv}$\uparrow$ \\
         \midrule
         Flow~\cite{haofei2023unifyingflowstereodepth} & 0.32 & 0.38 \\
         Track2Act~\cite{bharadhwaj2024track2act} & \textbf{0.77} & 0.75 \\
        \arrayrulecolor{gray!50!white}
        \midrule
        \arrayrulecolor{black}
         \oursours {\tiny 0-shot} & 0.76 & \textbf{0.77} \\
         \bottomrule
    \end{tabular}}
    \caption{\textbf{Robotic Planning.} Given final goals \ours~can infer robot arm motion with similar quality to prior art without finetuning. We extend~\cite{bharadhwaj2024track2act}'s table.}\label{tab:exp:robotics}
\end{table}

%% file: fig/4_exp/plot/calibration.tex
\begin{figure}[htb]
    \centering
    \begin{subfigure}{0.3\linewidth}
        \includegraphics[width=\linewidth]{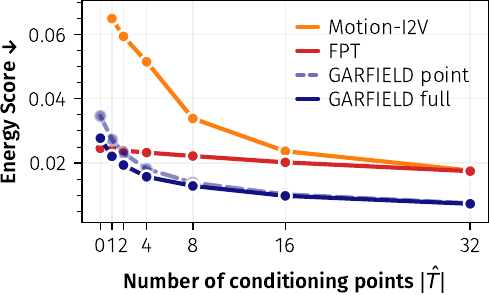}
        \caption{Energy Score.}\label{fig:exp:calibration:ES}
    \end{subfigure}
    \hfill
    \begin{subfigure}{0.3\linewidth}
        \includegraphics[width=\linewidth]{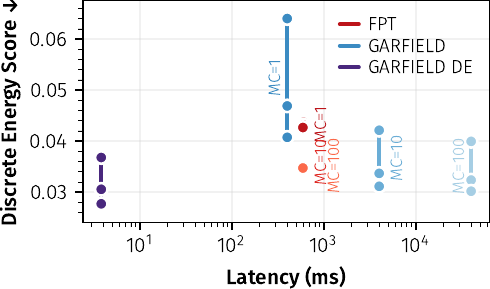}
        \caption{Discretized energy scores.}\label{fig:exp:energy_v_latency}
    \end{subfigure}
    \hfill
    \begin{subfigure}{0.3\linewidth}
        \includegraphics[width=\linewidth]{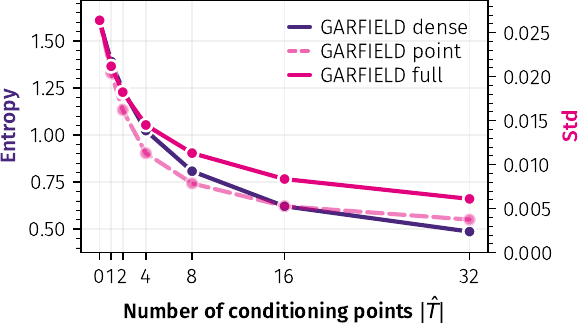}
        \caption{Distribution tightening.}\label{fig:exp:calibration:tightening}
    \end{subfigure}
    \caption{\textbf{Evaluation of Distributions.} \ours achieves the best energy score. The density decoder $\decoderdense$ attains superior discretized energy scores with orders-of-magnitude lower latency. Additional conditioning reduces entropy and variance.}
    \label{fig:exp:calibration}
\end{figure}

%% file: tab/4_exp/pedestrian_merged.tex
\begin{table*}[b]
    \centering
    {\footnotesize
    \setlength{\tabcolsep}{5pt}
    \begin{tabular}{l cc cc cc}
        \toprule
        \multicolumn{1}{c}{\textbf{Method}} 
        & \multicolumn{2}{c}{\textbf{ETH}~\cite{pellegrini2009eth}} 
        & \multicolumn{2}{c}{\textbf{HOTEL}~\cite{pellegrini2009eth}} 
        & \multicolumn{2}{c}{\textbf{SDD}~\cite{robicquet2016sdd}} \\
        \cmidrule(lr){2-3}\cmidrule(lr){4-5}\cmidrule(lr){6-7}
        & \textbf{ADE} $\downarrow$ & \textbf{FDE} $\downarrow$ & \textbf{ADE} $\downarrow$ & \textbf{FDE} $\downarrow$ & \textbf{ADE} $\downarrow$ & \textbf{FDE} $\downarrow$ \\
        \midrule
        Social-LSTM~\cite{alahi2016social}      & 1.09 & 2.35 & 0.79 & 1.76 & 57.00 & 31.20 \\
        Social-GAN~\cite{gupta2018social}       & 0.81 & 1.52 & 0.72 & 1.61 & 27.23 & 41.44 \\
        Trajectron++~\cite{salzmann2021trajectron}     & 0.39 & 0.83 & 0.12 & 0.21 & 8.98  & 19.02 \\
        IDM~\cite{liu2024intentionaware}              & 0.41 & 0.62 & 0.15 & 0.25 & 7.46 & 13.83 \\
        \arrayrulecolor{gray!50!white}
        \midrule
        \arrayrulecolor{black}
        \oursours {\ \tiny 0-shot} & 1.09 & 1.79 & 0.41 & 0.71 & 36.77 & 49.86 \\
        \oursours {\ \tiny FT}     & 0.84 & 1.44 & 0.29 & 0.56 & 19.92 & 34.98 \\
        \bottomrule
    \end{tabular}
    }
    \caption{\textbf{Pedestrian trajectory prediction.} Errors are reported in pixels w.r.t. the original video resolution for Stanford Drone and meters for ETH/UCY.
    Baseline numbers are taken from IDM~\cite{liu2024intentionaware}. \ours~performs competitively.}
    \label{tab:exp:sdd_eth_combined}
\end{table*}

%% file: tab/4_exp/ablation/enc_ablations.tex
\begin{table}[t]
    \centering
    \newcommand{\inte}{\color{ourgray}}
    \newcommand{\vcat}[1]{\rotatebox{90}{\scriptsize#1}}{
    \adjustbox{max width=0.4\linewidth}{
    \begin{tabular}{l c c}
    \toprule
    \multicolumn{1}{c}{\makecell{\textbf{Latent}\\\textbf{structure}}} & \textbf{EPE $\downarrow$} & \textbf{FDE $\downarrow$} \\
    \midrule
    Global & 0.479 & 0.482 \\
    Entangled       & 0.509 & 0.510 \\
    Ours        & 0.011 & 0.015 \\
    \bottomrule
    \end{tabular}
    }}
    {
    \adjustbox{max width=0.54\linewidth}{
    \begin{tabular}{c c c }
    \toprule
    \makecell{\textbf{Latent}\\\textbf{size}} & \makecell{\textbf{EPE} $\downarrow$\\\textbf{$\frac{\condtracks}{\trackset} = 1\%$}} & \makecell{\textbf{EPE (OOD)} $\downarrow$\\\textbf{$\frac{\condtracks}{\trackset} = 90\%$}} \\
    \midrule
    16 & 0.012 & 0.009 \\
    64 & 0.012 & 0.008 \\
    256 & 0.012 & 0.019 \\
    \bottomrule
    \end{tabular}
    }}
    \caption{\textbf{Embeddings.} Using non-disentangled representations yields performance degradation. Smaller embeddings achieve superior performance in OOD settings. Evaluated on held-out validation set of training data.}
    \label{tab:exp:enc_ablation}
\end{table}

%% file: fig/4_exp/plot/training_objectives.tex
\begin{figure}[b]
    \centering
    \begin{subfigure}{0.4\textwidth}
        \includegraphics[width=\textwidth]{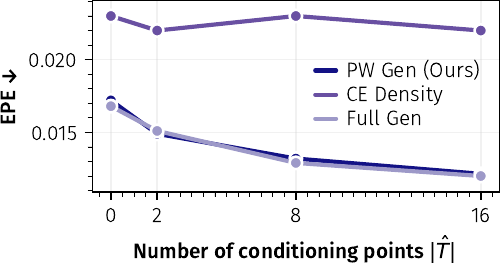}
        \caption{Full Decoder}\label{fig:sup:comp_training:planning}
    \end{subfigure}
    \hfill
    \begin{subfigure}{0.4\textwidth}
        \includegraphics[width=\textwidth]{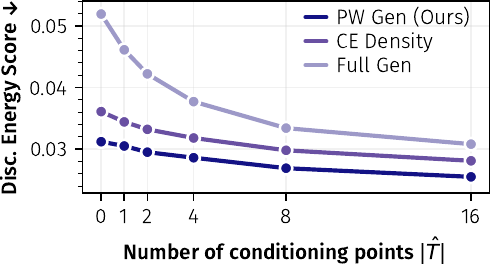}
        \caption{Density Estimator}\label{fig:sup:comp_training:des}
    \end{subfigure}
    \caption{\textbf{Comparison of Training Protocol.} We compare our pretraining with a point-wise generative decoder $\decoderhead$ (\textit{PW Gen}) to training the encoder with the full decoder $\decoder$ (\textit{Full Gen}) or the deterministic density estimator $\decoderdense$ (\textit{CE Density}). The former achieves comparable track prediction performance but fails for density decoding, while the latter fails to produce a representation that can be used for motion planning. Evalauted on held-out validation set of training data.}
    \label{fig:sup:comp_training}
\end{figure}

%% file: tab/4_exp/ablation/latency_pw_v_full_merged.tex
\begin{table}[t]
\centering

\begin{minipage}[t]{0.48\linewidth}
\centering
\begin{tabular}{lccc}
\toprule
\multicolumn{1}{c}{\textbf{Decoder}} &
\makecell{$|\condtracks|$} &
\makecell{\textbf{EPE} $\downarrow$} &
\makecell{\textbf{FDE} $\downarrow$} \\
\midrule
\multirow{3}{*}{\makecell{Point-wise\\$\decoderhead$}}
 & 2  & 0.035 & 0.060 \\
 & 4  & 0.026 & 0.047 \\
 & 16 & 0.014 & 0.024 \\
\cmidrule{1-4}
\multirow{3}{*}{\makecell{Full sample\\$\decoder$}}
 & 2  & 0.023 & 0.029 \\
 & 4  & 0.019 & 0.024 \\
 & 16 & 0.013 & 0.017 \\
\bottomrule
\end{tabular}
\captionof{table}{\textbf{Point-wise vs.\ Full Decoder.} Given limited constraints the full decoder outperforms the point-wise model.}
\label{tab:exp:full_v_pointwise}
\end{minipage}
\hfill
\begin{minipage}[t]{0.48\linewidth}
\centering
\begin{tabular}{l c}
\toprule
\multicolumn{1}{c}{\textbf{Component}} & \textbf{Latency [ms]} \\
\midrule
Encoder $\encoder$ & 3.0 \\
Density $\decoderdense$ & 0.8 \\
\midrule
Point-wise $\decoderhead$ & 37.0 \\
Full $\decoder$ & 330.4 \\
\bottomrule
\end{tabular}
\captionof{table}{\textbf{Latency.} The density decoder predicts heatmaps faster than either generative decoder produces a sample.}
\label{tab:exp:runtime}
\end{minipage}

\end{table}

%% file: fig/4_exp/plot/ablation_grid_size.tex
\begin{figure}[b]
    \centering
    \begin{minipage}[t]{0.32\linewidth}
        \centering
        \includegraphics[width=\linewidth]{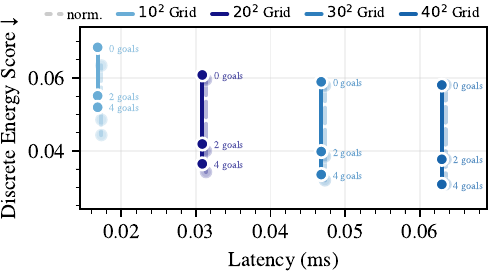}
        \caption{\textbf{Grid size ablation.} Grid size $g$ trades quality for latency. Evaluated on held-out validation split of training data.}
        \label{fig:exp:ablation_grid_size}
    \end{minipage}
    \hfill
    \begin{minipage}[t]{0.65\linewidth}
        \centering
        \includegraphics[width=\linewidth]{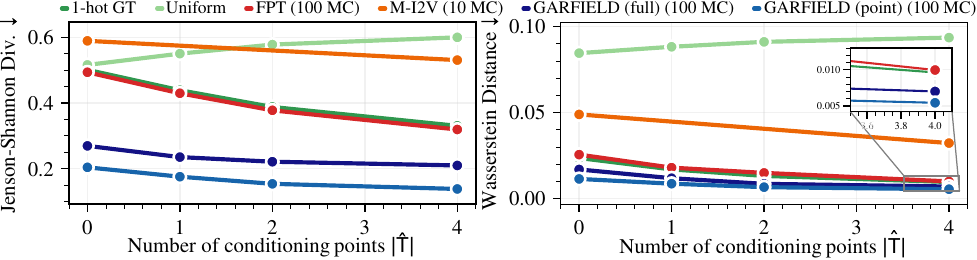}
        \caption{\textbf{Alignment of MC samples and densities.} (\textbf{left}) MC samples from $\decoder$ and $\decoderhead$ best match $\decoderdense$ in Jensen-Shannon divergence. (\textbf{right}) Their MC heatmaps also achieve the lowest Wasserstein Distance to $\decoderdense$.}
        \label{fig:exp:alignment}
    \end{minipage}
\end{figure}

%% file: sec_tech/5_conclusion.tex
\section{Conclusion}\label{sec:conclusion}

Reasoning about partially observable and uncertain environments requires modeling not only what will happen, but what \emph{could happen}.
We presented \ours, a model that represents the space of possible scene kinematics through a structured probabilistic latent representation learned with a joint motion encoder and a point-wise diffusion decoder.

Conditioned on sparse goal positions, this representation captures the distribution of plausible future motions and progressively sharpens as additional constraints become available.
From the same latent representation, we obtain both coherent trajectory samples and probability densities over future positions through an efficient deterministic density decoder.
These densities provide direct access to uncertainty about where scene elements may appear, thus letting users identify underdetermined future regions and refine them through sparse conditioning.
This enables interactive exploration and motion planning with minimal user intervention.

\ours~learns an open-set motion representation that performs competitively with large video models at substantially lower latency.
The learned distributions collapse as additional information is given and transfer to downstream tasks such as robotic planning and pedestrian trajectory prediction.
More broadly, we view \ours~as a step toward models that reason about the space of possible futures rather than single outcomes, enabling interpretable, controllable, and efficient probabilistic modeling of scene evolution.

%% file: sec_tech/X_acknowledgment.tex
\section*{Acknowledgements}

This project has been supported by the Horizon Europe project ELLIOT (GA No.\ 101214398), the project ``GeniusRobot'' (01IS24083) funded by the Federal Ministry of Research, Technology and Space (BMFTR), the BMWE ZIM-project (No.\ KK5785001LO4) ``conIDitional LoRA'', the German Federal Ministry for Economic Affairs and Energy within the project ``NXT GEN AI METHODS - Generative Methoden für Perzeption, Prädiktion und Planung'', and the bidt project KLIMA-MEMES. The authors gratefully acknowledge the Gauss Center for Supercomputing for providing compute through the NIC on JUWELS/JUPITER at JSC and the HPC resources supplied by the NHR@FAU Erlangen.
We thank Stefan Baumann, Johannes Schusterbauer, Ming Gui, and Ulrich Prestel for helpful discussions and feedback, Nick Stracke, Frank Fundel, and Thomas Ressler-Antal for initial data work, Stefan Baumann for creating the \LaTeX~template, and Owen Vincent for continuous technical support.

%% file: sec_tech/X_suppl.tex
\clearpage
\ifeccv
    \setcounter{page}{1}
    \crefalias{section}{appsec}
    \crefalias{figure}{appfig}
    \crefalias{table}{apptab}
    \crefalias{equation}{appeq}
\fi

\appendix
\setcounter{figure}{0}
\setcounter{table}{0}
\setcounter{equation}{0}
\setcounter{section}{0}
\renewcommand\thesection{\Alph{section}}
\renewcommand\thefigure{\Alph{section}.\arabic{figure}}
\renewcommand\thetable{\Alph{section}.\arabic{table}}
\renewcommand\theequation{\Alph{section}.\arabic{equation}}

\section{Implementation Details}\label{sec:sup:implementation}

\input{tab/X_sup/hyperparam}

\paragraph{Hyperparameters}
The hyperparameters for all \ours~models used in evaluations and for qualitative results are summarized in \cref{tab:supp:hyperparam}.
We train the encoder model with the pointwise decoder for a total of 450k steps using the AdamW~\cite{loshchilov2019decoupled} optimizer with a linear learning rate warm-up of 300 steps and a final constant learning rate of $0.0001$. We use a global batch size of 256 for all training runs. During encoder training, we start with $\condtracks / \trackset = 50\%$ known positions and linearly decrease the amount of future knowledge to $\condtracks / \trackset = 1\%$ over the first 50k training iterations. We find that this curriculum w.r.t.\ conditional sparsity results in better adherence to conditionings, while starting from more than $\condtracks / \trackset=50\%$ known positions results in training instabilities.
The full decoder and density estimator are trained with a frozen encoder model using the same batch size, optimizer, and learning rate.
We keep conditioning sparsity constant for decoder training and train these models for 150k (density decoder) and 200k (full decoder) iterations.
All training is performed with $\mathrm{bfloat16}$ precision and parallelized across 16 Nvidia H200 GPUs for efficiency.

Starting from 20M video clips with high variety, we apply TapNext~\cite{zholus2025tapnexttrackingpointtap} to obtain our training data.
The tracker is seeded with random space-time positions.
During training, we apply a visibility filter to ensure all points our model observes are visible in the first frame $I$.
We further utilize the tracker's predicted certainty to ensure high-quality tracks for training.
We normalize positions w.r.t.\ the image size, such that visible tracks are in $[0,1]$ range.
Start frames are resized to $224 \times 224$ resolution before obtaining image features from the pre-trained foundational image feature extractor~\cite{simeoni2025dinov3}, which is finetuned during encoder training.

For tokenization, known conditional track positions $\condtracks$ are encoded as Fourier features, while unknown track positions are replaced with a learnable query token. For known positions, we concatenate image features from the corresponding position. Further, the encoder performs cross-attention to all image features. The resulting token sequence is passed to the encoder's transformer which is 24 layers deep and has a width of 1024. The encoder produces latents for all known and unknown positions. While we supervise all positions, we neglect conditional positions for evaluation. Encoder transformer tokens are finally downprojected to a fixed latent dimensionality of 64, and a $\tanh$ function is applied to obtain the final latent representation. We further add a small noise perturbation during training to ensure a locally smooth latent space~\cite{schusterbauer2026probabilistic}.
We use standard Llama-style transformers with RMSNorm~\cite{zhang2019root}, SwiGLU~\cite{shazeer2020glu} activations, and RoPE~\cite{su2023roformerenhancedtransformerrotary} positional embeddings.
The point-wise decoder is a three layer MLP with a width of 1024, using GELU~\cite{hendrycks2023gaussian} activations, and LayerNorm~\cite{ba2016layer}. The point-wise decoder receives a noisy version of the sampled point as input and is conditioned on the diffusion timestep and latent via adaptive normalization~\cite{xu2019understandinga, peebles2023scalablediffusionmodelstransformers}. A final layer downprojects the hidden dimensionality to the point dimensionality.
The full decoder is implemented as a DiT~\cite{peebles2023scalablediffusionmodelstransformers} with a similar architecture to the encoder. It is conditioned on the diffusion timestep via adaptive normalization, and conditioned on the full set of encoder latents by concatenation of noisy inputs and corresponding latent representations. The transformer tokens are finally downprojected to the data dimensionality.
The density estimator is implemented as an S-sized ViT~\cite{dosovitskiy2020image} with a depth of 12 and width of 768. It is conditioned on a single latent by prepending the latent to the sequence of grid tokens. These grid tokens are initialized as Fourier embedded grid positions. We further add a learnable query token for out of bounds positions. The density decoder transformer tokens are ultimately downprojected to a dimensionality of one for each grid cell and the out of bounds position, providing cross-entropy logits.

We pre-train the encoder and pointwise decoder with a point-wise flow matching objective (\cref{eq:flow-loss}). The full decoder is trained with a flow matching objective over all track positions. The density estimator is trained with a grid Cross Entropy loss (\cref{eq:met:grid_ce}).
We do not enforce any attention masks in the encoder or decoder models and thus provide the model with the ability to leverage information about known parts of the future and track interdependencies for other track positions.

\paragraph{Conditioning}
We condition decoders on encoder latents in different ways. For the full decoder $\decoder$ each trackpoint $t^i_\tau$ corresponds to exactly one specific processed token in the DiT and latent component $\indplatent$. Hence, we enforce this structure by using \emph{in-context conditioning}, concatenating noisy DiT inputs with corresponding latents before passing tokens to DiT layers.

For the pointwise decoder MLP $\decoderhead$, the conditioning signal is much stronger and easier to interpret, because it only receives a single latent $\indplatent$ to denoise a single trackpoint $t_\tau^i$. Thus, we resort to realizing the conditioning mechanism with \emph{adaptive normalization}~\cite{xu2019understandinga, peebles2023scalablediffusionmodelstransformers}. We linearly project the latent with a learned matrix $Z^i_{\timestep} = \mathrm{proj}_Z(\indplatent)$ and add projected latents to embeddings of the diffusion timestep $\kappa$. The combined conditioning $c = \mathrm{proj}_c(\{Z^i_{\timestep}, \kappa\})$ is then used to compute scale $\gamma(c)$ and shift $\beta(c)$ with learned linear projections. We then modify the standard normalization and compute
\begin{equation}\label{eq:supp:adanorm}
    \mathrm{AdaNorm}(x, \gamma(c), \beta(c)) = \gamma(c) \mathrm{Norm}(x)+\beta(c).
\end{equation}
We apply adaptive normalization before each MLP layer, conditioning each non-linear projection on the latent.

Similarly to the pointwise model, the density estimator $\decoderdense$ processes only a single latent $\indplatent$ corresponding to a single trackpoint $t_\tau^i$. Yet, similar to the full decoder, the density estimator processes a sequence of tokens enabling stronger conditioning. However, the density estimator is not a generative diffusion model, but a deterministic network. We opt for a variant of \emph{in-context conditioning} where the latent $\indplatent$ is projected to the model dimensionality $Z_\tau^i = \mathrm{proj}_{dens}(\indplatent)$ and prepended to the sequence of tokens corresponding to grid cells. Thus, the density estimator can utilize latent information via self-attention without the additional complexity introduced by cross-attention or adaptive normalizations.

\paragraph{Positional Embedding for Motion Modeling}
We use 3D Axial RoPE~\cite{su2023roformerenhancedtransformerrotary, heo2024rotary} for positional embedding in our Transformer models. However, we repurpose the spatial components of 3D RoPE to mitigate information leakage and foster trajectory coherence.
Thus, for our encoder model, we use ground truth timestep information, anchoring each grid latent to a specific timepoint, yet we use only the ground truth position for the first timestep $\tau=0$ for spatial positioning. We assume that the position at $\tau=0$ is available through observation, while future positions are generally unknown.
Thus, the input to RoPE for each token is a position $\mathrm{pos}$ given by
\begin{equation}
    \mathrm{pos} = (x_{\tau=0}, y_{\tau=0}, \tau),
\end{equation}
which informs the token about its spatial origin and temporal position.
Note that each point $t^i_\tau$ on the trajectory $t^i$ for scene element $i$ shares the same 2D start position $(x_{\tau=0}, y_{\tau=0})$, thus enabling the model to identify which points lie on the same trajectory.

Image tokens use the same coordinate system as spatial trajectory positions and are thus positionally embedded with the same positional embedding, setting $\tau = 0$. This eases the cross-attention of trajectory tokens to the corresponding image regions by aligning their rotation.

For the full decoder, we further prevent information leakage by choosing random positions $\tilde x_{\tau =0}, \tilde y_{\tau=0} \sim [0, 1]$ while still using true temporal information $\tau$. Thus, we retain the ability to infer which points lie on the same trajectory through shared spatial positions, yet true start positions have to be obtained from the latents, preventing any information leakage.
The density estimator uses standard 2D Axial RoPE to embed positions of grid cells, while the point-wise decoder model does not require any positional embedding.

\paragraph{Static Camera Conditioning}
We reuse the static camera detection from FPT~\cite{baumann2025whatif}. Thus, we consider the camera to be static if less than a fraction $\alpha$ of all movements in the scene exhibit motion magnitudes greater than $m$,
\begin{equation}
\begin{aligned}
     \mathrm{static}_{\alpha, m} = &\frac{1}{(H-1)N} \sum_{i=0}^N \sum_{\tau=0}^{H-1} \mathbf{1}_{\lVert t^i_{\tau+1} - t^i_{\tau}  \rVert_2 > m} < \alpha \\
     &\text{where, } \mathbf{1}_{c} =
    \begin{cases}
    1,& \text{if } c\\
    0,& \text{otherwise}
    \end{cases}
\end{aligned}
\end{equation}
where $N$ is the number of trajectories and $H$ is the time horizon.
We manually tune $\alpha = 0.45$ and $m = 0.0002$ on our training dataset, maximizing the $F1$ score of correct detections on 100 manually labeled training clips with TapNext annotations.
We then embed $c = \mathrm{emb}(\mathrm{static})$ and apply Adaptive Normalization~\cite{xu2019understandinga, peebles2023scalablediffusionmodelstransformers} defined in \cref{eq:supp:adanorm} in the encoder, where scale $\gamma(c)$ and shift $\beta(c)$ are determined by projections of the static camera embedding $c$, thus $(\gamma(c), \beta(c)) = \mathrm{proj(c)}$.
This enables training on mixed data with both moving and static cameras, while fixing the camera for interpretable inference results. Therefore, the latent embedding encodes the information as to whether the scene has a static or moving camera.

\paragraph{Robotics planning data processing} To match Track2Act's dense motion prediction, we select the 256 tracks with the highest motion magnitude per video sample to evaluate our method and dismiss the rest as static background. Because our model is trained with $i \in \{1\dots64\}$ tracks (s. \cref{sec:experiments:setup}), we split the 256 tracks into 4 disjoint subsets of 64 tracks which serve as the GT for our evaluation.

\section{General Flow Matching}
\label{sec:supp:general_flow_matching}
Generative models aim to produce samples from an unknown data distribution. We build upon the framework of flow matching~\cite{lipman2022flow, liu2022flow, albergo2022building, albergo2023stochastic}, which demonstrated strong mode coverage and high sample quality in complex domains such as image and video generation~\cite{ma2024sit, rombach2022highresolution, esser2024scaling, meta2024moviegen}.

Flow matching generates samples from a data distribution by transferring samples from a simple prior distribution $\mathbf{x}_0 \sim p_0$ to samples from the data distribution $\mathbf{x}_1 \sim p_1$.
This transformation is defined through a continuous time-dependent process $\mathbf{x}_{\kappa} = \alpha_{\kappa} \mathbf{x}_1 + \sigma_{\kappa} \mathbf{x}_0$, where $\alpha_{\kappa}$ and $\sigma_{\kappa}$ are functions that control the interpolation process over diffusion time $\kappa \in [0, 1]$.
Training a model to predict the \emph{velocity field}
\begin{equation}\label{eq:flow-ode}
    \mathbf{v}(\mathbf{x}_{\kappa}, \kappa) = \frac{\mathrm{d} x}{\mathrm{d}\kappa} = \dot \alpha_{\kappa} \mathbf{x}_1 + \dot \sigma_{\kappa} \mathbf{x}_0
\end{equation}
allows sampling from the data distribution by numerically integrating this process over diffusion time.

We train $\mathbf{v}_\theta(\mathbf{x}_i, i)$ using the standard flow matching loss
\begin{equation}\label{eq:flow-loss}
    \mathcal{L}_\mathbf{v}(\theta)
    =
    \int \mathbb{E}[\lVert \mathbf{v}_\theta(\mathbf{x}_{\kappa}, \kappa) - (\dot{\alpha}_{\kappa} \mathbf{x}_1 + \dot{\sigma}_{\kappa} \mathbf{x}_0) \rVert^2] \text{ d} \kappa .
\end{equation}
We use the Rectified Flow interpolant~\cite{liu2022flow} defined by $\alpha_{\kappa} = \kappa$ and $\sigma_{\kappa} = (1 - \kappa)$.

The density of the data distribution $p_1$ can be estimated via Monte Carlo (MC) sampling by generating $N_{MC}$ samples from the model starting from different prior samples. In theory, this estimate converges to the true density as $N_{MC} \rightarrow \infty$.
However, in practice, accurate density estimation remains prohibitively expensive for large models.

\section{Metric Details}\label{sec:sup:metrics}

\paragraph{End-Point Error}
We measure the end-point error (EPE) as the average distance from predicted to ground truth trajectories. Consider a set of trajectories $T = \{ t_\tau^i \}$ for $N$ scene elements $i$ and $H$ timesteps $\tau$. Given a prediction $\tilde T = \{ \tilde {t}_\tau^i \}$ the endpoint error for this example is computed as
\begin{equation}
    \mathrm{EPE}(T, \tilde T) = \frac{1}{H \cdot N} \sum_{i = 0}^{N} \sum_{\tau = 0}^{H - 1} d^i_{\timestep} ,
\end{equation}
where 
\begin{equation}\label{eq:sup:track_delta}
    d_\tau^i = \lVert t_\tau^i - \tilde t_\tau^i \rVert_2^2 .
\end{equation}
For a fair comparison, we evaluate EPE only for predicted trajectory points $\bar T = T \setminus \hat T$ and ignore conditional positions $\hat T$.

As multiple futures may align with the same constraints, we compute an ensemble of predictions $\tilde E = \{ \tilde T_n \}$ for $n \in [1, N_{MC}]$ and take the minimum value
\begin{equation}
    \mathrm{minEPE}(T, E) = \min_{\tilde T \in E} \mathrm{EPE}(T, \tilde T).
\end{equation}
Thus, each method has the opportunity to make a range of possible predictions.
Note that $\mathrm{minEPE}$ is equivalent to the commonly used $\mathrm{minADE}$ metric.
Unless noted otherwise, $\mathrm{EPE}$ refers to $\mathrm{minEPE}$ averaged over multiple scenes, simplifying notation for the sake of readability.
For pedestrian trajectory prediction evaluations on ETH/UCY in \cref{tab:exp:sdd_eth_combined} we follow standard practice and average $\mathrm{minADE}$ over observed agents instead of scenes.

\paragraph{Final Distance Error}
Following standard practice, we additionally report the Final Distance Error (FDE), the $\mathrm{L2}$-distance between predicted and ground truth final positions. Therefore, for trajectories $T = \{ t_\tau^i \}$ for $N$ scene elements $i$ and $H$ timesteps $\tau$ the FDE is given by the average error of predicted trajectories $\tilde T = \{ \tilde {t}_\tau^i \}$ at the last timestep $H-1$:
\begin{equation}
    \mathrm{FDE}(T, \tilde T) = \frac{1}{N} \sum_{i}^{N} \lVert t^i_{H-1} - \tilde t^i_{H-1} \rVert_2^1.
\end{equation}
Similar to EPE, we report $\mathrm{FDE}$ only for predicted trajectory points $\bar T = T \setminus \hat T$ and use the minimum $\mathrm{FDE}$ over an ensemble of predictions averaged over multiple scenes. For ETH/UCY we again average over agents instead of scenes, following common practice.

\paragraph{Percentage of Correct Keypoints} We report the Percentage of Correct Keypoints (PCK) as an accuracy metric for trajectory prediction given pre-defined thresholds.
Given trajectories $T = \{ t_\tau^i \}$ for $N$ scene elements $i$ and $H$ timesteps $\tau$ as well as predictions $\tilde T = \{ \tilde {t}_\tau^i \}$, we first compute the per-trackpoint distances $D = \{ d_\tau^i \}$ (s. \cref{eq:sup:track_delta}).
Then, for threshold $\alpha$ the $\mathrm{PCK}@\alpha$ is given by
\begin{equation}\label{eq:sup:pck}
\begin{aligned}
    \mathrm{PCK}@\alpha(D) &= \mathrm{PCK}@\alpha(\{ d_\tau^i \}) = \frac{1}{H \cdot N} \sum_{\tau = 0}^{H-1} \sum_{i=0}^{N} \mathbf{1}_{d_\tau^i < \alpha} \\ &\text{where, } \mathbf{1}_{c} =
    \begin{cases}
    1,& \text{if } c\\
    0, & \text{otherwise}.
\end{cases}
\end{aligned}
\end{equation}
Therefore, $\mathrm{PCK}@\alpha$ measures the fraction of predictions that land with-in a distance $\alpha$ of the round truth.

We define $\alpha$ relative to the image size and report $\mathrm{PCK}@10\%$ and $\mathrm{PCK}@1\%$ only for the predicted points, ignoring the points provided as conditional information. We again average $\mathrm{PCK}$ values over multiple scenes and take the best prediction from an ensemble of $N_{MC}$ samples by taking the maximum.

\paragraph{Area Under the Curve} For our robotic planning evaluation we follow Track2Act~\cite{bharadhwaj2024track2act} and report the \emph{area under the curve} over PCK thresholds.
Effectively, we estimate the integral over PCK (s. \cref{eq:sup:pck}) with a sum over ten $\alpha$ thresholds:
\begin{equation}
    \Delta = \frac{1}{A} \sum^{A}_{a=1} \mathrm{PCK}@\frac{a}{S}(D) \approx \int_{0}^{A} \mathrm{PCK}@\frac{a}{S}(D) \, \text{d}a ,
\end{equation}
where $D$ are L2-distances between track points, $A=10$ is the highest threshold in pixel units and $S=256$ is the pixel resolution such that $0.39\,\% \leq \alpha \leq 3,91\,\%$.

\paragraph{Energy Score}~\cite{gneiting2007strictly} (ES) is a proper scoring rule for evaluating probabilistic forecasts of multivariate variables. Given a predictive distribution $P$ over $X \in \mathbb{R}^d$ and an observed value $y$ the energy score is computed as
\begin{equation}
    \mathrm{ES}(P,y) = \mathbb{E}_{X \sim P} \big [\lVert X - y \rVert \big] -  \frac{1}{2} \mathbb{E}_{X, X' \sim P}\big [\lVert X - X' \rVert \big].
\end{equation}
The first term measures the expected distance of samples from the distribution to the observation, while the second term rewards diversity in the predictive distribution. Therefore, $\mathrm{ES}$ balances closeness to the observation with internal diversity.
Lower values indicate more calibrated predictions.

If the distribution $P$ is represented by an ensemble of $N_{MC}$ individual samples $\{x_1, \dots, x_{N_{MC}} \}$ the expectations can be approximated empirically, yielding the \emph{ensemble energy score}
\begin{equation}
    \mathrm{ES} \Big (y, \{ x_n \}_{n=1}^{N_{MC}} \Big) = \frac{1}{N_{MC}} \sum_{n=1}^{N_{MC}} \lVert x_n - y \rVert - \frac{1}{2 {N_{MC}}^2 } \sum_{n=1}^{N_{MC}} \sum_{m=1}^{N_{MC}} \lVert x_n - x_m \rVert.
\end{equation}
Note that, if the ensemble is overly dispersed, the second term reduces the score less than the first term increases, penalizing underconfident predictions.

\paragraph{Discrete Energy Score}
To evaluate calibration of a discretized probability distribution over a grid, we interpret each grid cell as a position by taking the grid cell center normalized to the full image size and ignore out-of-bounds positions.
Therefore, the discretized probability distribution is interpreted as a distribution over $g \times g$ positions $s_{u,v} \in [0,1]^2$ with grid indices $u,v \in \{1,\dots,g\}$.
The probability $P(X = s_{u,v})$ is given by the discrete probability $p_{u,v}$ assigned to the grid cell $(u,v)$.

With this, the multivariate energy score for this distribution can be computed exactly (i.e.\ without the need for empirical ensemble approximation) as
\begin{equation}\label{eq:sup:discrete_es}
\begin{aligned}
    \mathrm{ES}(P,y) = &\sum_{u=1}^{g}\sum_{v=1}^{g} p_{u,v}\,\lVert s_{u,v} - y \rVert_2 \\
    &- \frac{1}{2} \sum_{u=1}^{g}\sum_{v=1}^{g} \sum_{u'=1}^{g}\sum_{v'=1}^{g} p_{u,v} p_{u',v'} \, \lVert s_{u,v} - s_{u',v'} \rVert_2 .
\end{aligned}
\end{equation}
Note that discretization introduces different value ranges compared to the continuous energy score. However, discrete energy scores for ensemble-based methods can be computed by counting the relative frequencies of samples landing in grid cell regions and normalizing the resulting probabilities so that $\sum_{u=1}^{g}\sum_{v=1}^{g} p_{u,v} = 1$. Thus, discretization enables comparison of ensemble-based methods and methods that directly produce non-parametric probability heatmaps.

\paragraph{Latency}
Latency measurements are performed on a single Nvidia H200 GPU with a batch size of one and reported only for computing a single realization, not the full ensemble. We warm-up model runs before computing latency measurements and use compilation where possible. For fairness, the time for embedding the start frame is also included in the latency measurement. We use the pipelines in diffusers~\cite{von-platen-etal-2022-diffusers, wolf2020transformers} for LTX~\cite{HaCohen2024LTXVideo} and CogVideoX~\cite{yang2024cogvideox} and the official repositories for Motion-I2V~\cite{shi2024motioni2v}, Track2Act~\cite{bharadhwaj2024track2act}, and FPT~\cite{baumann2025whatif}.
Note that for FPT we use two auto-regressive loops, resulting in increased latency: auto-regressive updates of the conditioning pokes to achieve consistent predictions, and independent sampling of each timestep to obtain full tracks instead of single-step estimates.

\input{tab/X_sup/comparison_rel}

\section{Conceptual Differences to Prior Motion Models}

For clarification, we elaborate on conceptual differences between \ours~and previous open-set probabilistic motion models. We provide an argumentation, explaining the advantages of our formulation, for each difference.
We separate differences in conditioning the predicted motion (i.e.\ inputs) from differences in what is predicted (i.e.\ outputs). \cref{tab:met:concept_compare} summarizes our comparison.

\subsection{Conditioning}

\paragraph{Future Appearance Independence}
Prior methods such as Track2Act~\cite{bharadhwaj2024track2act} condition on the appearance of the expected result. They require not only a start frame from the present but also a goal frame from the end in full RGB detail.
We argue (i) while appearance differences include all information necessary for motion planning, the information about goals is not directly accessible to the motion model, (ii) goals in the form of images contain spurious additional information that should not influence motion planning (e.g.\ lighting), (iii) information about the final appearance of a system is often hard to obtain, while positions can be more easily provided by users, and (iv) while changes in positions can be inferred from start and goal images by estimating semantic correspondances, inferring RGB details from a start frame and changes in positions is non-trivial, highlighting the increased generality of \ours~over appearance-based conditioning.
Thus, we opt for appearance-free conditioning based only on positional information.

\paragraph{Localized Conditioning}
Textual descriptions used by \textit{What happens next?}~\cite{boduljak2025happensnextanticipatingfuture} and goal images~\cite{bharadhwaj2024track2act} can inform the model about the future. Yet, text is inherently coarse and ambiguous whereas full images are overly detailed.
Thus, for both input modalities models require additional capacity to extract localized information about which scene element should move where.
In comparison, we directly provide localized goals associated with specific scene elements, focussing modeling capacity on actual trajectory planning instead of goal understanding.

\paragraph{Sub-Goal Conditioning}
In addition, prior approaches such as Track2Act~\cite{bharadhwaj2024track2act} and FPT~\cite{baumann2025whatif} only condition the final goal positions.
\ours~additionally enables specification of sub-goals that should be passed on the way to the final goal.
By enabling spatially and temporally sparse, yet arbitrarily distributed conditioning on future positions, our conditioning is maximally flexible. This enables users to provide exactly the information and assumptions about the future they have available and feel confident about, if any.

\subsection{Prediction}

\paragraph{Spatially Sparse Motion}
Most prior open-set motion models such as Track2Act~\cite{bharadhwaj2024track2act}, \textit{What happens next?}~\cite{boduljak2025happensnextanticipatingfuture}, and Motion-I2V~\cite{shi2024motioni2v, pandey2024motionmodeshappennext} predict the future evolution of the scene densely. They predict either optical flow~\cite{shi2024motioni2v, pandey2024motionmodeshappennext} or point trajectories with points sampled on a dense grid~\cite{bharadhwaj2024track2act, boduljak2025happensnextanticipatingfuture}.
Inspired by canonical baselines in closed-domain trajectory prediction~\cite{salzmann2021trajectron, alahi2016social, gupta2018social}
and the more recent open-set FPT~\cite{baumann2025whatif} we argue (i) modeling motion for a spatially sparse subset of relevant points is sufficient to obtain meaningful estimates of the future. Additionally, (ii) modeling motion for spatially sparse points is computationally cheaper, as we focus on what is important instead of wasting effort to model static scene elements.
\begin{wrapfigure}[9]{r}{0.24\textwidth}
    \centering
    \includegraphics[width=\linewidth]{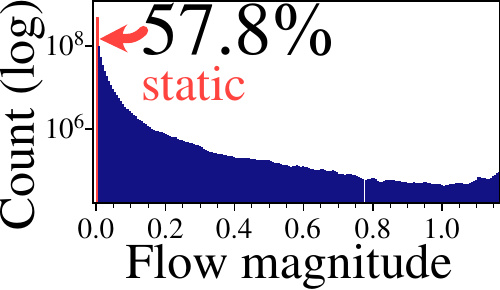}
    \caption{\textbf{Motion Magnitude Histogram.} Most open-set flow magnitudes are small or static.}
    \label{fig:sup:motion_histogram}
\end{wrapfigure}
We empirically find that in open-set scenes a large portion of trajectories is static and visualize a motion histogram in \cref{fig:sup:motion_histogram}. Further, sampling motion on a grid similar to the one used by \textit{What happens next?}~\cite{boduljak2025happensnextanticipatingfuture} incurs a $10\times$ latency increase with our model. Therefore, focusing on salient, moving, off-grid scene elements allows substantially faster inference.

\paragraph{Temporally Dense Motion}
The FPT model~\cite{baumann2025whatif} provides spatially sparse predictions. However, they only model motion between two discrete points in time.
We postulate (i) downstream tasks such as robotic planning benefit from a sequence of small changes rather than big changes between relatively far apart timesteps. Furthermore, (ii) single-step models such as FPT have to account for all interdependencies and interactions internally, limiting their performance in complex multi-object interaction scenarios (see \cref{tab:exp:fde_billiard}).
\ours~predicts temporally dense trajectory estimates for a spatially sparse set of relevant objects.

\paragraph{Sampling-free Uncertainty}
Probabilistic motion models based on generative diffusion~\cite{bharadhwaj2024track2act, shi2024motioni2v, pandey2024motionmodeshappennext, boduljak2025happensnextanticipatingfuture} provide implicit access to the underlying distribution of future motion through repeated sampling.
However, estimating Likelihoods and probabilities of future events as well as quantifying the uncertainty about what might happen in a scene, is costly as it requires Monte-Carlo approximation.
In contrast we propose learning a structured embedding of probability distributions over the future, that can be decoded into probability estimates in less than a millisecond, enabling orders of magnitude faster density estimation. 
Further, in our experiments direct density estimation consistently outperforms Monte-Carlo sampling.

\input{tab/X_sup/ar_vs_full}

\paragraph{Joint Modeling}
FPT~\cite{baumann2025whatif}, which is conceptually closest to our work predicts only a marginal distribution per scene element per timestep and maintains spatial coherence via autoregression (AR). Extending FPT's AR to multi-step horizons (similar to MYRIAD.~\cite{baumann2026envisioning}) requires early commitment which factorizes spatio-temporal inter-dependencies into marginals.
By using a bidirectional encoder, our latents model the joint distribution of possible kinematics where the uncertainty of each point is informed by the whole scene without greedy commitment.
From these latents, we sample full trajectories and estimate densities over future locations.
Empirically, \cref{tab:sup:ar_vs_ours} shows, that our \ours~joint full decoder improves sample quality and efficiency over a diffusion-based FPT-style AR variant extended to multiple steps, and sub-goal conditioning.

\section{Extended Evaluations}\label{sec:sup:extended_eval}

\input{fig/X_sup/calibration_sup}

\paragraph{Test-time Scalability}
\cref{fig:sup:es_nf_ens} demonstrates how more function evaluations and larger Monte Carlo ensembles impact the energy score.
For large ensembles, there is a consistent benefit of investing more compute, in the form of NFEs, into each individual member.
For small ensembles, however, energy score diverges slightly for high NFE. We attribute this to error accumulation in the sampling process.
The same trend applies to discretized energy scores visualized in \cref{fig:sup:des_nfe_mc}. The point-wise decoder requires three orders of magnitude more compute to produce ensembles that come close to the performance of the discrete model.
In general, sampling more and higher quality realizations results in better motion planning results, providing the users with the ability to balance this trade-off according to their current needs.

\paragraph{Extended Comparison of Discretized Energy Scores}
We extend \cref{fig:exp:energy_v_latency} with more expensive baselines in \cref{fig:sup:des_nc} to further emphasize the benefit of our proposed density estimator $\decoderdense$.
Our density decoder produces better predictions than large video generation models with five orders of magnitude reduced latency.

\input{fig/X_sup/calibration_additional}

\paragraph{Additional Calibration Metrics}
To further validate the calibration of densities decoded by $\decoderdense$, we additionally compute Negative log-Likelihoods of GT trajectories (NLL), reliability curves (with $\lvert \hat T \rvert = 4$), and coverage vs.\ nominal coverage curves (with $\lvert \hat T \rvert = 4$). Note that while (discrete) energy scores and NLL are proper scoring rules, reliability and coverage can improve under variance inflation and should therefore be considered only as supplementary to the other metrics.
\cref{fig:sup:nll_gt} shows that our density decoder $\decoderdense$ outperforms not only baseline approaches but also MC estimates from the full decoder $\decoder$ and the point-wise decoder $\decoderhead$ in NLL. This highlights the advantage of direct density decoding over estimating densities via Monte-Carlo sampling.
Additionally, \cref{fig:sup:reliability} and \cref{fig:sup:coverage} show that our density decoder $\decoderdense$ performs competitively with strong baselines for reliability and coverage while consistently outperforming MC sampling.

\input{tab/X_sup/openset_fde_vs_fpt}

\paragraph{Additional Comparison to FPT} The FPT~\cite{baumann2025whatif} model is designed to predict flow between the conditional image $I$ and a single timestep in the future, which differs from \ours~that models motion densely over time.
In particular, FPT can only be fed future knowledge about the exact timestep $\timestep$ it is supposed to predict while our method allows conditioning and prediction at different points in time. As we considered the task of motion planning, we evaluated FPT in \cref{tab:exp:motion_comparison} by rerunning it for different time horizons with spatio-temporally sparse conditioning, matching our method. 
However, \ours~can also be applied in the setting FPT was trained for, conditioning only a single timestep and predicting remaining points for that step.
\cref{tab:exp:openset_fde_vs_fpt} demonstrates that \ours~is competitive in this setting. While FPT achieves slightly superior performance, our method achieves similar performance without training for such motion interpolation tasks.

\input{tab/X_sup/mi2v_oldcond}

\paragraph{Additional Comparison to Motion-I2V} Motion-I2V~\cite{shi2024motioni2v} predicts optical flow to animate images based on text prompts and drags which correspond to our goals.
However, Motion-I2V is designed to accept full tracks over time as input which is inherently different from our spatio-temporally sparse conditioning. To make the baseline comparable, we follow the protocol for sparse conditioning outlined in the Motion-I2V paper~\cite{shi2024motioni2v} and linealy interpolate sparse known goals to full tracks for the comparison in \cref{tab:exp:motion_comparison}.
However, Motion-I2V performs noticeably better in its native setting with full tracks as conditioning (s. \cref{tab:supp:mi2v_oldcond}), but this requires dense knowledge of where the scene element will be at every timestep.

Further, the Motion-I2V repository recommends a cfg scale of 7.0 for inference, which we use throughout our main evaluation aligning. However, we find performance is improved when using cfg 2.0, which we attribute to higher variance predictions leading to better exploration of the search space.
\cref{tab:supp:mi2v_oldcond} shows our method remains superior on all metrics for the sparse conditioning case. However, combining dense conditioning with insights from a cfg sweep, Motion-I2V approaches the quality of our motion planning.

\input{tab/X_sup/pedestrians_eth_ucy_full}

\input{tab/X_sup/pedestrians_sdd_full}

\paragraph{More Details on Pedestrian Trajectory Prediction}
We expand on the evaluation of pedestrian trajectory prediction in \cref{tab:exp:sdd_eth_combined}. We report zero-shot performance as well as performance after minimal finetuning. For finetuned results we train the encoder with the pointwise decoder model for 4k iterations with a batch size of 128. We modify our masking to always provide the initial eight positions as conditions, and train the model to predict the following twelve positions without future knowledge, matching the inference setting. We further mask the loss to only supervise prediction. We then finetune the full decoder by substituting the finetuned encoder and training the full decoder for 4k iterations in the same prediction setting. For the Stanford Drones dataset~\cite{robicquet2016sdd} we use the official train-test-split, while we use a leave-one-out finetuning for the ETH/UCY~\cite{pellegrini2009eth, lerner2007ucy} scenes. As \ours~requires RGB input we only use the ETH, Hotel, Univ, Zara01, and Zara02 scenes for finetuning and evaluation, matching the evaluation performed by IDM~\cite{liu2024intentionaware}.

\cref{tab:exp:sdd_full} shows results for more baselines on the Stanford Drones Dataset (SDD)~\cite{robicquet2016sdd}. Without training for prediction, \ours~achieves zero-shot performance comparable with Social-LSTM~\cite{alahi2016social}. After finetuning, \ours~outperforms Social-GAN~\cite{gupta2018social} and is competitive with CGNS~\cite{li2019cgns} and Goal-GAN~\cite{dendorfer2020goalgan}. Although we do not outperform recent domain-specific methods, this result highlights the general world knowledge learned by \ours.

Similarly, \cref{tab:exp:eth_ucy_full} shows results for more baselines and more subsets of the ETH/UCY datasets. Again, \ours's zero-shot performance without ever training for prediction matches established approaches. While even our finetuned model does not outperform recent state-of-the-art baselines, the ADE of \ours~is below 1\,m still providing useful information for use in downstream applications such as autonomous driving.

Note that the methods we compare against in \cref{tab:exp:sdd_full} and \cref{tab:exp:eth_ucy_full} are specifically designed for pedestrian trajectory prediction and fail to predict the motion of other scene elements. Therefore, the usefulness of these methods in application is limited compared to \ours's general understanding of kinematics.

\input{tab/X_sup/public_vs_private_data}

\paragraph{Alternative Training on Public Data} In the main paper, we present a model trained on a diverse in-house dataset. To ensure reproducibility, we also train \ours~with the same configuration on a subset of the public Koala-36M~\cite{wang2024koala36mlargescalevideodataset} dataset.
\Cref{tab:sup:public_vs_private_data} compares our main model to the model trained on public data. We find negligable differences in performance. We provide checkpoints and results for the model trained on internal data, as we hope the diverse training data introduces more general world knowledge useful as a foundation in downstream domains.

\section{Extended Ablations}\label{sec:sup:extended_ablations}

\input{fig/X_sup/recon}

\paragraph{Reconstruction Quality}
\cref{fig:sup:recon} showcases reconstruction quality for points $\hat T$ provided as a part of $\cond$ to the encoder. We find near-perfect reconstructions ($EPE < 0.003$; $< 1$\,px on a $256^2$ image) with \ours~across the entire range of conditioning density. Therefore, \ours~serves not only as a powerful motion planning method, but is also able to encode given trajectories with minimal loss of information, which is not possible with previous approaches such as Motion-I2V~\cite{shi2024motioni2v}.

\input{tab/X_sup/tracker_agreement}

\paragraph{Performance beyond Tracker Artifacts}
\ours~is trained and evaluated using a pseudo-ground truth obtained by running the off-the-shelf point tracker TapNext~\cite{zholus2025tapnexttrackingpointtap} on video data.
Experiments using human annotated data~\cite{pellegrini2009eth, lerner2007ucy, robicquet2016sdd} and CoTracker3~\cite{karaev2024cotracker3} in \cref{tab:exp:robotics} and \cref{tab:exp:sdd_eth_combined} partially show that our model generalizes beyond its original training setting.
Still, tracker artifacts in supervision and evaluation targets may obstruct evaluation of the true motion understanding.
Consequently, we ablate \ours's motion understanding by testing a model trained using TapNext~\cite{zholus2025tapnexttrackingpointtap} annotations on TapNext as well as CoTracker3~\cite{karaev2024cotracker3} annotated open-set clips from OpenVid-1M in \cref{tab:sup:ours_tapnext_cotracker}.
We find that \ours~performs similarly well when using either tracker as the prediction target.
Furthermore, both TapNext and CoTracker3 estimate similar tracks when given the same videos and are seeded with the same tracking queries (s. \cref{tab:sup:tapnext_cotracker}).
While we do not inspect the biases of each tracking method further, this result validates that our model learns motion concepts beyond the artifacts produced by one tracking method.

\input{tab/X_sup/sm_vs_distill}

\paragraph{Density Estimator Objective}
We train our density estimator using a grid-Cross Entropy loss~\cref{eq:met:grid_ce}. Alternatively, we could distill the distribution learned by the point-wise decoder into a deterministic model by performing Monte-Carlo sampling with the point-wise decoder, constructing a heatmap, and training the density estimator with an MSE loss to regress the heatmap. With this, the alternative loss for the density estimator would become
\begin{equation}\label{eq:sup:distill_loss}
    \mathcal{L}^{\text{alt}}_\omega = \frac{1}{g^2} \sum_{(u,v) \in \mathcal{G}} \lVert p_\omega^{(u,v)}(t_\tau^i \mid \indplatent) - p^{(u, v)}_{\psi, MC}(t^i_\tau \mid \indplatent) \rVert_2^2,
\end{equation}
where $p_\omega^{(u,v)}(t_\tau^i \mid \indplatent)$ is the predicted probability for grid cell $(u,v)$ by the density estimator $\decoderdense$ and $p^{(u, v)}_{\psi, MC}(t^i_\tau \mid \indplatent)$ is obtained by sampling $N_{MC}$ realizations from the pointwise decoder $\decoderhead$ conditioned on latent $\indplatent$ and counting the relative frequency of a point-wise sample landing in grid cell $(u, v)$.

Training with this objective introduces a fundamental trade-off between accurate targets $p^{(u, v)}_{\psi, MC}(t^i_\tau \mid \indplatent)$ and the time required to sample $N_{MC}$ realizations, as $p^{(u, v)}_{\psi, MC}(t^i_\tau \mid \indplatent) \longrightarrow p^{(u,v)}$ if $N_{MC} \longrightarrow \infty$.
Furthermore, \cref{tab:sup:de_objective} shows that a distilled density estimator achieves substantially worse calibration, resulting in a detrimental increase in discrete energy scores.
We conclude that the distilled density estimator overfits to artifacts produced by the point-wise decoder when $N_{MC}$ is too small, while the cross-entropy variant has a more stable supervision target in the ground-truth position.

\input{tab/X_sup/fde_billiards}

\input{fig/X_sup/entropy_cond_robustness}

\paragraph{Robustness of Entropy-based Conditioning Signal}
We evaluate the robustness of using entropy as a signal for where to provide additional conditioning. For this purpose we perturb the additional conditioning provided with random noise, mimicking uncertainty of human annotators about correct constraints.
\cref{fig:sup:entropy_robustness} reveals that entropy-based conditioning is substantially more robust than EPE oracle-based conditioning. Our conditioning variant not only achieves consistently better performance at the same perturbation level but also diverges more slowly under stronger perturbation.

\paragraph{Multi-Object Interactions}
To test force-driven dynamics beyond kinematics, we train our model and FPT~\cite{baumann2025whatif} on data from a synthetic Billiard simulation~\cite{ebke2019billiards} and compare prediction accuracy with up to 16 balls. We select a time horizon of one second (40 steps, $\Delta t=0.025\,s$) as we find this horizon is sufficient for multiple ball-to-ball interactions. Note that this requires FPT to internally model many-ball interactions, while \ours~is able to reason about step-by-step interactions.
\cref{tab:exp:fde_billiard} showcases \ours~achieves relatively stable performance when introducing more balls and thus forcing more ball interactions to occur. While performance slightly deteriorates when moving from two to eight balls, performance remains stable when increasing to 16 interacting scene elements.
In comparison, FPT achieves substantially worse performance across the entire range of ball counts.

\cref{fig:exp:billiard_qual} qualitatively shows Billiard interactions over time given the starting velocity and end-point conditioning. \ours~is able to correctly infer interactions at intermediate timesteps. This highlights \ours's ability to understand and reason about multiple interacting entities.

\input{fig/X_sup/nll_alignment}

\paragraph{Alignment of Samples with Density}
As an additional verification of alignment between sampled trajectories by $\decoder$ and $\decoderhead$ and decoded distributions from $\decoderdense$ we compute the negative log-Likelihood (NLL) of samples under distributions from $\decoderdense$.
\cref{fig:sup:nll_alignment} shows that samples from either $\decoder$ or $\decoderhead$ are more likely under $\decoderdense$ than the ground truth trajectory or trajectories from an alternative model (Motion-I2V~\cite{shi2024motioni2v}).
This further strengthens the evidence for alignment between models that decode the same latent possibility space.

\input{tab/X_sup/single_shot}

\paragraph{Single-shot Performance}
Given only partial future knowledge multiple outcomes can be equally plausible. Therefore, trajectory prediction performance is typically evaluated in a best-of-N setting (cf. \cite{baumann2026envisioning, salzmann2021trajectron, alahi2016social, gupta2018social, boduljak2025happensnextanticipatingfuture}).
However, selecting the best trajectory might be impossible in application settings. Yet, \cref{tab:sup:best_of_one} highlights that \ours~remains highly competitive in the single-shot setting.

\section{Additional Qualitative Samples}

For animated qualitative examples, we refer to the attached video samples summarized in \texttt{video\_samples/index.html}.

We qualitatively compare our model to open-set baselines in \cref{fig:sup:qual_comparison}, qualitatively supplementing \cref{tab:exp:motion_comparison}. For the TI2V models, we show tracks extracted from generated videos using TapNext~\cite{zholus2025tapnexttrackingpointtap} matching the motion planning evaluation setting. CogVideoX~\cite{yang2024cogvideox}, LTX~\cite{HaCohen2024LTXVideo},  predict realistic motion, however due to the lack of appropriate conditioning, their samples are dominated by unwanted camera motion and fail to match the ground truth in speed and direction. FPT~\cite{baumann2025whatif} is unable to predict consistent trajectories due to modeling only the flow between two points in time, requiring independently rerunning the model for each horizon.
Motion-I2V~\cite{shi2024motioni2v} and Track2Act~\cite{bharadhwaj2024track2act} tend to predict low motion magnitudes, limiting their applicability to samples with complex motion.
\ours~is able to sample motion which has the closest alignment to the target with as few as $|\condtracks| = 4$ known positions.

Furthermore, we show examples of \ours's density estimation with $\decoderdense$ in \cref{fig:sup:qual_densities_cond0}, \cref{fig:sup:qual_densities_cond2} and \cref{fig:sup:qual_densities_cond4}. \ours~becomes more certain when $\timestep$ is close to 0 (the present) and/or more conditioning $\condtracks$ is given. Even when no future knowledge is provided, the ground-truth is within or close to the regions which \ours~considers likely and \ours~becomes more confident in the correct regions when conditioned on more knowledge about the future.

In \cref{fig:sup:qual_robotics} and \cref{fig:sup:qual_pedestrian}, we visualize predicted trajectories for robotics and pedestrians. When given only the final goal position of each trajectory in the robotics setting, \ours~is able to plan reasonable motion for robotics without any domain-specific training. After fine-tuning \ours, it can predict realistic trajectories of pedestrians in traffic scenes.

\ours~is able to handle and predict dynamics, e.g. multi-object inter-actions in Billiard simulations as shown in \cref{fig:exp:billiard_qual}. It is able to predict and resolve multiple collisions based on the final positions and initial velocities of each ball, which demonstrates its understanding of physical dynamics.

\FloatBarrier

\input{fig/X_sup/qualitative_comparison/qual_comparison}

\input{fig/X_sup/qualitative_densities/qual_densities}

\input{fig/X_sup/qualitative_robotics/robotics_qualitative}

\input{fig/X_sup/qualitative_pedestrians/qual_pedestrians}

\input{fig/X_sup/qualitative_billiard/billiard_qualitative}

\FloatBarrier

\section{Copyright}
\ifeccv\else
    The style used for this paper is adapted from the \href{https://arxiv.org/abs/2407.15595v2}{arXiv preprint \emph{Discrete Flow Matching} (Gat et al., 2024)}, licensed under \href{https://creativecommons.org/licenses/by/4.0/}{CC BY 4.0}.
    Throughout the paper and figures/plots, we use Fira Sans (licensed under the \href{https://openfontlicense.org/open-font-license-official-text/}{OFL v1.1}) for bold text.
\fi

%% file: tab/X_sup/hyperparam.tex
\begin{table}[t]
    \centering
    \footnotesize
    {
    \adjustbox{max width=\textwidth}{
    \begin{tabular}{lcccc}
        \toprule
        \multicolumn{1}{c}{\textbf{Parameter}} & \textbf{Encoder} & \textbf{Pointwise Decoder} & \textbf{Full Decoder} & \textbf{Density Estimator} \\
        \midrule
        Depth & 24 & 3 & 28 & 12 \\
        Width & 1024 & 1024 & 1152 & 768 \\
        Normalization & RMSNorm~\cite{zhang2019root} & LayerNorm~\cite{ba2016layer} & RMSNorm~\cite{zhang2019root} & RMSNorm~\cite{zhang2019root}\\
        FFN expand factor & 3 & \notavail & 3 & 3 \\
        Activation & SwiGLU~\cite{shazeer2020glu} & GELU~\cite{hendrycks2023gaussian} & SwiGLU~\cite{shazeer2020glu} & SwiGLU~\cite{shazeer2020glu} \\
        Positional Encoding & 3D RoPE~\cite{heo2024rotary, su2023roformerenhancedtransformerrotary} & \notavail & 3D RoPE~\cite{heo2024rotary, su2023roformerenhancedtransformerrotary}  & 2D RoPE~\cite{heo2024rotary, su2023roformerenhancedtransformerrotary} \\
        Static Scene Conditioning & AdaLN~\cite{peebles2023scalablediffusionmodelstransformers, xu2019understandinga} & \notavail & \notavail & \notavail \\
        Latent Conditioning & \notavail & AdaLN~\cite{peebles2023scalablediffusionmodelstransformers, xu2019understandinga} & In-Context & In-Context \\
        Image Encoder & DINOv2R-B~\cite{oquab2023dinov2, darcet2023vitneedreg} & \notavail & \notavail & \notavail \\
        \arrayrulecolor{gray!50!white}
        \midrule
        \arrayrulecolor{black}
        Batch size & \multicolumn{4}{c}{256} \\
        Optimizer &  \multicolumn{4}{c}{AdamW~\cite{loshchilov2019decoupled}} \\
        Learning rate & $1e-4$ & $1e-4$ & $1e-4$ & $1e-5$ \\
        LR schedule &  \multicolumn{4}{c}{Linear + Constant} \\
        LR schedule steps &  \multicolumn{4}{c}{300 Linear warmup steps} \\
        Betas &  \multicolumn{4}{c}{(0.9, 0.95)} \\
        Conditioning sparsity &  \multicolumn{4}{c}{1\% conditioning}\\
        Sparsity schedule & \multicolumn{3}{c}{Linear from 50\% + Constant}  &  \multicolumn{1}{c}{Constant} \\
        Sparsity schedule steps & \multicolumn{3}{c}{50k Linear warmup steps} & \notavail \\
        Training steps & \multicolumn{2}{c}{450k} & 200k & 150k \\
        Precision & \multicolumn{4}{c}{$\mathrm{bfloat16}$} \\
        Trainable Parameters & 510M & 34M & 522M & 97M \\
        \arrayrulecolor{gray!50!white}
        \midrule
        \arrayrulecolor{black}
        GPUs &  \multicolumn{4}{c}{16 Nvidia H200} \\
        Training Time & \multicolumn{2}{c}{4 days} & 1 day & 1 day \\
        \arrayrulecolor{gray!50!white}
        \midrule
        \arrayrulecolor{black}
        Dataset &  \multicolumn{4}{c}{Open-Set Videos} \\
        Number of clips &  \multicolumn{4}{c}{20M} \\
        Tracker &  \multicolumn{4}{c}{TapNext~\cite{zholus2025tapnexttrackingpointtap}} \\
        Tracker Position Seeding &  \multicolumn{4}{c}{1024 random positions} \\
        Position Scale &  \multicolumn{4}{c}{[0,1]} \\
        Image size &  \multicolumn{4}{c}{$224 \times 224$} \\
        \bottomrule
    \end{tabular}
    }}
    \caption{\textbf{Hyperparameters.} We summarize \ours~and training configurations for all components of our method below. }
    \label{tab:supp:hyperparam}
\end{table}

%% file: tab/X_sup/comparison_rel.tex
\begin{table*}[t]
    \centering    \newcommand{\inte}{\color{ourgray}}
    {
    \adjustbox{max width=\linewidth}{
    \begin{tabular}{l c c c c c c }
        \toprule
        \multirow{3.5}{*}{\textbf{Method}} & \multicolumn{3}{c}{\textbf{Conditioning}} & \multicolumn{3}{c}{\textbf{Prediction}}\\
        \cmidrule(lr){2-4} \cmidrule(lr){5-7}
        & \makecell{Independent\\of appearance} & \makecell{Localized\\motion} & \makecell{Allows\\sub-goals} & \makecell{Spatially\\sparse} & \makecell{Temporally\\dense} & \makecell{Sampling-free\\uncertainty} \\
        \midrule
        Track2Act ~\cite{bharadhwaj2024track2act}          &  \no &  \no &  \no & \yes & \yes &  \no \\
        \makecell[l]{Motion-I2V~\cite{shi2024motioni2v}/\\Motion Modes~\cite{pandey2024motionmodeshappennext}}       & \yes & \yes &  \yes &  \no & \yes &  \no  \\
        FPT ~\cite{baumann2025whatif}                & \yes & \yes &  \no & \yes &  \no & \yes  \\
        What happens next ~\cite{boduljak2025happensnextanticipatingfuture}  & \yes &  \no &  \no &  \no & \yes &  \no  \\
        \oursours & \yes & \yes & \yes & \yes & \yes & \yes  \\
        \bottomrule
    \end{tabular}
    }}
    \caption{\textbf{Conceptual comparison}. Unlike related approaches, we aim to model the distribution of motion densely over time with spatially sparse trajectories while also allowing single-NFE uncertainty estimation without expensive diffusion sampling.}
    \label{tab:met:concept_compare}
\end{table*}

%% file: tab/X_sup/ar_vs_full.tex
\begin{table}[t]
    \centering
    \footnotesize
    \setlength{\tabcolsep}{2pt}
    \begin{tabular}{l c c c}
        \toprule
        \textbf{Method}
        & \textbf{Goals}
        & \textbf{EPE} $\downarrow$
        & \textbf{Latency} $\downarrow$\\
        \midrule
        \multirow{2}{*}{AR ($\decoderhead$)}
            & 4  & 0.076 & \multirow{2}{*}{71.4} \\
            & 16 & 0.040 & \\
        \multirow{2}{*}{\oursours~$\decoder$}
            & 4  & 0.014 & \multirow{2}{*}{\textbf{0.40}} \\
            & 16 & \textbf{0.012} & \\
        \bottomrule
    \end{tabular}
    \captionof{table}{\textbf{AR vs.\ joint.} Joint prediction achieves better trajectory quality and lower latency than AR rollouts. Evaluated on held-out validation set of training data.}
    \label{tab:sup:ar_vs_ours}
\end{table}

%% file: fig/X_sup/calibration_sup.tex
\begin{figure}[t]
    \centering
    \begin{subfigure}{0.32\textwidth}
        \includegraphics[width=\textwidth]{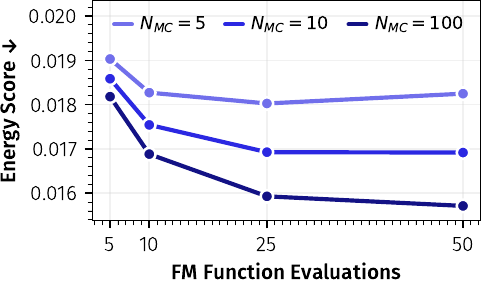}
        \caption{Energy Score vs.\ NFE and MC samples for point-wise $\decoderhead$.}
        \label{fig:sup:es_nf_ens}
    \end{subfigure}
    \hfill
    \begin{subfigure}{0.32\textwidth}
        \includegraphics[width=\textwidth]{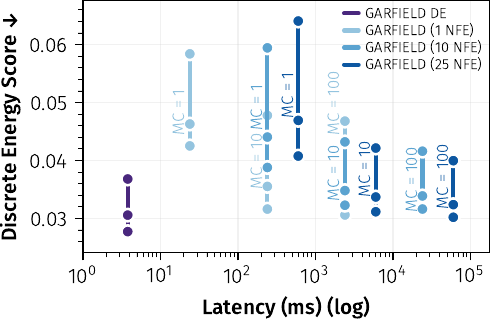}
        \caption{Discrete Energy Score vs.\ NFE and MC samples.
        }
        \label{fig:sup:des_nfe_mc}
    \end{subfigure}
    \hfill
    \begin{subfigure}{0.32\textwidth}
        \includegraphics[width=\textwidth]{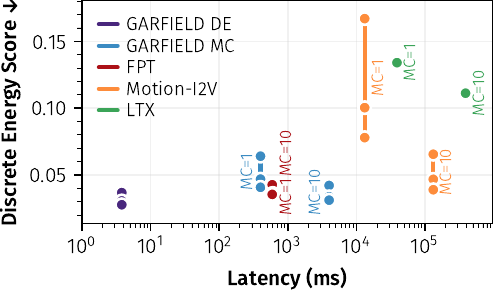}
        \caption{Discrete Energy Score vs.\ Latency for more baselines.
        }
        \label{fig:sup:des_nc}
    \end{subfigure}
    \caption{\textbf{Calibration trade-offs.} We visualize trade-offs between well-calibrated density estimation, indicated by low (discrete) energy score, and the compute latency to estimate these densities. While more NFEs and larger ensembles improve motion prediction, they also induce additional latency. In \textbf{(b)} and \textbf{(c)}, we also show how adding conditionals ($|\condtracks| \in \{0,1,2\}$) further lowers the discrete energy score for the same model and ensemble size.}
    \label{fig:sup:calibration_nfe_ens}
\end{figure}

%% file: fig/X_sup/calibration_additional.tex
\begin{figure}[t]
    \centering
    \begin{subfigure}{0.3\textwidth}
        \includegraphics[width=\linewidth]{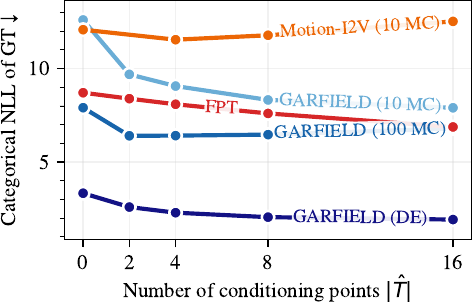}
        \caption{NLL of GT trajectory.}\label{fig:sup:nll_gt}
    \end{subfigure}
    \begin{subfigure}{0.3\textwidth}
        \includegraphics[width=\linewidth]{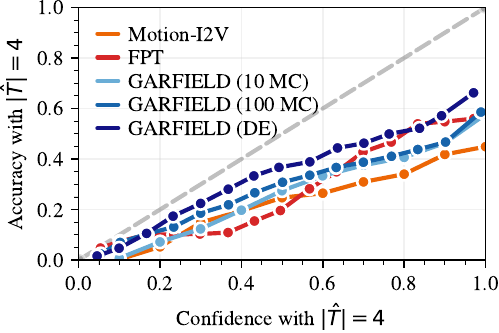}
        \caption{Reliability Diagram.}\label{fig:sup:reliability}
    \end{subfigure}
    \begin{subfigure}{0.3\textwidth}
        \includegraphics[width=\linewidth]{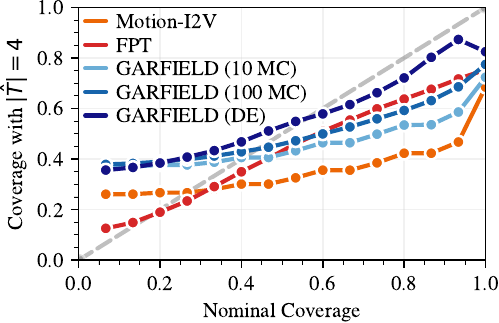}
        \caption{Coverage vs.\ Nominal Coverage.}\label{fig:sup:coverage}
    \end{subfigure}
    \caption{\textbf{Calibration of Densities.} Densities from the density decoder $\decoderdense$ consistently outperform Monte-Carlo sampling and outperforms strong baselines in terms of negative log-Likelihood of the ground truth trajectory. Evaluated on held-out validation set of training data.}
    \label{fig:sup:add_calibration}
\end{figure}

%% file: tab/X_sup/openset_fde_vs_fpt.tex
\begin{table}[t]
    \setlength{\tabcolsep}{5pt}
    \centering
    {
    \adjustbox{max width=\linewidth}{
    \begin{tabular}{l c c c c c c}
        \toprule
        \multicolumn{1}{c}{\multirow{2}{*}{\textbf{Method}}} &
        \multicolumn{6}{c}{\textbf{FDE with $|\condtracks|$ End-Point Goals}}\\
        \cmidrule{2-7}
        & 0
        & 1
        & 2
        & 4
        & 8
        & 16 \\
        \midrule

        FPT~\cite{baumann2025whatif}
        & 0.017
        & 0.017
        & 0.015
        & 0.014
        & 0.013
        & 0.009 \\

        \oursours
        & 0.022
        & 0.020
        & 0.019
        & 0.018
        & 0.014
        & 0.013 \\

        \bottomrule
    \end{tabular}
    }}
    \caption{\textbf{End-point motion interpolation.} We compare our model's ability to infer where scene elements end up given end-point for other elements in the scene. Evaluated on held-out validation set of training data.}
    \label{tab:exp:openset_fde_vs_fpt}
\end{table}

%% file: tab/X_sup/mi2v_oldcond.tex
\begin{table}[b]
    \centering
    \newcommand{\inte}{\color{ourgray}}
    \newcommand{\vcat}[1]{\rotatebox{90}{\scriptsize#1}}
    {
    \adjustbox{max width=\linewidth}{
    \footnotesize
    \begin{tabular}{l c c c c c c}
        \toprule
        \textbf{Method}
        & \makecell{\textbf{Num.}\\\textbf{Goals}}
        & \textbf{CFG}
        & $\mathbf{\mathrm{\mathbf{EPE}}\downarrow}$
        & \makecell{$\mathrm{\mathbf{PCK}}\uparrow$\\@$\mathbf{10\%}$}
        & \makecell{$\mathrm{\mathbf{PCK}}\uparrow$\\@$\mathbf{1\%}$}
        & $\mathbf{\mathrm{\mathbf{FDE}}\downarrow}$
        \\
        \midrule

        \multirow{4}{*}{Motion-I2V~\cite{shi2024motioni2v}} 
            & 4 & 7.0  & 0.055 & 0.864 & 0.150 & 0.060 \\
            & 16 & 7.0 & 0.033 & 0.957 & 0.263 & 0.037 \\
            & 4 & 2.0  & 0.022 & 0.982 & 0.428 & 0.029 \\
            & 16 & 2.0 & 0.016 & 0.992 & 0.577 & 0.030 \\
            & $4 \times H$  & 7.0 & 0.051 & 0.886 & 0.166 & 0.053 \\
            & $16 \times H$ & 7.0 & 0.027 & 0.981 & 0.312 & 0.030 \\
            & $4 \times H$  & 2.0 & 0.021 & 0.986 & 0.447 & 0.026 \\
            & $16 \times H$ & 2.0 & 0.013 & 0.996 & 0.602 & 0.016 \\
        \midrule
        \multirow{2}{*}{\oursours} 
            & 4  & 1.0 & 0.020 & 0.973 & 0.544 & 0.026 \\
            & 16 & 1.0 & 0.013 & 0.989 & 0.666 & 0.017  \\
        \bottomrule
    \end{tabular}
    }}
    \caption{\textbf{Extended comparison to Motion-I2V}~\cite{shi2024motioni2v} When fed full trajectories with horizon $H$ as input, Motion-I2V performs noticeably better at motion planning.}\label{tab:supp:mi2v_oldcond}
\end{table}

%% file: tab/X_sup/pedestrians_eth_ucy_full.tex
\begin{table}[t]
    \centering
    {
    \adjustbox{max width=\linewidth}{
    \footnotesize
    \begin{tabular}{lcccccccccc}
        \toprule
        \multicolumn{1}{c}{\textbf{Methods}} &
        \multicolumn{2}{c}{\textbf{ETH}} &
        \multicolumn{2}{c}{\textbf{Hotel}} &
        \multicolumn{2}{c}{\textbf{Univ}} &
        \multicolumn{2}{c}{\textbf{Zara1}} &
        \multicolumn{2}{c}{\textbf{Zara2}} \\
        \cmidrule(lr){2-3} \cmidrule(lr){4-5} \cmidrule(lr){6-7} \cmidrule(lr){8-9} \cmidrule(lr){10-11}
         & ADE & FDE & ADE & FDE & ADE & FDE & ADE & FDE & ADE & FDE \\
        \midrule
        Social-LSTM      & 1.09 & 2.35 & 0.79 & 1.76 & 0.67 & 1.40 & 0.47 & 1.00 & 0.56 & 1.17 \\
        Social-STGCN     & 0.64 & 1.11 & 0.49 & 0.85 & 0.44 & 0.79 & 0.34 & 0.53 & 0.30 & 0.48 \\
        Causal-STGCN     & 0.64 & 1.00 & 0.38 & 0.45 & 0.49 & 0.81 & 0.34 & 0.53 & 0.32 & 0.49 \\
        STAR             & 0.36 & 0.65 & 0.17 & 0.36 & 0.31 & 0.62 & 0.26 & 0.55 & 0.22 & 0.46 \\
        Expert-GMM       & 0.37 & 0.65 & \textbf{0.11} & \textbf{0.15} & \textbf{0.20} & 0.44 & \textbf{0.15 }& 0.31 & 0.12 & 0.26 \\
        PCCSNET          & \textbf{0.28} & 0.54 & \textbf{0.11} & 0.19 & 0.29 & 0.60 & 0.21 & 0.44 & 0.15 & 0.34 \\
        Social-GAN       & 0.81 & 1.52 & 0.72 & 1.61 & 0.60 & 1.26 & 0.34 & 0.69 & 0.42 & 0.84 \\
        Goal-GAN         & 0.59 & 1.18 & 0.19 & 0.35 & 0.60 & 1.19 & 0.43 & 0.87 & 0.32 & 0.65 \\
        Social-BiGAT     & 0.69 & 1.29 & 0.49 & 1.01 & 0.55 & 1.32 & 0.30 & 0.62 & 0.36 & 0.75 \\
        MG-GAN           & 0.47 & 0.91 & 0.14 & 0.24 & 0.54 & 1.07 & 0.36 & 0.73 & 0.29 & 0.60 \\
        CGNS             & 0.62 & 1.40 & 0.70 & 0.93 & 0.48 & 1.22 & 0.32 & 0.59 & 0.35 & 0.71 \\
        PECNET           & 0.54 & 0.87 & 0.18 & 0.24 & 0.35 & 0.60 & 0.22 & 0.39 & 0.17 & 0.30 \\
        Trajectron++     & 0.39 & 0.83 & 0.12 & 0.21 & \textbf{0.20} & 0.44 & \textbf{0.15} & 0.33 & \textbf{0.11} & \textbf{0.25} \\
        LB-EBM           & 0.30 & \textbf{0.52} & 0.13 & \textbf{0.20} & 0.27 & 0.52 & 0.20 & 0.37 & 0.15 & 0.29 \\
        MID              & 0.39 & 0.66 & 0.13 & 0.22 & 0.22 & 0.45 & 0.17 & 0.30 & 0.13 & 0.27 \\
        IDM              & 0.41 & 0.62 & 0.15 & 0.25 & \textbf{0.20} & \textbf{0.42} & 0.17 & \textbf{0.28} & 0.12 & \textbf{0.25} \\
        \arrayrulecolor{gray!50!white}
        \midrule
        \arrayrulecolor{black}
        \oursours {\tiny 0-shot}       & 1.09 & 1.79 & 0.41  & 0.71 & 0.60 & 1.12 &  0.72 & 1.37 &  0.55 & 1.05  \\
        \oursours {\tiny FT}           & 0.84 & 1.44 &  0.29 & 0.56 & 0.58 & 1.19 &  0.33 & 0.65 &  0.33 & 0.67  \\
        \bottomrule
    \end{tabular}}}
    \caption{\textbf{ETH/UCY Pedestrian Prediction.} We show an extended variant of \cref{tab:exp:sdd_eth_combined} with all scenes from ETH/UCY~\cite{pellegrini2009eth, lerner2007ucy} and more baselines. Errors are in meters. Our method achieves competitive performance while enabling open-set motion modeling.}
    \label{tab:exp:eth_ucy_full}
\end{table}

%% file: tab/X_sup/pedestrians_sdd_full.tex
\begin{table}[t]
    \centering
    {\footnotesize
    \begin{tabular}{lcc}
        \toprule
        \multicolumn{1}{c}{\textbf{Method}}          & \textbf{minADE-20} & \textbf{minFDE-20} \\
        \midrule
        Social-LSTM     & 57.00  & 31.20 \\
        Expert+GMM       & 10.67  & 14.38 \\
        SimAug          & 10.27 & 19.71 \\
        PCCSNET         & 8.62 & 16.16 \\
        Y-Net           & 8.97 & 14.61 \\
        Social-GAN      & 27.23 & 41.44 \\
        Goal-GAN        & 12.20 & 22.10 \\
        MG-GAN          & 13.60 & 25.80\\
        CGNS            & 15.60 & 28.20 \\
        Trajectron++    & 8.98 & 19.02 \\
        PECNET          & 9.96 & 15.88 \\
        LB-EBM          & 8.87 & 15.61 \\
        MID             & 7.61 & 14.30 \\
        IDM             & \textbf{7.46} & \textbf{13.83 }\\
        \arrayrulecolor{gray!50!white}
        \midrule
        \arrayrulecolor{black}
        \oursours~{\tiny 0-shot} & 36.77  & 49.86 \\
        \oursours~{\tiny FT} & 19.92  & 34.98 \\
        \bottomrule
    \end{tabular}}
    \caption{\textbf{Stanford Drone Dataset}  We show an extended variant of \cref{tab:exp:sdd_eth_combined} for the Stanford Drone Dataset~\cite{robicquet2016sdd}. Errors are in pixels w.r.t.\ the original video resolution. Our method achieves comparable performance, without domain-specific adaptations.}
    \label{tab:exp:sdd_full}
\end{table}

%% file: tab/X_sup/public_vs_private_data.tex
\begin{table}[tb]
    \centering
    \setlength{\tabcolsep}{5pt}
    \begin{subtable}{0.48\linewidth}
    \centering
    \caption{Point-wise $\decoderhead$}
    \begin{tabular}{c c c c c}
    \toprule
         $\mathbf{\lvert \hat T \rvert}$ & \multicolumn{2}{c}{\textbf{EPE} $\downarrow$} & \multicolumn{2}{c}{\textbf{FDE} $\downarrow$}  \\
         \cmidrule(lr){2-3} \cmidrule(lr){4-5}
         & Ours & Koala & Ours & Koala \\
         \midrule
        0  & 0.017 & 0.015 & 0.023 & 0.020 \\
        4  & 0.013 & 0.012 & 0.017 & 0.016 \\
        8  & 0.011 & 0.012 & 0.015 & 0.015 \\
        16 & 0.011 & 0.011 & 0.015 & 0.015 \\
         \bottomrule
    \end{tabular}
    \end{subtable}
    \hfill
    \begin{subtable}{0.48\linewidth}
    \centering
    \caption{Full $\decoder$}
    \begin{tabular}{c c c c c}
    \toprule
         $\mathbf{\lvert \hat T \rvert}$ & \multicolumn{2}{c}{\textbf{EPE} $\downarrow$} & \multicolumn{2}{c}{\textbf{FDE} $\downarrow$} \\
         \cmidrule(lr){2-3} \cmidrule(lr){4-5}
         & Ours & Koala & Ours & Koala \\
         \midrule
        0  & 0.017 & 0.016 & 0.022 & 0.021 \\
        4  & 0.014 & 0.014 & 0.018 & 0.018 \\
        8  & 0.013 & 0.013 & 0.016 & 0.016 \\
        16 & 0.012 & 0.012 & 0.015 & 0.015 \\
         \bottomrule
    \end{tabular}
    \end{subtable}
    \caption{\textbf{Public vs Private Dataset} We compare the effect of training our model with two different video datasets: an in-house dataset and a subset of Koala-36M~\cite{wang2024koala36mlargescalevideodataset}. Training on publicly available data achieves comparable performance, ensuring reproducibility of results. Evalauted on a subset of OpenVid-1M~\cite{nan2024openvid} data.}
    \label{tab:sup:public_vs_private_data}
\end{table}

%% file: fig/X_sup/recon.tex
\begin{figure}
    \centering
    \includegraphics[width=0.4\linewidth]{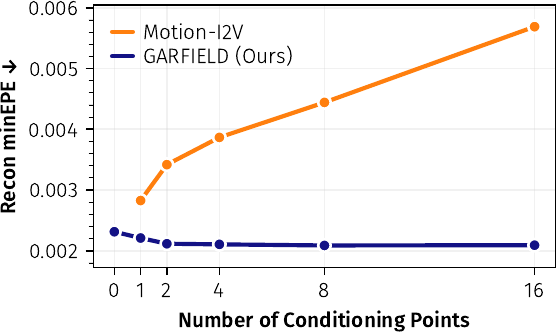}
    \caption{\textbf{Reconstruction Error} \ours~produces accurate reconstructions of conditional points, showcasing its ability to serve as a representation for given points beyond a planning method.}
    \label{fig:sup:recon}
\end{figure}

%% file: tab/X_sup/tracker_agreement.tex
\begin{table}[tb]
    \centering
    \setlength{\tabcolsep}{5pt}
    \begin{subtable}{0.65\textwidth}
    \centering
    \caption{\textbf{Inference performance}}
    \label{tab:sup:ours_tapnext_cotracker}
    \begin{tabular}{ccccccc}
    \toprule
         $\mathbf{\lvert \hat T \rvert}$ & \multicolumn{2}{c}{\textbf{EPE} $\downarrow$} & \multicolumn{2}{c}{\textbf{FDE} $\downarrow$} & \multicolumn{2}{c}{\textbf{PCK@1\%} $\uparrow$}  \\
         \cmidrule(lr){2-3} \cmidrule(lr){4-5} \cmidrule(lr){6-7}
         & TN & CT & TN & CT & TN & CT \\
         \midrule
        0  & 0.017 & 0.017 & 0.022 & 0.022 & 0.731 & 0.725 \\
        2  & 0.015 & 0.015 & 0.018 & 0.019 & 0.786 & 0.771 \\
        8  & 0.013 & 0.013 & 0.016 & 0.016 & 0.807 & 0.799 \\
        16 & 0.012 & 0.012 & 0.015 & 0.015 & 0.821 & 0.812 \\
    \bottomrule
    \end{tabular}
    \end{subtable}
    \hfill
    \begin{subtable}{0.3\textwidth}
    \centering
    \caption{\textbf{Tracker Agreement}}
    \label{tab:sup:tapnext_cotracker}
    \begin{tabular}{l|c}
    \toprule
    \textbf{EPE} $\downarrow$ & 0.003 \\
    \textbf{PCK@}$\mathbf{10\%}$ $\uparrow$ & 0.995 \\
    \textbf{PCK@}$\mathbf{1\%}$ $\uparrow$ & 0.942 \\
    \bottomrule
    \end{tabular}
    \end{subtable}
    \caption{\textbf{Alternative Tracker Pseudo-Ground Truth.} We evaluate the performance of a TapNext-trained model on TapNext (TN)~\cite{zholus2025tapnexttrackingpointtap} and CoTracker (CT)~\cite{karaev2024cotracker3} annotated data and find similar performance. We further find high agreement between TapNext and CoTracker. Therefore, our model generalizes beyond artifacts of a specific tracker, learning useful motion representations. Evaluated on a subset of OpenVid~\cite{nan2024openvid} evaluation data.}
    \label{tab:sup:tracker_agreement}
\end{table}

%% file: tab/X_sup/sm_vs_distill.tex
\begin{table}[tb]
    \centering
    {
    \setlength{\tabcolsep}{7pt}
    \footnotesize
    \begin{tabular}{ccc}
        \toprule
        \multirow{2.2}{*}{$\mathbf{\lvert \hat T \rvert}$} & \multicolumn{2}{c}{\textbf{Discrete Energy Score} $\downarrow$} \\
        \cmidrule(lr){2-3}
        & Cross Entropy & Distillation \\
        \midrule
         0 & 0.061 & 0.202\\
         2 & 0.043 & 0.199 \\
         8 & 0.030 & 0.198 \\
         16 & 0.026 & 0.197 \\
         \bottomrule
    \end{tabular}}
    \caption{\textbf{Density Estimator Objective} Training with a cross-entropy (CE) objective (\cref{eq:met:grid_ce}) results in a lower discrete energy score (dES; \cref{eq:sup:discrete_es}) than training with a distillation objective (\cref{eq:sup:distill_loss}) across any number of conditioning points $\lvert \hat T \rvert$. Evalauted on held-out validation set of training data.}
    \label{tab:sup:de_objective}
\end{table}

%% file: tab/X_sup/fde_billiards.tex
\begin{table}[b]
    \centering
    {\footnotesize
    \begin{tabular}{l c c c c}
    \toprule
         \multirow{2.5}{*}{\textbf{Method}} & \multicolumn{4}{c}{\textbf{EPE} $\downarrow$} \\
         \cmidrule(lr){2-5}
         & 2 Balls & 4 Balls & 8 Balls & 16 Balls \\
         \midrule
         FPT~\cite{baumann2025whatif} & 0.260 & 0.237 & 0.225 & 0.218 \\
         \oursours                         & 0.140 & 0.182 & 0.196 & 0.186 \\
         \bottomrule
    \end{tabular}}
    \caption{\textbf{Multi-object Interactions.} \ours~shows consistent performance with more balls, highlighting it's ability to model interactions between many objects.}
    \label{tab:exp:fde_billiard}
\end{table}

%% file: fig/X_sup/entropy_cond_robustness.tex
\begin{figure}[t]
    \centering
    \includegraphics[width=0.35\linewidth]{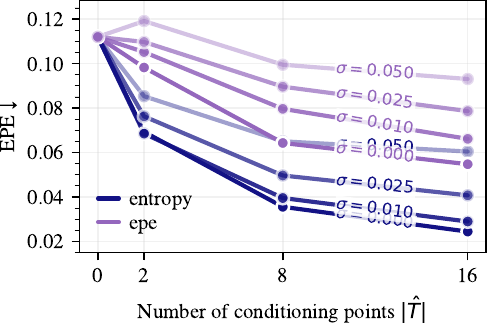}
    \caption{\textbf{Robustness of Entropy-based Conditioning.} Selecting conditioning based on entropy is more robust to perturbed conditioning than EPE-based conditioning. Evaluated on held-out validation set of training data.}
    \label{fig:sup:entropy_robustness}
\end{figure}

%% file: fig/X_sup/nll_alignment.tex
\begin{figure}[t]
    \centering
    \includegraphics[width=0.4\linewidth]{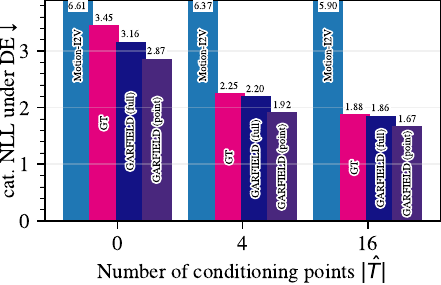}
    \caption{\textbf{NLL of samples under decoded density.} The samples from $\decoder$ or $\decoderhead$ are more likely under $\decoderdense$ than the ground truth trajectories or samples from an alternative model (Motion-I2V~\cite{shi2024motioni2v}). This strengthens the assumption that all our decoders decode a shared possibility space.}
    \label{fig:sup:nll_alignment}
\end{figure}

%% file: tab/X_sup/single_shot.tex
\begin{table}[b]
    \centering
    \footnotesize
    \setlength{\tabcolsep}{4.7pt}
    \begin{tabular}{lcc}
        \toprule
        \textbf{Method}
        & \textbf{Goals}
        & \textbf{EPE}$\downarrow$
        \\
        \midrule
        CogVideoX~\cite{yang2024cogvideox}      & \notavail & 0.020 \\
        Motion-I2V~\cite{shi2024motioni2v}    & 4         & 0.036 \\
        FPT~\cite{baumann2025whatif}      & 4         & 0.123 \\
        \oursours & 4      & \textbf{0.015} \\
        \bottomrule
    \end{tabular}
    \caption{\textbf{Single-Shot performance.} \ours~remains competitive with baselines in the single-shot setting. Evaluated on a subset of the full OpenVid~\cite{nan2024openvid} validation data.}
    \label{tab:sup:best_of_one}
\end{table}

%% file: fig/X_sup/qualitative_comparison/qual_comparison.tex
\begin{figure}[tbh]
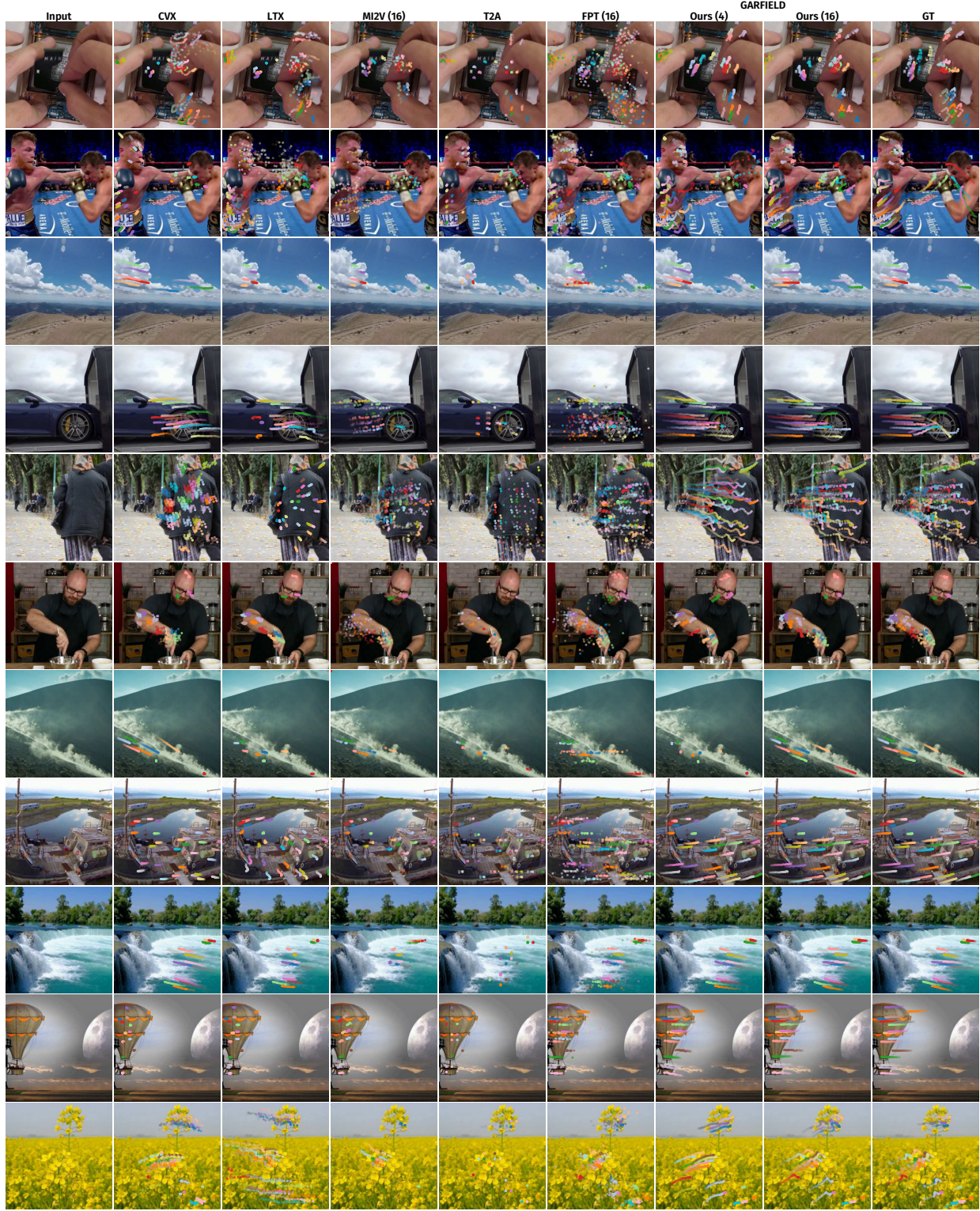

    \centering
    \newcommand{\heading}[1]{\tiny \textbf{#1}}
    \newcommand{\imgwidth}[0]{0.1\textwidth}
    \newcommand{\img}[2]{\includegraphics[width=\imgwidth]{fig/X_sup/qualitative_comparison/sample_#1/#2.pdf}}
    \newcommand{\imgrow}[1]{
    \img{#1}{input} & \img{#1}{CogVideoX} & \img{#1}{LTX} & \img{#1}{MotionI2V_16} & \img{#1}{track2act} & \img{#1}{FPT_16} & \img{#1}{Ours_4} & \img{#1}{Ours_16} & \img{#1}{gt}
    }
    \setlength{\tabcolsep}{.5pt}
    \renewcommand{\arraystretch}{0.25}
    \adjustbox{width=\textwidth}{
    \begin{tabular}{ccccccccc}
        \heading{} & \heading{} & \heading{} & \heading{} & \heading{} & \heading{} & \multicolumn{2}{c}{\tiny \textbf{GARFIELD}} & \heading{}\\
        \heading{Input} & \heading{CVX} & \heading{LTX} & \heading{MI2V (16)} & \heading{T2A} & \heading{FPT (16)} & \heading{Ours (4)} & \heading{Ours (16)} & \heading{GT}\\
        \imgrow{36} \\
        \imgrow{39} \\
        \imgrow{62} \\
        \imgrow{72} \\
        \imgrow{76} \\
        \imgrow{82} \\
        \imgrow{410} \\
        \imgrow{552} \\
        \imgrow{570} \\
        \imgrow{582} \\
        \imgrow{586} \\
    \end{tabular}
    }
    \caption{\textbf{Qualitative Comparison on OpenVid-1M~\cite{nan2024openvid}.} We select samples with high motion magnitude and ignore static and nearly static points for visualization. Similar to \cref{tab:exp:motion_comparison} we compare CogVideoX~\cite{yang2024cogvideox}(CVX), LTX~\cite{HaCohen2024LTXVideo}, Motion-I2V~\cite{shi2024motioni2v} with 16 conditioning points (MI2V (16)), Track2Act~\cite{bharadhwaj2024track2act} (T2A), and the Flow Poke Transformer~\cite{baumann2025whatif} with 16 conditioning points (FPT (16)) to \ours.}
    \label{fig:sup:qual_comparison}
\end{figure}

%% file: fig/X_sup/qualitative_densities/qual_densities.tex
\begin{figure}[tbh]
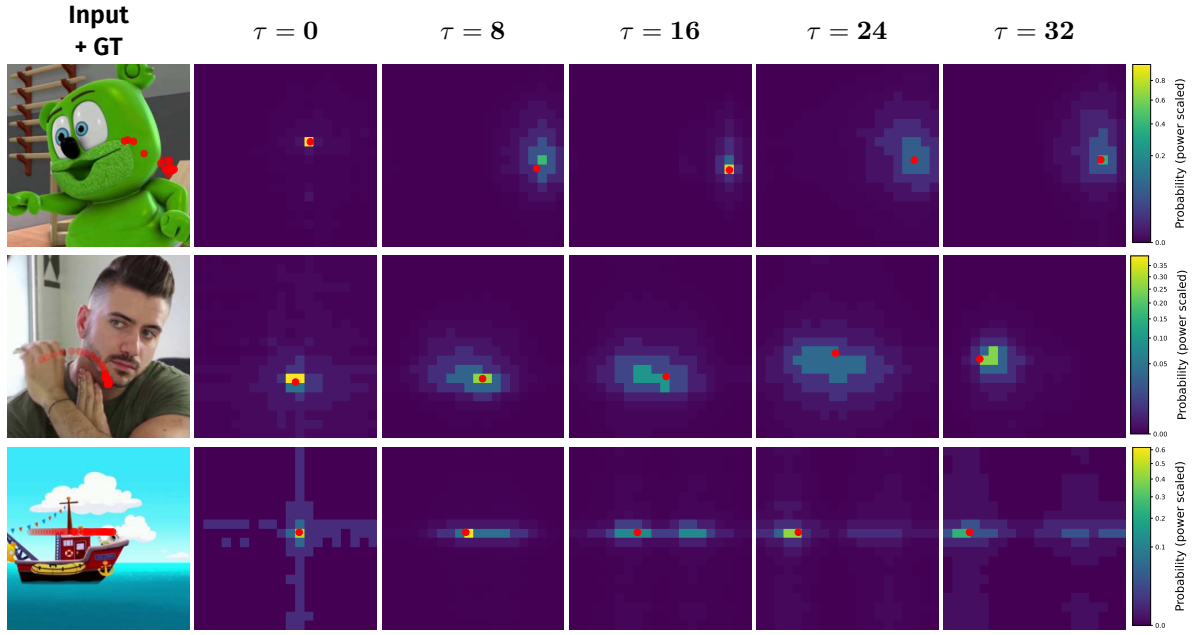

    \centering
    \setlength{\tabcolsep}{1pt}
    \newcommand{\imgsize}{0.145\textwidth}
    \newcommand{\img}[3]{\includegraphics[width=\imgsize]{fig/X_sup/qualitative_densities/cond#1/example_#2/#3.pdf}}
    \newcommand{\colorbar}[2]{\includegraphics[height=\imgsize]{fig/X_sup/qualitative_densities/cond#1/example_#2/colorbar.pdf}}
    \newcommand{\imgrow}[2]{\img{#1}{#2}{input} & \img{#1}{#2}{density_t0} & \img{#1}{#2}{density_t4} & \img{#1}{#2}{density_t7} & \img{#1}{#2}{density_t12} & \img{#1}{#2}{density_t15} & \colorbar{#1}{#2}}
    \begin{tabular}{ccccccc}
        \makecell[c]{\textbf{Input}\\\textbf{+ GT}} & $\mathbf{\tau = 0}$ & $\mathbf{\tau = 8}$ & $\mathbf{\tau = 16}$ & $\mathbf{\tau = 24}$ & $\mathbf{\tau = 32}$ &  \\
        \imgrow{4_ps}{1} \\
        \imgrow{4_ps}{2} \\
        \imgrow{4_ps}{3} \\
        \end{tabular}
    \caption{\textbf{Density Samples when conditioned on $|\condtracks| = 4$ future positions.} \ours~consistently predicts high certainty close to the ground-truth, but remains reasonably uncertain about exact speed.}
    \label{fig:sup:qual_densities_cond4}
\end{figure}

\begin{figure}[tbh]
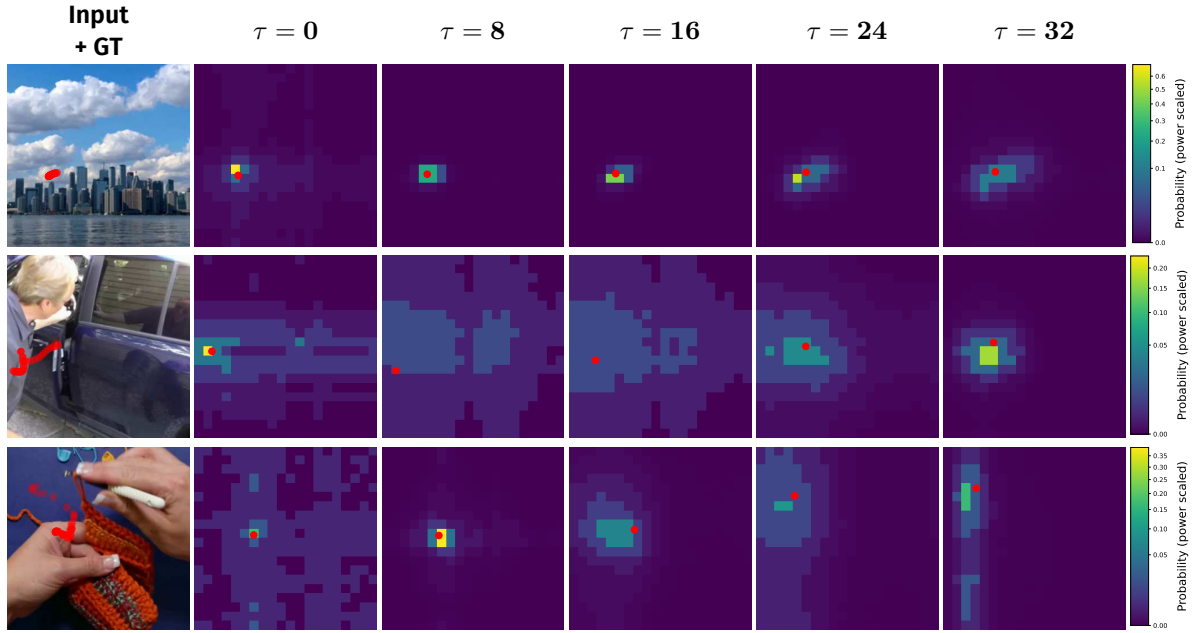

    \centering
    \setlength{\tabcolsep}{1pt}
    \newcommand{\imgsize}{0.145\textwidth}
    \newcommand{\img}[3]{\includegraphics[width=\imgsize]{fig/X_sup/qualitative_densities/cond#1/example_#2/#3.pdf}}
    \newcommand{\colorbar}[2]{\includegraphics[height=\imgsize]{fig/X_sup/qualitative_densities/cond#1/example_#2/colorbar.pdf}}
    \newcommand{\imgrow}[2]{\img{#1}{#2}{input} & \img{#1}{#2}{density_t0} & \img{#1}{#2}{density_t4} & \img{#1}{#2}{density_t7} & \img{#1}{#2}{density_t12} & \img{#1}{#2}{density_t15} & \colorbar{#1}{#2}}
    \begin{tabular}{ccccccc}
        \makecell[c]{\textbf{Input}\\\textbf{+ GT}} & $\mathbf{\tau = 0}$ & $\mathbf{\tau = 8}$ & $\mathbf{\tau = 16}$ & $\mathbf{\tau = 24}$ & $\mathbf{\tau = 32}$ &  \\
        \imgrow{2_ps}{2} \\
        \imgrow{2_ps}{1} \\
        \imgrow{2_ps}{3} \\
        \end{tabular}
    \caption{\textbf{Density Samples when conditioned on $|\condtracks| = 2$ future positions.} While \ours~remains uncertain at some points in the spatio-temporal volume, motion tends to align with the observed direction.}
    \label{fig:sup:qual_densities_cond2}
\end{figure}

\begin{figure}[tbh]
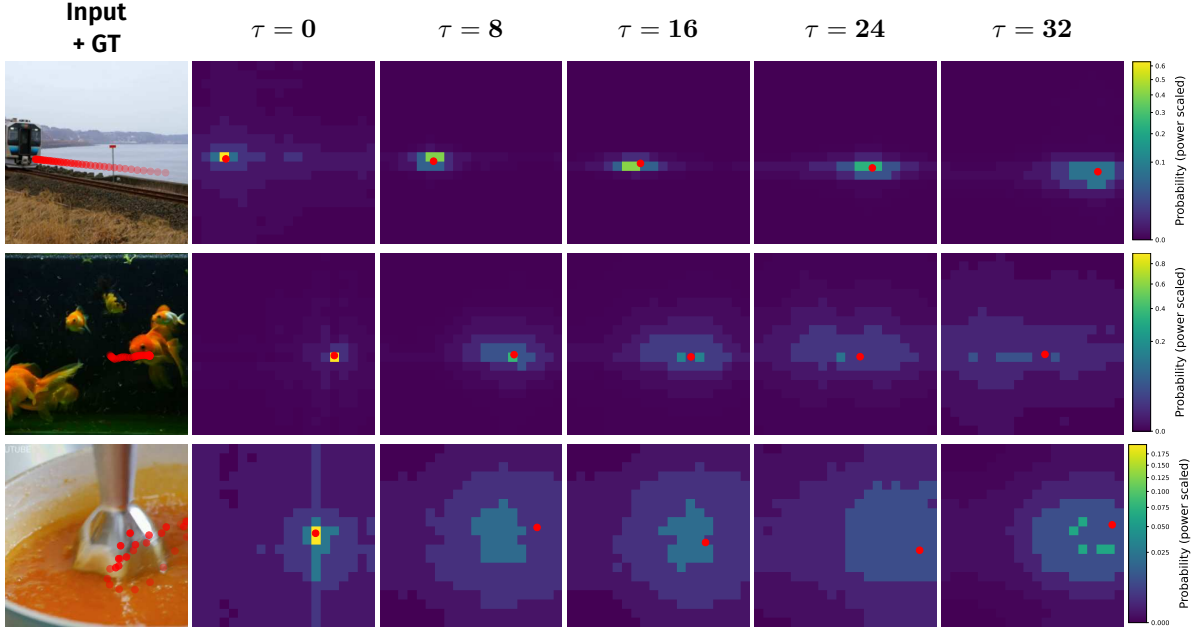

    \centering
    \setlength{\tabcolsep}{1pt}
    \newcommand{\imgsize}{0.145\textwidth}
    \newcommand{\img}[3]{\includegraphics[width=\imgsize]{fig/X_sup/qualitative_densities/cond#1/example_#2/#3.pdf}}
    \newcommand{\colorbar}[2]{\includegraphics[height=\imgsize]{fig/X_sup/qualitative_densities/cond#1/example_#2/colorbar.pdf}}
    \newcommand{\imgrow}[2]{\img{#1}{#2}{input} & \img{#1}{#2}{density_t0} & \img{#1}{#2}{density_t4} & \img{#1}{#2}{density_t7} & \img{#1}{#2}{density_t12} & \img{#1}{#2}{density_t15} & \colorbar{#1}{#2}}
    \begin{tabular}{ccccccc}
        \makecell[c]{\textbf{Input}\\\textbf{+ GT}} & $\mathbf{\tau = 0}$ & $\mathbf{\tau = 8}$ & $\mathbf{\tau = 16}$ & $\mathbf{\tau = 24}$ & $\mathbf{\tau = 32}$ &  \\
        \imgrow{0_ps}{2} \\
        \imgrow{0_ps}{3} \\
        \imgrow{0_ps}{1} \\
        \end{tabular}
    \caption{\textbf{Density Samples without future conditioning.} Without future knowledge \ours~remains uncertain about exact motion that occurs but is able to estimate the correct area, based on it's general world knowledge.}
    \label{fig:sup:qual_densities_cond0}
\end{figure}

%% file: fig/X_sup/qualitative_robotics/robotics_qualitative.tex
\begin{figure}[tbh]
    \centering
    \setlength{\tabcolsep}{1.5pt}
    \newcommand{\img}[1]{\includegraphics[width=0.155\textwidth]{fig/X_sup/qualitative_robotics/sample_#1.pdf}}
    \begin{tabular}{ccccccc}
         \img{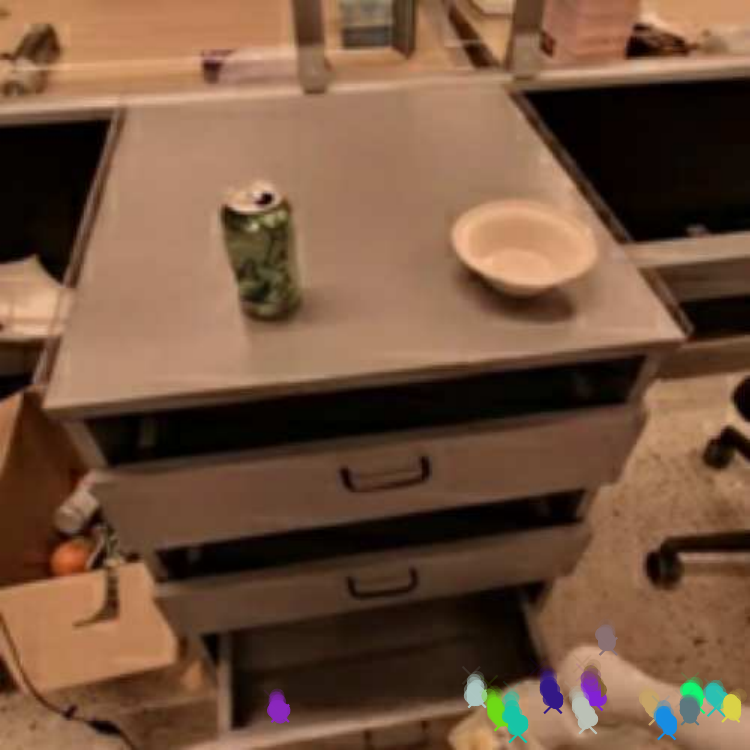} & \img{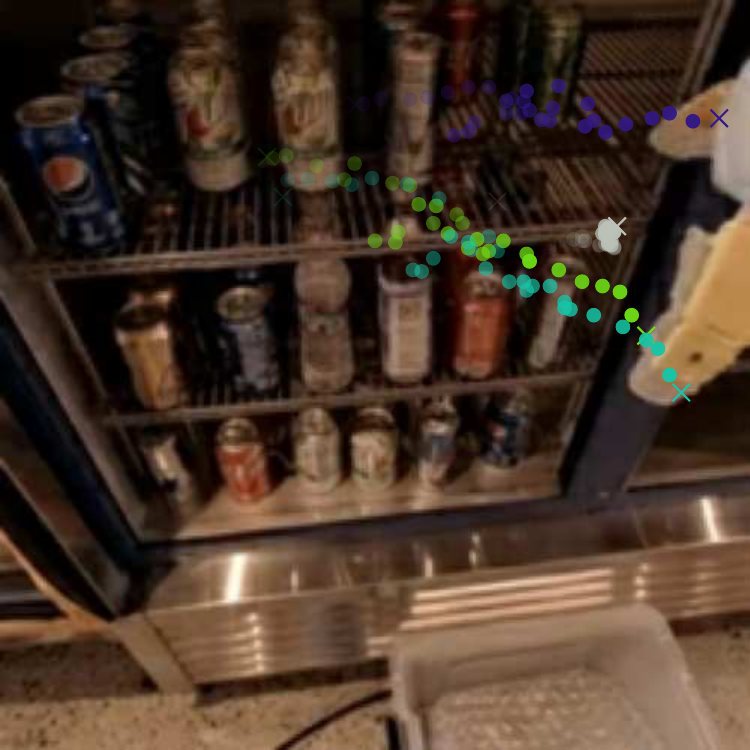} & \img{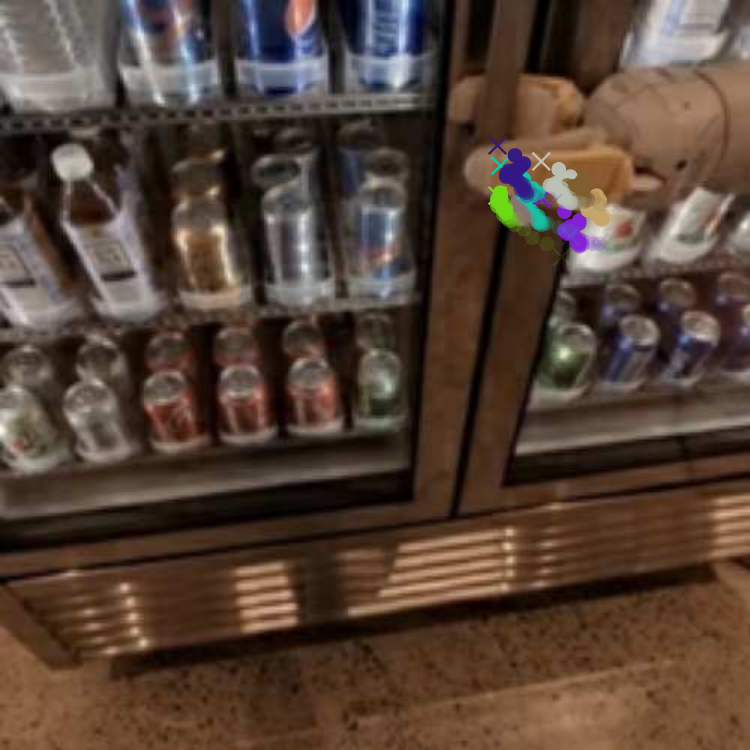} & \img{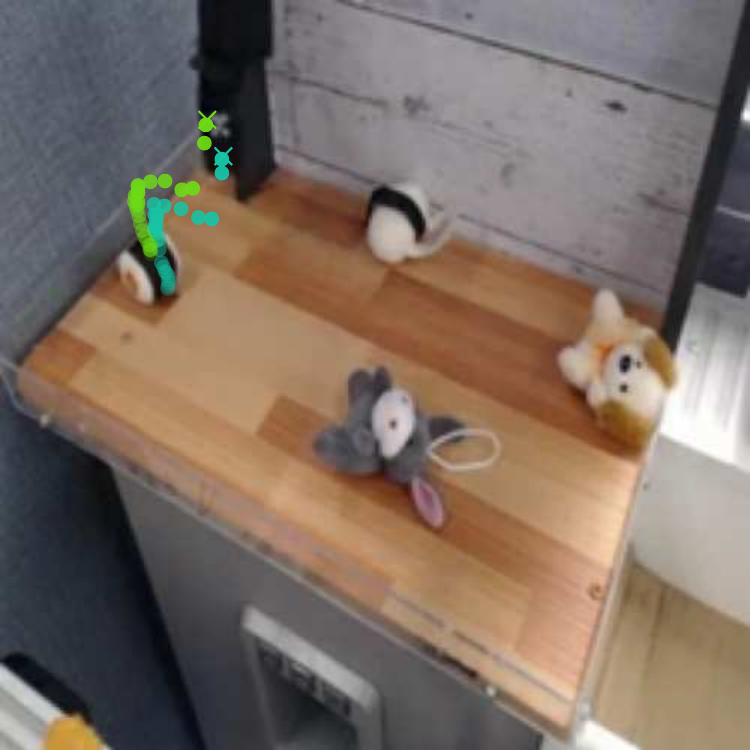} & & \img{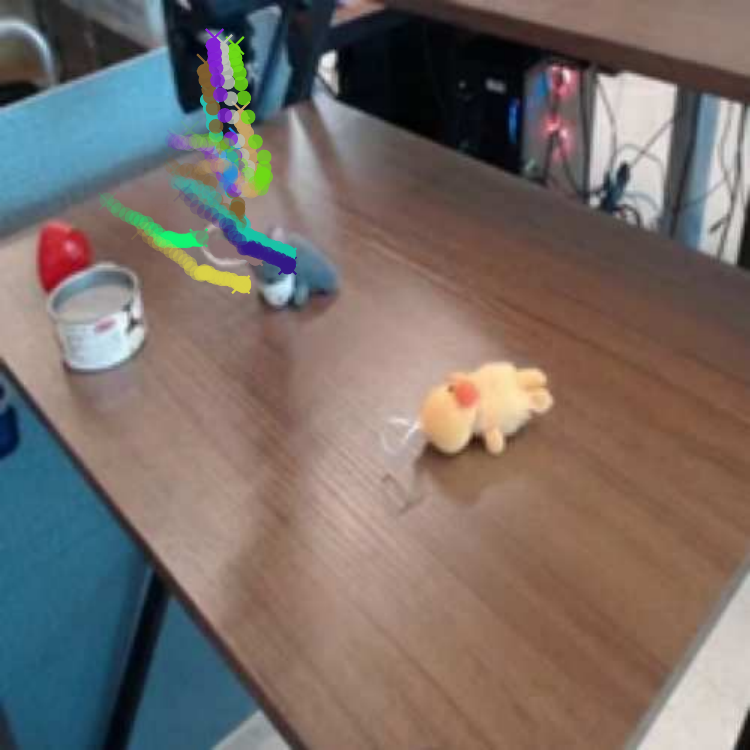} & \img{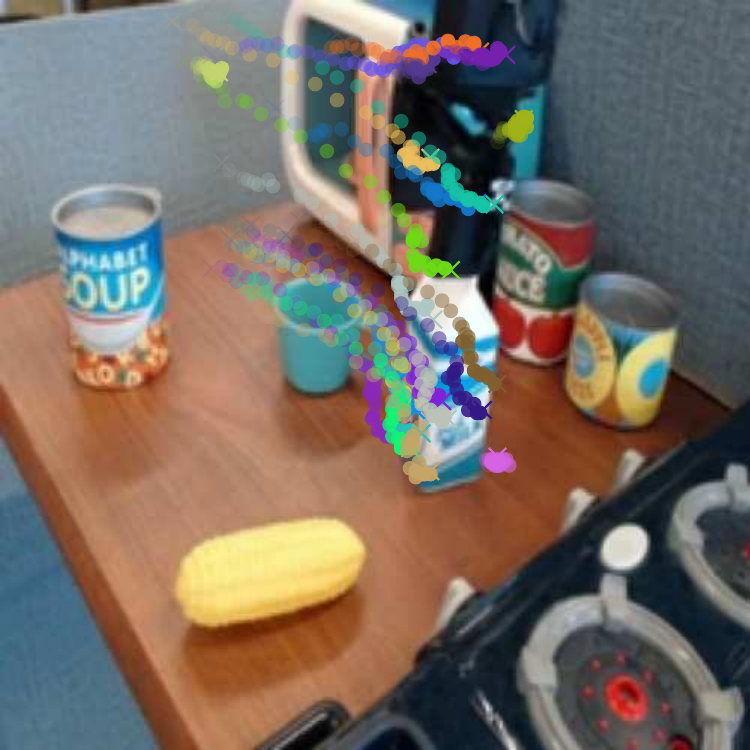}\\
    \end{tabular}
    \caption{\textbf{Qualitative Robotic Planning.} Samples from \ours~on examples in the RT1~\cite{rt12022arxiv} (first three) and BridgeData~\cite{walke2023bridgedata} (last three) datasets. We show only trajectories exceeding a motion threshold and further subsample trajectories if necessary to achieve an interpretable visualization.}
    \label{fig:sup:qual_robotics}
\end{figure}

%% file: fig/X_sup/qualitative_pedestrians/qual_pedestrians.tex
\begin{figure}[tbh]
    \setlength{\tabcolsep}{1.5pt}
    \centering
    \newcommand{\img}[1]{\includegraphics[width=0.19\textwidth]{fig/X_sup/qualitative_pedestrians/#1.pdf}}
    \newcommand{\sddimg}[1]{\includegraphics[height=0.26\textwidth]{fig/X_sup/qualitative_pedestrians/#1.pdf}}
    \newcommand{\imgrotated}[1]{\includegraphics[width=0.26\textwidth, angle=90]{fig/X_sup/qualitative_pedestrians/#1.pdf}}
    \begin{tabular}{ccccc}
        \img{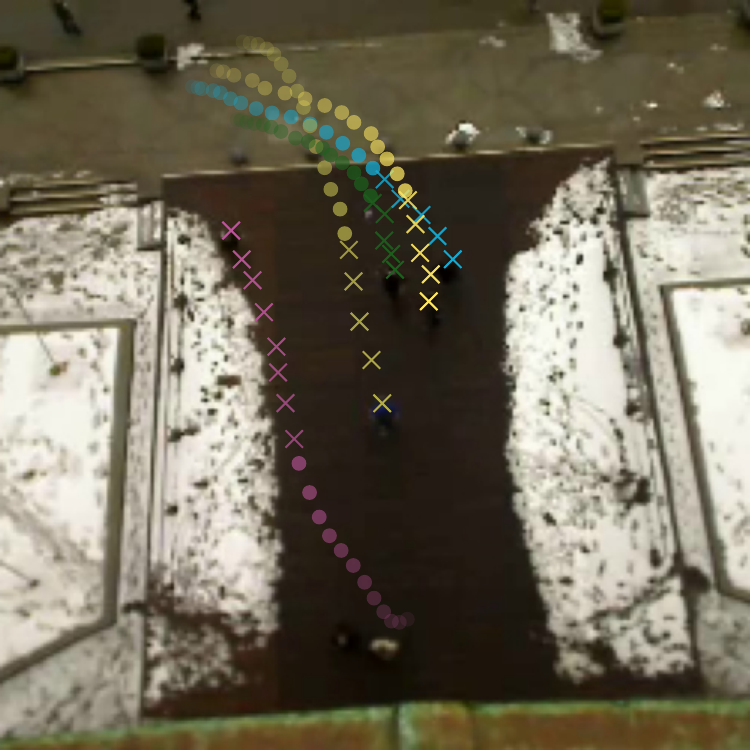} & \img{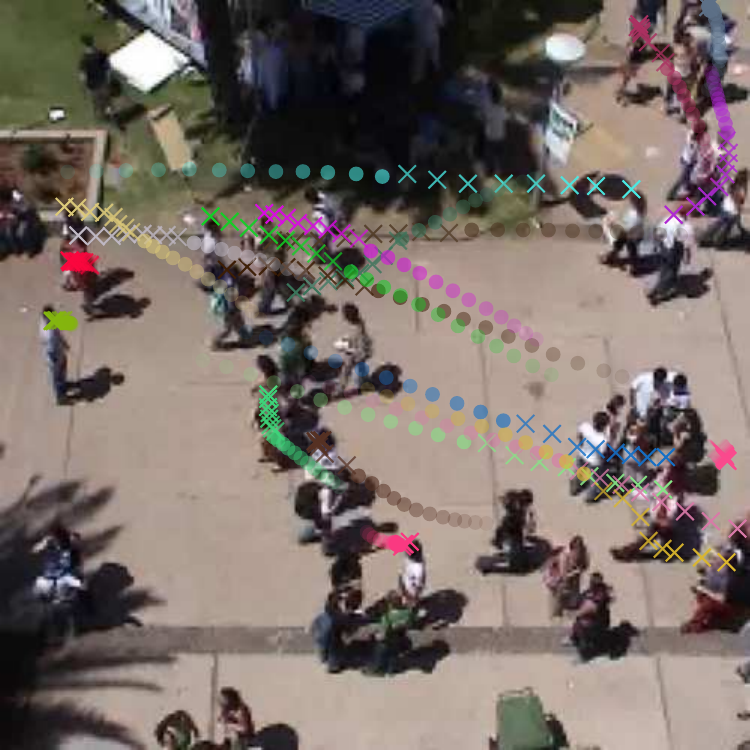} & \img{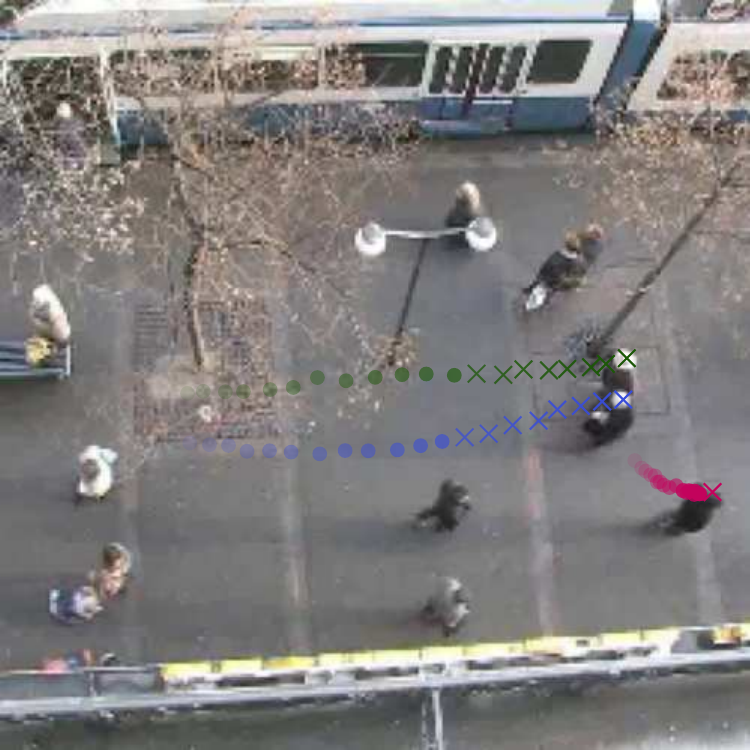} & \img{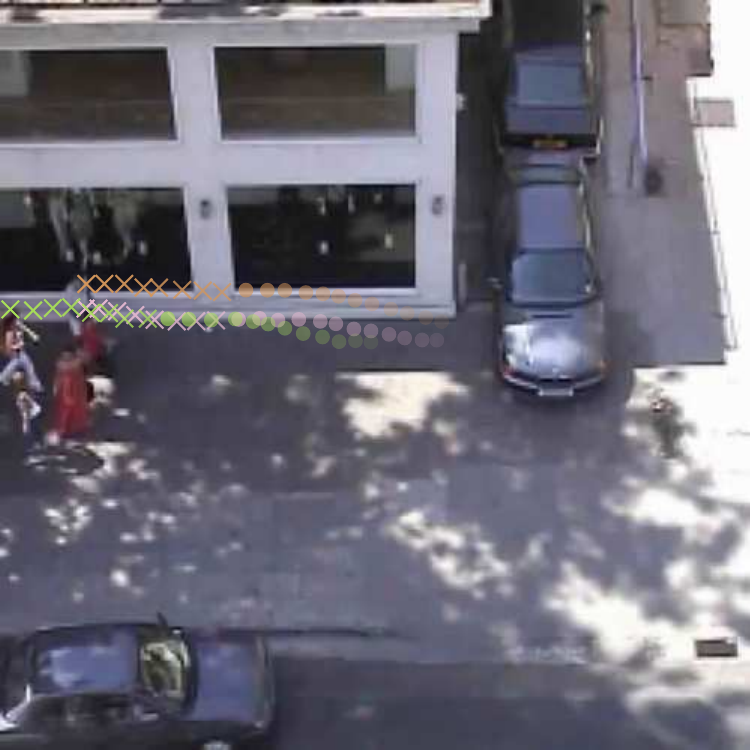} & \img{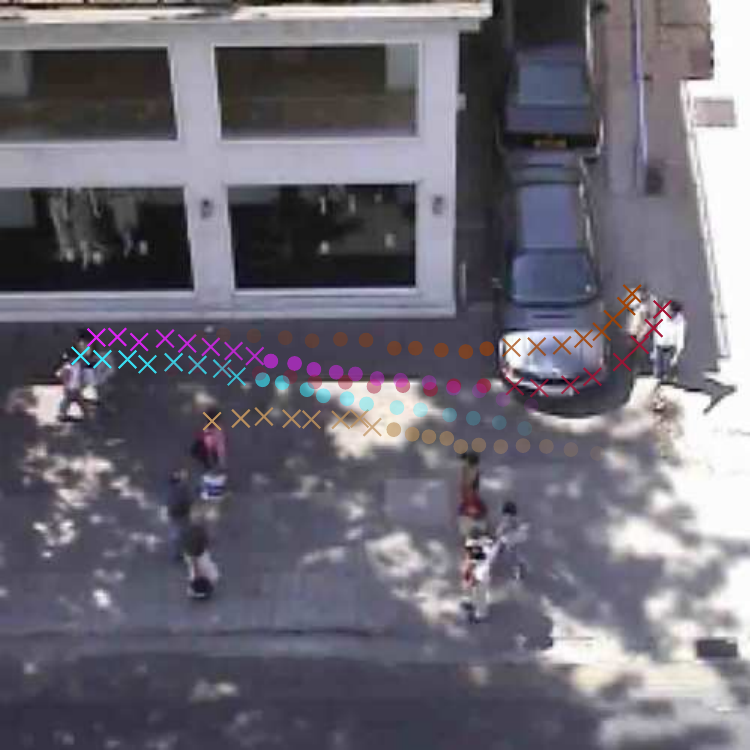} \\
        \sddimg{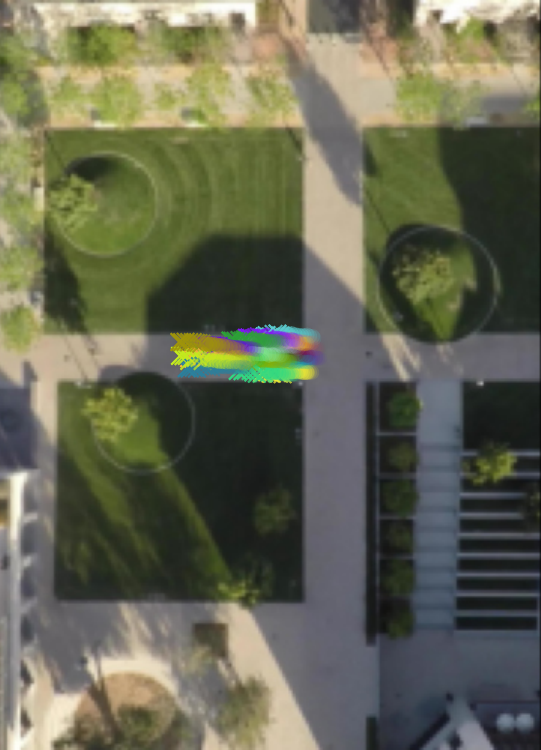} & \sddimg{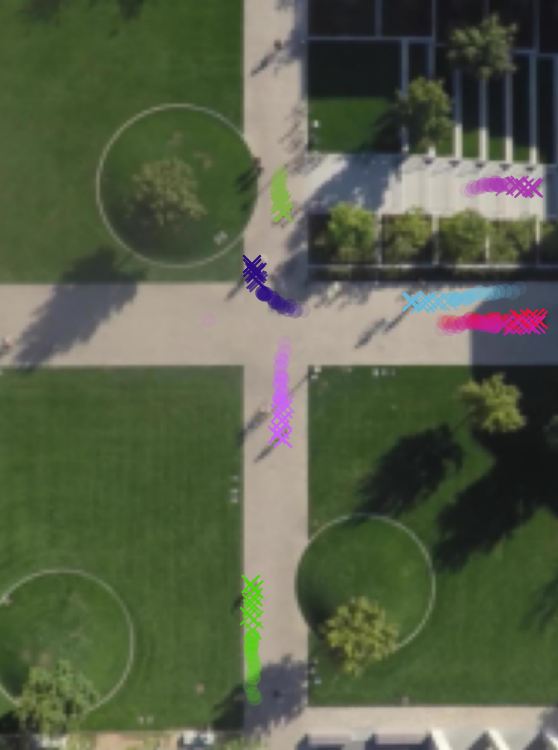} & \sddimg{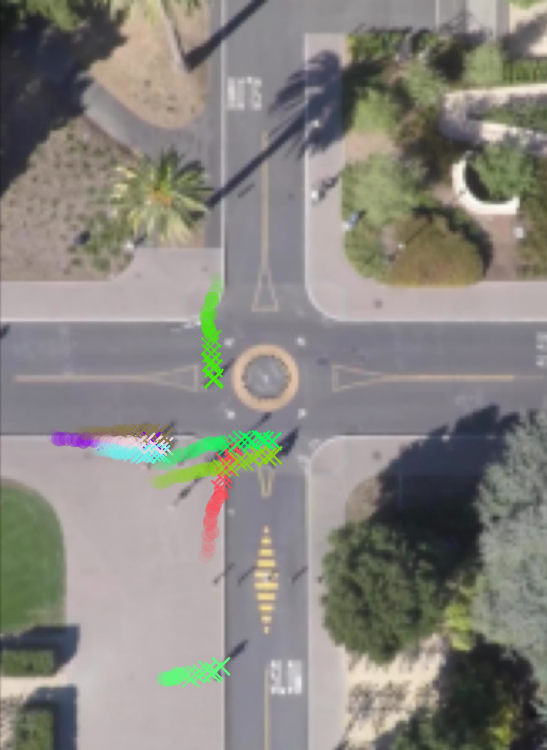} & \sddimg{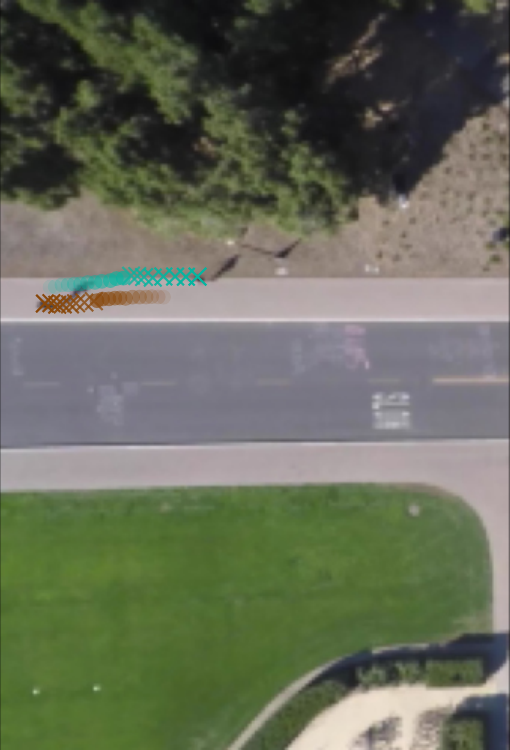} & \imgrotated{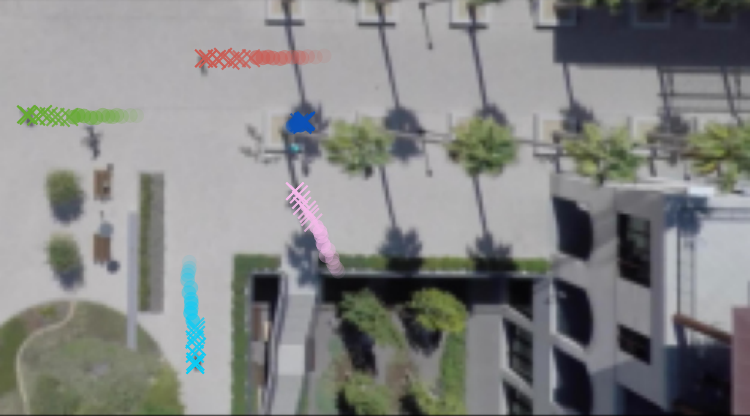} \\
    \end{tabular}
    \caption{\textbf{Qualitative Pedestrian Trajectory Prediction.} Trajectories predicted with \ours~for ETH/UCY~\cite{pellegrini2009eth, lerner2007ucy} (top row) and SDD~\cite{robicquet2016sdd}. Crosses ($\times$) indicate given positions, dots ($\cdot$) are predictions. Trajectories are subselected to trajectories that are visible in the first and last frame to ensure proper visualization.}
    \label{fig:sup:qual_pedestrian}
\end{figure}

%% file: fig/X_sup/qualitative_billiard/billiard_qualitative.tex
\begin{figure}[tbh]
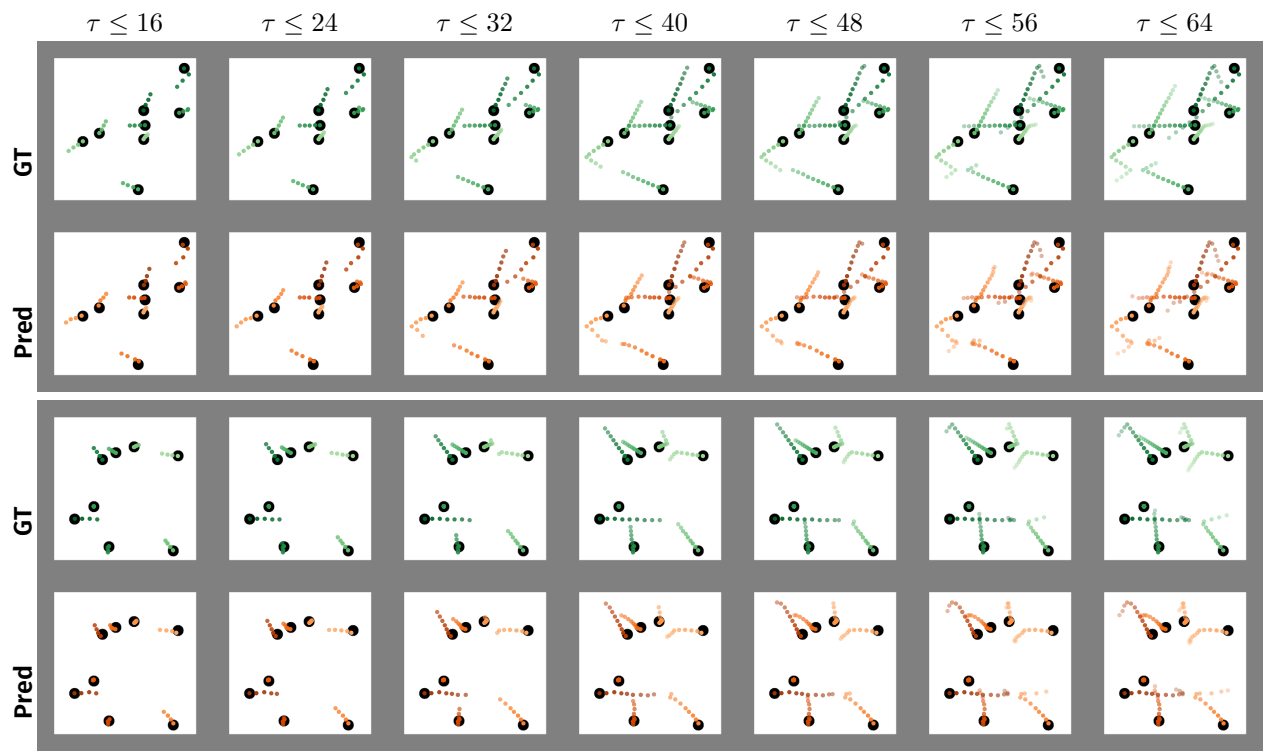

    \centering
    \setlength{\tabcolsep}{-0.1pt}
    \newcommand{\samplewidth}{0.14\textwidth}
    \newcommand{\rowlabel}[2]{%
        \raisebox{#2pt}[0pt][0pt]{\rotatebox{90}{\textbf{#1}}}%
        \hspace{2pt}%
    }
    \newcommand{\img}[3]{\includegraphics[width=\samplewidth]{fig/X_sup/qualitative_billiard/billiards_2/#1/billiard_example_#2_mean_h#3.pdf}}
    \newcommand{\imgrow}[2]{\img{#1}{#2}{16} & \img{#1}{#2}{24} & \img{#1}{#2}{32} & \img{#1}{#2}{40} & \img{#1}{#2}{48} & \img{#1}{#2}{56} & \img{#1}{#2}{60}}
    \renewcommand{\arraystretch}{1.0}
    \begin{tabular}{c ccccccc}
       & \multicolumn{1}{c}{\textbf{$\tau\leq16$}} &
       \multicolumn{1}{c}{\textbf{$\tau\leq24$}} &
       \multicolumn{1}{c}{\textbf{$\tau\leq32$}} & 
       \multicolumn{1}{c}{\textbf{$\tau \leq 40$}} &
       \multicolumn{1}{c}{\textbf{$\tau \leq 48$}} &
       \multicolumn{1}{c}{\textbf{$\tau \leq 56$}} &
       \multicolumn{1}{c}{\textbf{$\tau \leq 64$}}
       \\
    \rowlabel{GT}{15} & \imgrow{gt}{14} \\[-4pt] %
    \rowlabel{Pred}{12} & \imgrow{pred}{14} \\ %
    \rowlabel{GT}{15}  & \imgrow{gt}{33} \\[-4pt]  %
    \rowlabel{Pred}{12} & \imgrow{pred}{33} \\ %
    \end{tabular}
    \caption{\textbf{Multi-object Interactions.} \ours~is able to resolve dynamics of multiple objects colliding given conditioning on final positions and initial velocity.}\label{fig:exp:billiard_qual}
\end{figure}